%% file: lei.tex
\PassOptionsToPackage{table}{xcolor}
\documentclass[lettersize,journal]{IEEEtran}
\usepackage{amsmath,amsfonts}
\usepackage{algorithmic}
\usepackage{array}
\usepackage[caption=false,font=normalsize,labelfont=sf,textfont=sf]{subfig}
\usepackage{textcomp}
\usepackage{stfloats}
\usepackage{url}
\usepackage{verbatim}
\usepackage{graphicx}
\usepackage{booktabs}
\usepackage{multirow}
\usepackage{xspace}
\usepackage{cite}
\usepackage{newtxtext,newtxmath}
\usepackage{rotating}
\usepackage{booktabs}
\usepackage{tabularx}
\usepackage{caption}
\usepackage{hyperref}
\usepackage{enumitem}
\usepackage{xcolor}

\usepackage{tikz}
\usetikzlibrary{trees}

\definecolor{innerblue}{HTML}{CFE8FF}
\definecolor{outergreen}{HTML}{D7F9D0}

\usepackage[table]{xcolor}

\definecolor{motion}{RGB}{255,220,220}       
\definecolor{procedural}{RGB}{220,255,220}   
\definecolor{temporal}{RGB}{220,220,255}     
\definecolor{vlm}{RGB}{230,210,255}          

\definecolor{motionprocedural}{RGB}{255,235,220}    
\definecolor{proceduralvlm}{RGB}{220,235,255}       
\definecolor{motionvlm}{RGB}{255,220,255}           
\definecolor{temporalprocedural}{RGB}{220,235,255}  

\def\BibTeX{{\rm B\kern-.05em{\sc i\kern-.025em b}\kern-.08em
    T\kern-.1667em\lower.7ex\hbox{E}\kern-.125emX}}
\usepackage{balance}

\makeatletter
\DeclareRobustCommand\onedot{\futurelet\@let@token\bmv@onedotaux}
\def\bmv@onedotaux{\ifx\@let@token.\else.\null\fi\xspace}
\def\eg{\emph{e.g}\onedot} 
\def\ie{\emph{i.e}\onedot}

\def\etal{\emph{et al}\onedot}
\makeatother

\begin{document}
\title{Video Understanding by Design: How Datasets Shape Video Models
}
\author{Lei Wang, Syuan-Hao Li, Piotr Koniusz, Yongsheng Gao
\thanks{
This work is supported in part by the Australian Research Council (ARC) under Industrial Transformation Research Hub Grant IH180100002 (Corresponding author: Yongsheng Gao).

Lei Wang, Syuan-Hao Li, and Yongsheng Gao are with the School of Engineering and Built Environment, Electrical and Electronic Engineering, Griffith University (email: l.wang4@griffith.edu.au, syuan-hao.li@griffithuni.edu.au, yongsheng.gao@griffith.edu.au).

Piotr Koniusz is with School of Computer Science and Engineering, University of New South Wales (email: piotr.koniusz@unsw.edu.au).
}}


\maketitle

\begin{abstract}
Research in video understanding has advanced rapidly, driven by increasingly diverse datasets and more powerful model architectures. While existing surveys typically organize progress by tasks, benchmarks, or model families, they provide limited insight into why particular architectures emerged and succeeded. In this survey, we argue that the evolution of video understanding is fundamentally shaped by dataset structure. Properties such as motion complexity, temporal span, compositional hierarchy, multi-agent interactions, and multimodal richness impose distinct learning challenges and reasoning requirements on models.
We present \emph{a dataset-centric perspective} that connects dataset structure, inductive biases, and architectural design within a unified framework. We show that different datasets require models to capture specific invariances and capabilities, such as robustness to viewpoint changes, sensitivity to temporal ordering, reasoning over long-range dependencies, relational interactions, and cross-modal alignment. These requirements naturally give rise to inductive biases, \ie, architectural assumptions that favor particular patterns of reasoning and generalization. From this perspective, milestone architectures, including two-stream networks, 3D CNNs, temporal models, transformers, graph-based methods, and multimodal foundation models, can be understood as architectural responses to the challenges posed by evolving datasets.
Building on this framework, we systematically analyze how dataset characteristics have shaped architectural innovation across video understanding tasks and discuss the representational biases induced by different data regimes. We further provide practical guidance for aligning model design with dataset properties and task requirements. By unifying datasets, inductive biases, and architectures into a coherent perspective, this survey offers both a retrospective explanation of the field's evolution and a forward-looking roadmap toward general-purpose video understanding systems. Code and dynamic video visualizations of dataset-induced biases are available at \url{https://time.griffith.edu.au/paper-sites/video-understanding/}.
\end{abstract}


\section{Introduction}

\IEEEPARstart{V}{ideo} understanding has become one of the central problems in computer vision, enabling applications ranging from surveillance, autonomous driving, and robotics to healthcare, education, and multimedia retrieval \cite{karpathy2014large,tran2015learning,feichtenhofer2019slowfast,chen2025motion,wang2024taylor,ding2025learnable,wang2024meet,qiu2025evolving,wang2025feature,ding2025language,ding2024quo,ding2025journey,liurepresentation}. Unlike images, videos contain rich temporal dynamics, human-object interactions, long-term activities, and multimodal signals that evolve over time. As a result, video understanding extends beyond recognizing visual appearance and requires models to reason about motion, temporal dependencies, procedural structure, social interactions, and semantic context \cite{madan2024foundation,tang2025video}.

\begin{figure}[tbp]
\centering
\includegraphics[trim=0 0 0 0, clip=true, width=\linewidth]{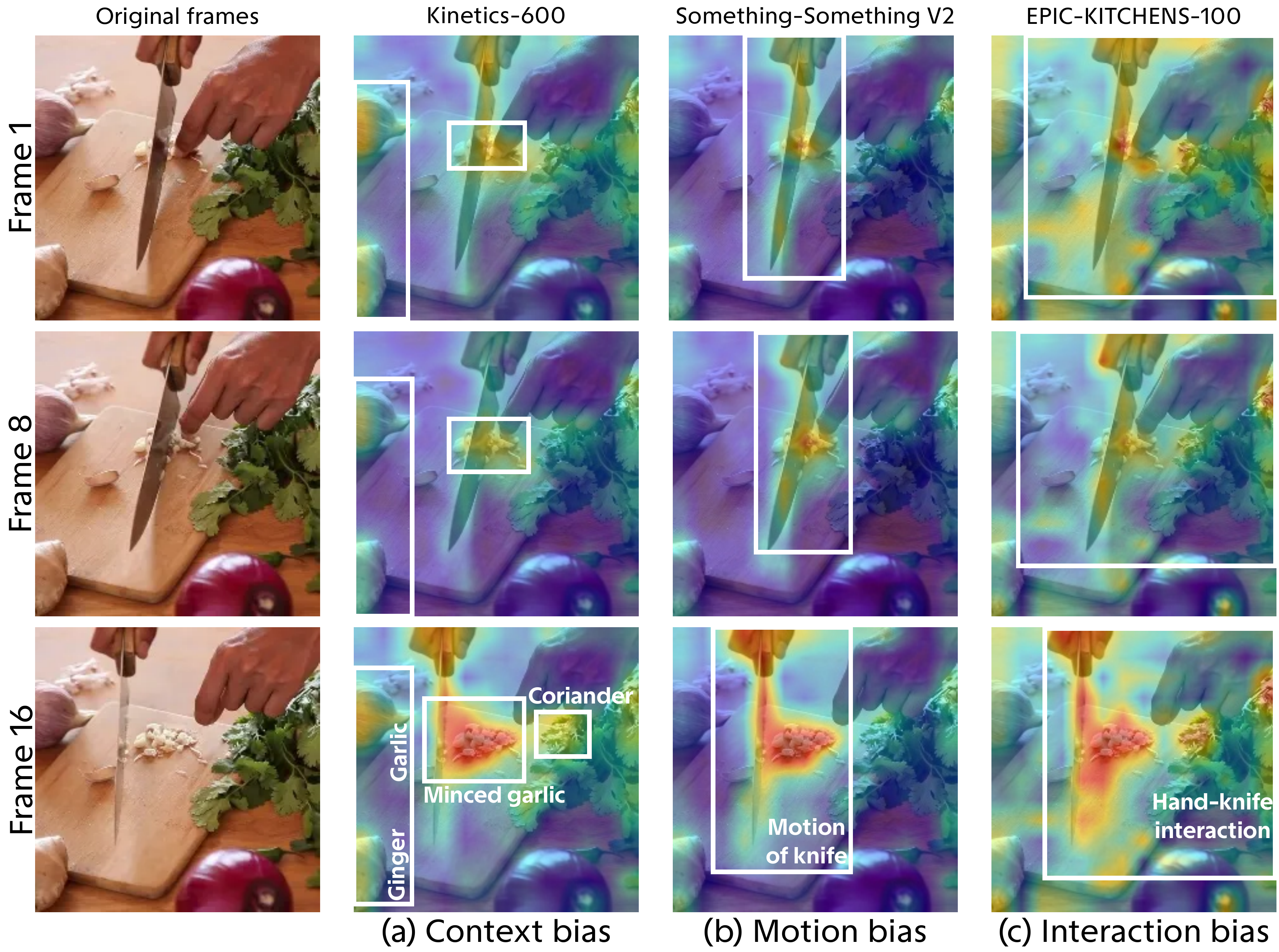}
\caption{
Dataset-induced representational biases. DAAM~\cite{daam} visualizations of MotionFormer~\cite{patrick2021keeping} pretrained on different datasets. (a) Kinetics-600 dataset exhibits a \emph{context bias}, attending to surrounding objects, including the ginger, garlic, and coriander; (b) Something-Something V2 dataset exhibits a \emph{motion bias}, emphasizing the knife's motion trajectory during cutting; and (c) EPIC-KITCHENS-100 dataset exhibits an \emph{interaction bias}, focusing on the hand-knife interaction and object manipulation. 
Videos are from OWM~\cite{baumann2026envisioning}.
}
\label{fig:data-driven}
\end{figure}

Over the past two decades, remarkable progress has been achieved in video understanding. Early research focused primarily on recognizing short and isolated actions using handcrafted representations and compact benchmarks \cite{soomro2012ucf101,kuehne2011hmdb,klaser2008spatio,wang2013action}. Today, modern video systems are expected not only to classify actions, but also to localize events, understand complex procedures, reason about human-object interactions, answer questions, anticipate future activities, and interact through natural language \cite{madan2024foundation,tang2025video}. This evolution has been accompanied by successive generations of architectures, ranging from two-stream networks and 3D convolutional models to graph neural networks, transformers, vision-language models, and video foundation models \cite{simonyan2014two,tran2015learning,wang2016graph,yan2018spatial,arnab2021vivit,bertasius2021space,wang2022internvideo,wang2024internvideo2}.
Most existing accounts describe this progress chronologically, emphasizing the emergence of new architectures, learning paradigms, or benchmark datasets. However, such a perspective explains \emph{what} changed, but not necessarily \emph{why} it changed. Why did temporal reasoning models emerge after conventional convolutional architectures? Why did transformers become increasingly dominant in long-horizon and multimodal settings? Why are graph-based models effective for relational reasoning, while vision-language models flourish in semantically rich environments? Although these developments are often discussed independently, they are closely connected through a common underlying factor: the changing structure of video datasets.

\begin{table*}[tbp]
\centering
\small
\caption{Comparison of representative 
surveys. Existing surveys 
organize the literature around tasks, architectures, modalities, learning paradigms, or benchmarks. In contrast, our survey adopts a dataset-centric perspective that explains how dataset characteristics introduce distinct learning requirements and, in turn, drive the evolution of video understanding architectures.}
\resizebox{\linewidth}{!}{
\begin{tabular}{p{3.0cm}p{3.9cm}p{4.0cm}p{6.8cm}}
\hline
\textbf{Survey category} &
\textbf{Primary focus} &
\textbf{Perspective gap} &
\textbf{Dataset-centric perspective} \\
\hline

{Classical \& deep-learning surveys}
(\eg, \cite{aggarwal2011human,herath2017going,pareek2021survey})
&
Handcrafted representations, two-stream networks, and early deep architectures.
&
Emphasize model evolution and benchmark performance.
&
Explain how increasing motion complexity, temporal diversity, and dataset scale motivated the transition from handcrafted representations to deep spatiotemporal architectures.
\\
\hline

{Transformer \& modern architecture surveys}
(\eg, \cite{arnab2021vivit,ulhaq2022vision,selva2023video})
&
Transformers, attention mechanisms, and architectural design trade-offs.
&
Focus on architectural innovations and efficiency considerations.
&
Relate the emergence of transformers to growing demands for long-range temporal reasoning, compositional understanding, and multimodal learning.
\\
\hline

{Multimodal, VLM \& LLM surveys}
(\eg, \cite{miech2020end,xu2023multimodal,madan2024foundation,tang2025video})
&
Video-language learning, multimodal pretraining, and video foundation models.
&
Focus on alignment objectives, datasets, and multimodal capabilities.
&
Analyze how multimodal richness, semantic grounding, and cross-modal alignment requirements influence model design and learning strategies.
\\
\hline

{Structured representation surveys}
(\eg, \cite{subetha2016survey,yan2018spatial,wang2019comparative,cheng2020skeleton,ahmad2021graph,mourot2022survey,ren2024survey,liurepresentation})
&
Skeleton-based recognition, graph neural networks, and relational reasoning.
&
Primarily examine modality-specific representations.
&
Connect representational choices to dataset properties such as relational complexity, multi-agent interactions, and human-object dynamics.
\\
\hline

{Self-supervised, generative \& pretraining surveys}
(\eg, \cite{jing2020self,liu2021self,ericsson2022self,oussidi2018deep,suzuki2022survey,cho2024sora,xing2024survey,wang2025feature})
&
Contrastive learning, masked modeling, generative learning, and pretraining strategies.
&
Focus on learning objectives and downstream transfer.
&
Show how dataset characteristics influence the suitability of contrastive, masked, generative, and hybrid pretraining paradigms.
\\
\hline

{Benchmark \& dataset surveys}
(\eg, \cite{liu2016benchmarking,carreira2017quo,singh2019video,yao2019review,guo2024benchmarking,plizzari2024outlook,liurepresentation})
&
Datasets, benchmarks, evaluation protocols, and performance comparisons.
&
Primarily characterize datasets and evaluation practices.
&
Examine how properties such as motion complexity, temporal span, compositional structure, relational interactions, and multimodal richness influence architectural development.
\\
\hline

{Foundation-model surveys}
(\eg, \cite{madan2024foundation,ding2025language})
&
Video foundation models, large-scale pretraining, and multimodal reasoning.
&
Focus on scaling, capabilities, and emerging applications.
&
Place foundation models within a broader evolutionary framework shaped by increasingly large, diverse, and multimodal data regimes.
\\
\hline

\textbf{Ours (2026)}
&
Comprehensive coverage from classical action recognition to multimodal and foundation-model-based video understanding.
&
-

&
Provides a unified dataset-centric framework that explains how dataset characteristics shape learning requirements and drive architectural evolution. Offers both a retrospective explanation of the field's progression and a roadmap toward general-purpose video understanding.
\\
\hline
\end{tabular}}
\label{tab:survey_comparison}
\end{table*}

We argue that datasets are not merely evaluation benchmarks; they define the learning requirements that models should satisfy. Different datasets emphasize different aspects of video understanding. Motion-centric benchmarks require fine-grained discrimination of temporal dynamics. Procedural datasets demand reasoning over hierarchical and multi-step activities. Egocentric datasets introduce hand-object interactions and long-term dependencies. Multimodal corpora require alignment across visual, auditory, and linguistic information. As datasets evolve, they introduce new challenges and reasoning requirements, which in turn shape the representations, learning paradigms, and architectural designs that prove most effective. 
%
Viewed through this lens, the history of video understanding becomes a sequence of architectural adaptations to increasingly complex data regimes. Datasets such as UCF101 \cite{soomro2012ucf101} and HMDB51 \cite{kuehne2011hmdb} primarily emphasized salient human motion and short-term temporal dynamics, motivating two-stream and 3D convolutional architectures \cite{simonyan2014two,tran2015learning}. The emergence of Something-Something \cite{goyal2017something}, ActivityNet \cite{caba2015activitynet}, Charades \cite{sigurdsson2016hollywood}, AVA \cite{li2020ava}, Breakfast \cite{kuehne2014language}, and EPIC-KITCHENS \cite{damen2018scaling} introduced challenges involving object-centric interactions, procedural understanding, long-range temporal dependencies, and multi-agent reasoning, accelerating the development of temporal reasoning networks, graph-based methods, and transformers \cite{goyal2017something,caba2015activitynet,sigurdsson2018charades,gu2018ava,damen2018scaling,wang2018non,wang2016graph,yan2018spatial,arnab2021vivit,bertasius2021space}. More recently, large-scale multimodal datasets such as HowTo100M \cite{miech2019howto100m}, WebVid \cite{bain2021frozen}, InternVid \cite{wang2023internvid}, Panda \cite{chen2024panda}, and Koala \cite{wang2025koala} have fostered the rise of vision-language pretraining and video foundation models capable of semantic grounding, multimodal reasoning, and open-world understanding \cite{miech2019howto100m,wang2022internvideo,wang2024internvideo2,madan2024foundation}. Rather than viewing these architectures as isolated innovations, we interpret them as responses to the learning requirements imposed by evolving datasets.

Despite extensive progress, existing surveys remain fragmented. Some focus on specific tasks, such as action recognition, temporal localization, video retrieval, or video question answering. Others review particular architectural families, including CNNs, recurrent networks, transformers, and multimodal foundation models \cite{sun2022human,nguyen2024video,madan2024foundation,10.1016/j.jvcir.2024.104320,zhou2024survey,tang2025video}. While these surveys provide valuable summaries of methods, benchmarks, and performance trends, they primarily describe \emph{what} architectures have been developed and \emph{how} they perform. Comparatively little attention has been devoted to understanding \emph{why} particular architectural paradigms emerged and succeeded. As a result, the field still lacks a unifying perspective that explains how dataset characteristics, learning requirements, learned representations, and architectural evolution interact over time.
Fig.~\ref{fig:data-driven} provides a motivating example. Despite sharing the same architecture and processing identical video inputs, MotionFormer models pretrained on Kinetics-600, Something-Something V2, and EPIC-KITCHENS-100 attend to markedly different aspects of the scene. In this example, Kinetics-600 focuses on surrounding objects such as the ginger, garlic, and coriander, Something-Something V2 emphasizes the knife's motion trajectory, and EPIC-KITCHENS-100 concentrates on the hand-knife interaction. These contrasting attention patterns suggest that different datasets encourage different representational biases, shaping what information models learn to prioritize. This observation motivates a central question of this survey: \emph{how do dataset characteristics influence representational biases and, ultimately, architectural evolution?}

In this survey, we address this question through a dataset-centric perspective on video understanding. Rather than organizing the literature solely by tasks, architectures, modalities, or learning paradigms, we examine how dataset characteristics shape learning requirements, representational biases, and architectural design. We show that recurring dataset properties, including motion complexity, temporal span, compositional structure, relational interactions, and multimodal richness, have repeatedly driven the emergence of new model architectures and learning paradigms throughout the history of video understanding. This perspective not only provides a coherent explanation for the field's evolution, but also offers practical guidance for future dataset construction, model development, and video foundation-model design.
Our key contributions are summarized as follows:
\renewcommand{\labelenumi}{\roman{enumi}.}
\begin{enumerate}[leftmargin=0.5cm]
\item 
We present a unified framework that connects dataset characteristics, learning requirements, inductive biases, learning paradigms, and architectural design, providing a principled explanation of how video understanding has evolved over the past two decades.
\item 
We analyze how motion complexity, temporal span, compositional structure, relational interactions, and multimodal richness have shaped the development of major architectural paradigms, including two-stream networks, 3D CNNs, graph-based models, transformers, and multimodal foundation models.
\item 
By examining diverse benchmarks spanning action recognition, temporal localization, procedural understanding, egocentric perception, and multimodal reasoning, we identify recurring patterns that explain both strengths and limitations of existing approaches.
\item 
Building on these insights, we discuss future directions for aligning dataset design, learning paradigms, and architectural principles, providing guidance for the development of scalable and robust video foundation models.
\end{enumerate}


\section{Positioning Within Prior Surveys}
\label{sec:related}

Video understanding has attracted extensive survey efforts. 
Representative examples include task- and architecture-centric surveys covering action recognition, temporal localization, retrieval, CNNs, transformers, and graph neural networks \cite{shih2017survey,herath2017going,aafaq2019video,sun2022human,selva2023video,madan2024foundation,tang2025video}; modality- and paradigm-centric surveys focusing on skeleton-based recognition, audio-visual learning, vision-language modeling, self-supervised learning, and generative modeling \cite{wang2019comparative,jing2020self,xu2023multimodal,xing2024survey,madan2024foundation,ding2025language}; and benchmark-oriented surveys that summarize dataset construction, evaluation protocols, and performance trends \cite{liu2016benchmarking,carreira2017quo,singh2019video,yao2019review,guo2024benchmarking,liurepresentation}.
A detailed discussion of existing surveys is provided in Appendix A.
These surveys provide valuable summaries of methods, datasets, and evaluation practices. However, they primarily focus on \emph{what} models have been developed and \emph{how} they perform. Comparatively less attention has been devoted to understanding \emph{why} particular architectural paradigms emerged and how advances in datasets, learning paradigms, and model architectures have co-evolved.

This survey adopts a complementary perspective by examining how dataset characteristics shape the learning requirements that drive architectural evolution. We argue that recurring properties of video datasets 
repeatedly introduce new challenges that motivate advances in representations, learning paradigms, and model architectures. From this perspective, major developments in video understanding 
can be viewed as responses to increasingly complex data regimes.
Table~\ref{tab:survey_comparison} summarizes the relationship between our survey and representative prior efforts. Unlike existing surveys that primarily organize the literature around tasks, architectures, modalities, learning paradigms, or benchmarks, we focus on how dataset characteristics shape the learning requirements that drive architectural evolution. By connecting datasets, learning requirements, and architectural development within a unified framework, our survey provides both a coherent explanation of the field's historical progression and practical guidance for future dataset design, model development, and general-purpose video understanding systems.

\begin{table*}[tbp]
\caption{Datasets as structural lenses for video understanding. 
The table organizes major datasets by structural properties: supervision, compositionality, multi-agent density, and temporal span, and highlights the architectural advances they spurred.
Row colors indicate the dataset's primary focus: motion/fine-grained actions (red), procedural/compositional tasks (green), temporal/stepwise tasks (blue), VLM/video-language tasks (purple), and mixed categories for overlaps.
View: 3P = third-person, Ego = egocentric. 
Mods: R = RGB, F = Flow, A = Audio, D = Depth, P = Pose/Skeleton, I = IR, T = Text, M = IMU. 
Anno: Cls = class labels, Temp = temporal segments, ST = spatiotemporal boxes, Cap = captions, Step = procedural steps, QA = question answering, Grnd = text grounding. 
Struct: Amp/Span/Comp/Agents, where Amp = motion amplitude (H/M/L), Span = temporal span (S/M/L), Comp = compositionality (-/C/H = none/compositional/hierarchical), Agents = agent density (L/M/H). 
Impulse: 2S = two-stream, 3D = 3D CNN, TSN/TCN, TRF = transformers/attention, ST-GCN = graph/skeleton, VLM = vision-language pretraining, HOI = hand-object interaction, Det = localized detection.
}
\vspace{-4pt}
\setlength{\tabcolsep}{0.15em}
\renewcommand{\arraystretch}{0.70}
\begin{center}
\resizebox{0.82\linewidth}{!}{%
\begin{tabular}{l c c c l l l l l}
\toprule
{\textbf{Dataset}} & {\textbf{Year}} & {\textbf{Scale}} & {\textbf{View}} & {\textbf{Mods}} & {\textbf{Anno}} & \textbf{Struct} & {\textbf{Primary Task}} & {\textbf{Impulse}} \\
\midrule
\rowcolor{motion}KTH actions \cite{schuldt2004recognizing} & 2004 & 2.4K clips & 3P & R & Cls & H/S/-/L & Simple actions & 2S, 3D \\
\rowcolor{motion}Weizmann \cite{blank2005actions} & 2005 & 90 clips & 3P & R & Cls & H/S/-/L & Simple actions & 2S \\
\rowcolor{motion}IXMAS Actions \cite{weinland2006free} & 2006 & 1148 clips & 3P (multi-view) & R & Cls & M/S/-/L & View invariance & 2S \\
\rowcolor{motion}Hollywood \cite{laptev2008learning} & 2008 & 1.4K clips & 3P & R & Cls & M/S/-/M & Movie actions in-the-wild & 2S \\
\rowcolor{motion}Hollywood2 \cite{marszalek2009actions} & 2009 & 1.7K clips & 3P & R & Cls & M/S/-/M & Movie actions & 2S \\
\rowcolor{motion} Collective Activity \cite{choi2009they} & 2009 & 44 videos & 3P & R & ST (groups) & M/S/-/H & Social/group acts & GNN \\
\rowcolor{motion} Olympic Sports \cite{niebles2010modeling} & 2010 & 783 clips & 3P & R & Cls & H/S/-/L & Sports actions & 2S \\
\rowcolor{motion} MSRAction3D \cite{li2010action} & 2010 & 567 clips & Kinect frontal & D,P & Cls & M/S/-/L & Depth/skeleton actions & ST-GCN \\
\rowcolor{motion} MSRActionPairs3D \cite{li2010action} & 2010 & 360 clips & Kinect frontal & R,D,P & Cls & M/S/-/L & Interaction pairs & ST-GCN \\
\rowcolor{motion} HMDB51 \cite{kuehne2011hmdb} & 2011 & 6.8K clips & 3P & R & Cls & M/S/-/M & Diverse actions & 2S, 3D \\
\rowcolor{motion} UCF101 \cite{soomro2012ucf101} & 2012 & 13K clips & 3P & R & Cls & H/S/-/L & Actions/sports & 3D, TSN \\
\rowcolor{motion} UTKinect-Action3D \cite{xia2012view} & 2012 & 199 clips & Kinect frontal & R,D,P & Cls & M/S/-/L & Skeleton actions & ST-GCN \\
\rowcolor{motion} G3D-Gaming \cite{bloom2012g3d} & 2012 & 20 classes & Kinect frontal & R,D,P & Cls & H/S/-/L & Gaming actions & ST-GCN \\

\rowcolor{motion} UCF50 \cite{reddy2013recognizing} & 2013 & 6.7K clips & 3P & R & Cls & H/S/-/L & Actions/sports & 3D \\
\rowcolor{motion} Florence3D \cite{seidenari2013recognizing} & 2013 & 215 clips & Kinect frontal & R,D,P & Cls & M/S/-/L & Skeleton actions & ST-GCN \\
\rowcolor{motion} JHMDB \cite{jhuang2013towards} & 2013 & 928 clips & 3P & R,P & ST & M/S/-/M & Pose + localized acts & Det \\
\rowcolor{motion} Sports-1M \cite{karpathy2014large} & 2014 & 1M clips & 3P & R & Cls & H/S/-/L & Large-scale sports & 3D pretrain \\

\rowcolor{motion} Northwestern-UCLA \cite{wang2014cross} & 2014 & 1.5K clips & Kinect multi-view & R,D,P & Cls & M/S/-/L & Cross-view actions & ST-GCN \\
\rowcolor{motion} UWA3D (I/II) \cite{wang2019comparative} & 2014/15 & 701/1.1K & Kinect multi-view & R,D,P & Cls & M/S/-/L & View-invariant depth & ST-GCN \\
\rowcolor{motion} NTU RGB+D 60 \cite{shahroudy2016ntu} & 2016 & 56K clips & 3P & R,D,P,IR & Cls & M/S/-/L & Large-scale skeleton & ST-GCN \\
\rowcolor{motion} InfAR \cite{gao2016infar} & 2016 & 600 clips & 3P & I & Cls & M/S/-/L & Infrared actions & Robustness \\
\rowcolor{motion} Thermal Simulated Fall \cite{vadivelu2016thermal} & 2016 & 44 clips & 3P & I & Cls & L/S/-/L & Fall detection (IR) & Safety \\
\rowcolor{motion} DALY \cite{weinzaepfel2016human} & 2016 & 3.6K ann. & 3P & R & ST+Temp & M/M/-/M & Daily ST actions & Det \\
\rowcolor{motion} MultiTHUMOS \cite{yeung2018every} & 2016 & 400 vids & 3P & R & Temp (dense) & M/M/C/M & Dense multilabel acts & TRF \\
\rowcolor{motion} Volleyball (group
activity) \cite{ibrahim2016hierarchical} & 2016 & 4830 clips & 3P & R & ST (players)+Cls & M/S/-/H & Group activity & GNN, Det \\
\rowcolor{motion} NfS (object tracking) \cite{kiani2017need} & 2017 & 100 vids & 3P & R & Boxes & M/M/-/M & Object tracking & Det \\
\rowcolor{motion} Kinetics-400 \cite{kay2017kinetics} & 2017 & 306K clips & 3P & R & Cls & H/S/-/M & General actions & 3D (I3D), TRF \\
\rowcolor{motion} AudioSet (video) \cite{gemmeke2017audio} & 2017 & 2M clips & 3P & R,A & Weak labels & M/S/-/M & AV tagging/pretrain & AV Fusion \\
\rowcolor{motion} Kinetics-Skeleton \cite{yan2018spatial} & 2018 & 260K & 3P & P & Cls & M/S/-/M & Pose-only actions & ST-GCN \\

\rowcolor{motion} AVA \cite{gu2018ava} & 2018 & 211K ann. & 3P & R,F & ST & L/M/-/H & Atomic ST actions (multi-agent) & Det, TRF \\

\rowcolor{motion} Diving48 \cite{kanojia2019attentive} & 2018 & 18K & 3P & R,F & Cls & L/S/-/L & Fine-grained dives & TRF \\
\rowcolor{motion} Moments in Time \cite{monfort2019moments} & 2019 & 1M+ & 3P & R,A & Cls & M/S/-/M & Event recognition & 3D, TRF \\
\rowcolor{motion} Kinetics-600/700 \cite{kay2017kinetics} & 2018/19 & 496K/650K & 3P & R & Cls & H/S/-/M & Scale for pretraining & 3D, TRF \\
\rowcolor{motion} NTU RGB+D 120 \cite{liu2019ntu} & 2019 & 114K & 3P & R,D,P,IR & Cls & M/S/-/M & Larger skeleton & ST-GCN \\
\rowcolor{motion} FineGym \cite{shao2020finegym} & 2020 & 32K & 3P & R & Cls & L/S/H/L & Fine-grained hierarchy & TRF \\
\rowcolor{motion} VGGSound \cite{chen2020vggsound} & 2020 & 210K clips & 3P & R,A & Cls & M/S/-/M & Audio-visual events & AV Fusion \\
\rowcolor{motion} AVA-ActiveSpeaker \cite{roth2020ava} & 2020 & 3.65M frames & 3P & R,A & ST (speaker) & L/S/-/H & AV diarization & AV Fusion, Det \\

\rowcolor{motion} UAV-Human \cite{li2021uav} & 2021 & 22K clips & 3P (UAV) & R,P & Cls & M/S/-/M & Aerial human acts & Robustness \\
\rowcolor{motion} UCF101-24 \cite{gritsenko2024end} & 2024 & 24 classes & 3P & R & ST & H/S/-/M & ST detection & Det, 3D \\
\rowcolor{motion} EPIC-SOUNDS \cite{huh2025epic} & 2025 & 100h & Ego & A,R & Temp & L/M/C/M & Ego audio events & AV Fusion \\
\rowcolor{motionprocedural} Berkeley MHAD \cite{ofli2013berkeley} & 2013 & 660 clips & 3P & R,D,P,A & Cls & M/S/-/L & Multisensor fusion & ST-GCN \\
\rowcolor{motionprocedural} Something-Something V1/V2 \cite{goyal2017something} & 2017/18 & 108K/221K clips & 3P & R & Cls & L/S/C/M & Object-centric relations & TRN, TRF \\
\rowcolor{procedural} CAD-60 \cite{sung2012unstructured} & 2011 & 68 clips & Kinect single-view & R,D,P & Cls & L/S/-/L & ADL+HOI (depth) & ST-GCN \\
\rowcolor{procedural} GTEA Gaze \cite{fathi2012learning} & 2012 & 17 vids & Ego & R,A (gaze) & Temp & L/M/C/M & Egocentric HOI + gaze & HOI \\
\rowcolor{procedural} CAD-120 \cite{koppula2013learning} & 2013 & 120 clips & Kinect frontal & R,D,P & Temp & L/M/C/M & HOI sequences (procedural) & ST-GCN, TRN \\
\rowcolor{procedural} 50 Salads \cite{stein2013combining} & 2013 & 50 vids & 3P & R & Temp+Step & L/M/H/L & Fine-grained cooking & RNN/TCN \\
\rowcolor{procedural} Breakfast \cite{kuehne2014language} & 2014 & 77h & 3P & R & Temp+Step & L/M/H/L & Procedural activities & RNN/TCN \\
\rowcolor{procedural} GTEA Gaze+ \cite{li2015delving} & 2015 & 37 vids & Ego & R,A (gaze) & Temp & L/M/C/M & Egocentric HOI + gaze & HOI \\
\rowcolor{procedural} SYSU 3D HOI \cite{hu2015jointly} & 2015 & 480 clips & 3P & R,D,P & Cls & L/S/-/M & HOI (depth) & ST-GCN \\
\rowcolor{procedural} EPIC-KITCHENS \cite{damen2018scaling} & 2018 & 39K clips & Ego & R,F & Temp & L/M/C/M & HOI, fine-grained egocentric & HOI, TRF \\
\rowcolor{procedural} YouCook2 \cite{zhou2018towards} & 2018 & 15K segs & 3P & R,T & Temp+Step & L/M/H/M & Cooking segmentation & TRF \\
\rowcolor{procedural} EGTEA Gaze+ \cite{li2018eye} & 2018 & 28 h & Ego & R,A (gaze) & Temp & L/M/C/M & Egocentric HOI + gaze & HOI \\
\rowcolor{procedural} YouCook2-BoundingBox \cite{zhou2018weakly} & 2018 & 15K seg & 3P & R & ST (HOI) & L/M/H/M & Obj-centric cooking & Det, HOI \\
\rowcolor{procedural} COIN \cite{tang2019coin} & 2019 & 12K vids & 3P & R,T & Step & L/M/H/M & Instructional steps & TRF \\
\rowcolor{procedural} CATER \cite{girdhar2019cater} & 2019 & 5.5K & Synth & R & Temp & L/S/C/L & Compositional reasoning & TRF \\
\rowcolor{procedural} CLEVRER \cite{yi2019clevrer} & 2019 & 20K & Synth & R,T & QA & L/S/C/L & Causal reasoning & TRF \\
\rowcolor{procedural} CrossTask \cite{zhukov2019cross} & 2019 & 4.7K & 3P & R,T & Step & L/M/H/M & Weakly sup. steps & TRF \\
\rowcolor{procedural} MOMA \cite{luo2021moma} & 2021 & 2.4K vids & 3P/Ego & R,T & Hier & L/M/H/H & Multi-agent hierarchy & TRF, GNN \\
\rowcolor{procedural} MOMA-LRG \cite{luo2022moma} & 2022 & 148 h & 3P/Ego & R,T & Hier & L/M/H/H & Multi-agent hierarchy & TRF, GNN \\
\rowcolor{temporalprocedural} ADL \cite{pirsiavash2012detecting} & 2009 & 10h & 3P & R & Cls & L/S/-/L & Household ADL & 2S \\
\rowcolor{temporalprocedural} MSRDailyActivity3D \cite{wang2012mining} & 2012 & 320 clips & Kinect frontal & R,D,P & Cls & L/S/-/L & ADL (depth) & ST-GCN \\
\rowcolor{temporalprocedural} MPII Cooking \cite{rohrbach2012database} & 2012 & 3.7K segs & 3P & R & Temp & L/M/C/M & Cooking steps (procedural) & TSN, TRN \\
\rowcolor{temporalprocedural} MPII Cooking 2 \cite{rohrbach2012database} & 2015 & 273 clips & 3P & R & Temp & L/M/C/M & Fine-grained cooking & TRN \\
\rowcolor{temporalprocedural} EPIC-KITCHENS-100 \cite{damen2022rescaling} & 2020 & 90K clips & Ego & R,F,A & Temp+Step & L/M/H/M & HOI + narration & HOI, TRF, VLM \\
\rowcolor{temporal} THUMOS'14 \cite{idrees2017thumos}  & 2014 & 24 classes & 3P & R & Temp & H/M/-/L & Temporal detection & 3D, TSN \\
\rowcolor{temporal} ActivityNet \cite{caba2015activitynet} & 2015 & 28K clips & 3P & R & Temp & M/M/-/M & Temporal localization & 3D, TSN \\
\rowcolor{temporal} Charades \cite{sigurdsson2016hollywood} & 2016 & 66K segs & 3P & R,F & Temp & L/L/C/H & Overlapping indoor actions & TRF, Det \\
\rowcolor{temporal} PKU-MMD I \cite{liu2017pku} & 2017 & 1.1K clips & 3P & R,D,IR,P & Temp & M/M/-/M & Multimodal detection & ST-GCN \\
\rowcolor{temporal} FineAction \cite{yeung2018every} & 2018 & 11.6K clips & 3P & R & Cls & H/S/-/M & Temporal action localization & 2S \\
\rowcolor{temporal} Charades-Ego \cite{sigurdsson2018charades} & 2018 & 68K & Ego+3P & R & Temp & L/M/C/H & Ego/3P alignment & TRF, HOI \\
\rowcolor{temporal} SoccerNet \cite{giancola2018soccernet} & 2018 & 500 games & 3P & R & Temp & H/M/-/M & Sports spotting & TRF \\
\rowcolor{temporal} HACS \cite{zhao2019hacs} & 2019 & 1.5M clips & 3P & R,F & Temp & M/M/-/M & Temporal localization (large) & TRF \\
\rowcolor{temporal} PKU-MMD II \cite{liu2020benchmark} & 2020 & 1K & 3P & R,D,IR,P & Temp & M/M/-/M & Multimodal detection & ST-GCN \\
\rowcolor{temporal} BDD100K (video) \cite{yu2020bdd100k} & 2020 & 100K vids & 3P & R,GPS,IMU & Boxes/Tracks & M/M/-/H & Driving perception/TA & Det, TRF \\
\rowcolor{vlm} MSVD (YouTube2Text) \cite{chen2011collecting} & 2011 & 1.9K vids & 3P & R,T & Cap & L/S/-/M & Captioning & VLM \\
\rowcolor{vlm} MSR-VTT \cite{xu2016msr} & 2016 & 200K pairs & 3P & R,T & Cap & M/S/-/M & Captioning/retrieval & VLM \\
\rowcolor{vlm} LSMDC (Movies) \cite{rohrbach2017movie} & 2016 & 128K sent. & 3P & R,T & Cap & M/S/-/H & Movie-description & VLM \\
\rowcolor{vlm} DiDeMo \cite{anne2017localizing} & 2017 & 10K & 3P & R,T & Grnd & L/S/-/M & Moment grounding & TRF, VLM \\
\rowcolor{vlm} ActivityNet Captions \cite{krishna2017dense} & 2017 & 20K vids & 3P & R,T & Temp+Cap & M/M/C/M & Dense captioning & TRF, VLM \\
\rowcolor{vlm} TGIF-QA \cite{jang2017tgif} & 2017 & 104K QA & 3P & R,T & QA & L/S/-/M & Video QA & TRF \\
\rowcolor{vlm} MSRVTT-QA \cite{xu2017video} & 2017 & 10K vids & 3P & R,T & QA & L/S/-/M & QA & TRF \\
\rowcolor{vlm} Charades-STA \cite{gao2017tall} & 2017 & 9.8K pairs & 3P & R,T & Grnd & L/M/C/H & Language grounding & TRF \\
\rowcolor{vlm} TVQA \cite{lei2018tvqa} & 2018 & 153K QA & 3P & R,T & QA & L/M/-/H & Multimodal QA (subs) & TRF \\
\rowcolor{vlm} VATEX \cite{wang2019vatex} & 2019 & 41K & 3P & R,T & Cap & M/S/-/M & Multilingual captions & VLM \\
\rowcolor{vlm} NExT-QA \cite{xiao2021next} & 2021 & 5.4K vids & 3P & R,T & QA & L/M/C/M & Temporal reasoning QA & TRF \\
\rowcolor{vlm} AGQA \cite{grunde2021agqa} & 2021 & 192M QA & 3P & R,T & QA & L/M/C/M & Compositional QA & TRF \\
\rowcolor{vlm} WebVid-2M/10M \cite{bain2021frozen} & 2021 & 2.5M/10M pairs & 3P & R,T & Cap (weak) & M/L/-/M & VLM pretraining & VLM \\
\rowcolor{vlm} HD-VILA-100M \cite{xue2022advancing} & 2022 & 100M & 3P & R,T,A & Cap (weak) & M/L/-/M & Hi-res pretrain & VLM \\
\rowcolor{vlm} Ego4D \cite{grauman2022ego4d} & 2022 & 3.7K h & Ego & R,A,T & Multi-task (NLQ, AV, gaze) & L/L/H/H & Long-term egocentric & HOI, TRF, VLM \\
\rowcolor{vlm} VidChapters-7M \cite{yang2023vidchapters} & 2023 & 7M seg & 3P & R,T & Temp+Cap & L/M/C/M & Chaptering/summary & TRF, VLM \\
\rowcolor{vlm} InternVid \cite{wang2023internvid} & 2023 & 234M & 3P & R,T & Cap (weak) & M/L/-/M & Video-text pretrain & VLM \\
\rowcolor{vlm} Panda-70M \cite{chen2024panda} & 2024 & 70M & 3P & R,T,A & Cap (weak) & M/L/-/M & Multimodal pretrain & VLM \\
\rowcolor{vlm} MiraData \cite{ju2024miradata} & 2024 & 16K h & 3P & R,T,A & Cap (weak) & M/L/-/M & Multimodal pretrain & VLM \\
\rowcolor{vlm} OpenVid-1M \cite{nan2024openvid} & 2025 & 1M vids & 3P & R,T & Cap (weak) & M/L/-/M & VLM pretraining & VLM \\
\rowcolor{vlm} OpenVidHD-0.4M \cite{nan2024openvid} & 2025 & 433K vids & 3P & R,T & Cap (weak) & M/L/-/M & VLM pretraining & VLM \\
\rowcolor{vlm} Koala-36M \cite{wang2025koala} & 2025 & 36M & 3P & R,T,A & Cap (weak) & M/L/-/M & Multimodal pretrain & VLM \\
\rowcolor{motionvlm} AVA-Kinetics \cite{li2020ava} & 2020 & 230K ann. & 3P & R & ST & M/M/-/H & Large ST localization & TRF, Det \\
\rowcolor{proceduralvlm} TACOS \cite{regneri2013grounding} & 2013 & 127 vids & 3P & R,T & Grnd & L/M/C/M & Grounding in cooking & TRF, VLM \\
\rowcolor{proceduralvlm} HowTo100M \cite{miech2019howto100m} & 2019 & 136M clips & 3P & R,A,T & Weak Align & L/L/H/M & Instructional pretraining & VLM, TRF \\
\bottomrule
\end{tabular}}
\label{tab:datasets_structural_lenses}
\end{center}
\vspace{-4pt}
\footnotesize \textbf{Reading guide.} \emph{Anno} shows what the dataset supervises (\eg, procedural steps, spatiotemporal boxes, captions/QA). \emph{Struct} summarizes motion, temporal, and relational aspects. \emph{Impulse} highlights modeling advances spurred by the dataset (\eg, Kinetics$\rightarrow$I3D/Transformers; Something-Something$\rightarrow$relation reasoning; AVA/Volleyball$\rightarrow$ detection/graph models; EPIC/Ego4D$\rightarrow$ HOI/egocentric transformers; HowTo100M/WebVid$\rightarrow$ vision-language pretraining).
\end{table*}

\section{Datasets as Structural Drivers}
\label{sec:dataset}

Datasets are commonly viewed as benchmarks for evaluating competing algorithms. In this survey, we adopt a different perspective: datasets are active drivers of innovation that shape what video models learn, how they represent information, and ultimately how architectural paradigms evolve. Rather than merely defining task difficulty, dataset characteristics determine the learning requirements that models should satisfy. 
These requirements influence not only model performance but also the representations that models acquire during training.

A key observation underlying this survey is that different datasets induce different representational biases, even when the underlying architecture remains unchanged. Models trained on large-scale action recognition datasets often prioritize appearance, scene context, and short-term motion cues, whereas models trained on motion-centric datasets become more sensitive to object trajectories and temporal ordering. Egocentric and multimodal datasets further encourage representations centered on human-object interactions, semantic grounding, and cross-modal correspondence. In this sense, datasets leave identifiable representational fingerprints on learned models, shaping which cues are considered most informative for prediction.
A detailed discussion of how different dataset characteristics shape learning requirements and architectural design can be found in Appendix B.

This perspective suggests that progress in video understanding is driven not solely by larger models or improved optimization, but by an ongoing interaction between dataset evolution and representation learning. As datasets become increasingly rich in temporal, relational, and multimodal structure, they redefine the learning requirements imposed on video models and reshape the representations that successful models should acquire. Architectural evolution can therefore be interpreted as a response to these changing representational demands rather than as a sequence of independent innovations.

\subsection{A Dataset-Bias-Architecture Framework}

Building on the preceding discussion, we formalize the relationship between datasets and model evolution through a \emph{dataset-bias-architecture} framework: dataset characteristics $\rightarrow$ learning requirements $\rightarrow$ representational biases $\rightarrow$ architectural responses.
The framework is motivated by a simple observation: architectures do not emerge in isolation. Instead, they arise in response to the dominant challenges presented by a particular data regime. Dataset characteristics define the information required for successful prediction, thereby shaping the learning requirements. 
During training, models develop representational biases toward the cues that are most informative for these requirements. Architectural innovations can then be interpreted as mechanisms that exploit such biases more effectively and at larger scales.
This perspective helps explain why different datasets often favor different representations despite sharing similar tasks. For example, Something-Something V2 emphasizes object manipulations and temporal state transitions, encouraging motion-centric representations that are sensitive to temporal ordering. EPIC-KITCHENS prioritizes hand-object interactions and object affordances, inducing interaction-centric representations. In contrast, multimodal corpora such as HowTo100M, WebVid, and InternVid emphasize semantic grounding and cross-modal correspondence, encouraging representations that align visual content with language. Although all of these datasets contribute to video understanding, they impose fundamentally different learning requirements and therefore promote different representational biases.

A key implication is that architectural success depends on the alignment between dataset characteristics and model inductive biases. Architectures that encode assumptions consistent with the dominant signals in the data are more likely to learn effective representations, whereas architectures optimized for one data regime may transfer poorly to another when the underlying learning requirements differ. This provides a principled explanation for several recurring trends in the literature, including why appearance-centric models perform strongly on large-scale action recognition benchmarks, why motion-centric datasets stimulate temporal reasoning architectures, and why multimodal datasets accelerate the development of vision-language models and video foundation models.
Across the literature, five recurring dataset characteristics repeatedly reshape the learning requirements imposed on video understanding systems:
\renewcommand{\labelenumi}{\roman{enumi}.}
\begin{enumerate}[leftmargin=0.5cm]
\item Motion complexity: actions are distinguished primarily through temporal dynamics and object state transitions rather than static appearance.
\item Temporal span: relevant information is distributed across extended temporal horizons and long-range dependencies.
\item Compositional structure: activities are organized hierarchically into actions, sub-actions, and procedural steps.
\item Relational interactions: understanding depends on interactions among agents, objects, and their surrounding context.
\item Multimodal richness: visual information should be integrated with language, audio, or other modalities.
\end{enumerate}

These characteristics constitute the principal dimensions along which video datasets evolve. Table \ref{tab:datasets_structural_lenses} summarizes the most popular datasets from the past 20 years, with row colors indicating their primary categories and overlaps. Early benchmarks were dominated by motion-focused datasets, while procedural, temporal, and video-language datasets have grown steadily, reflecting the community's increasing interest in fine-grained actions, stepwise reasoning, and multimodal understanding. Importantly, they also correspond to successive shifts in the dominant bottlenecks of video understanding, from motion recognition and temporal reasoning to relational understanding and multimodal grounding. Throughout the remainder of this survey, we use these dimensions as a unifying lens for interpreting dataset evolution and representation learning. 

\textbf{Operational definitions.} The structural descriptors reported in Table \ref{tab:datasets_structural_lenses} are intended as comparative characterizations of dataset properties rather than strict quantitative measurements. Temporal span (Span) reflects the temporal horizon over which relevant information should be integrated, ranging from short clip-level actions to long-duration activities involving extended dependencies. Motion amplitude (Amp) characterizes the extent to which discriminative information is conveyed through visible motion, ranging from subtle object-state changes to large body, object, or camera movements. Compositionality (Comp) describes whether activities can be decomposed into constituent actions or procedural steps, with hierarchical datasets additionally providing explicit multi-level activity structures. Agent density (Agents) reflects the prevalence and importance of interactions among multiple humans, objects, or agents within a scene. The labels in Table \ref{tab:datasets_structural_lenses} were assigned through a combination of benchmark documentation, annotation protocols, representative examples reported in the original dataset papers, and qualitative inspection of dataset characteristics. Accordingly, these labels should be interpreted as coarse-grained descriptors that facilitate cross-dataset comparison rather than precise statistical measurements. 
Future work could formalize these dimensions through quantitative statistics such as average video duration, optical-flow magnitude distributions, hierarchical annotation depth, and interaction frequency.

\begin{figure*}[tbp]
\vspace{-1.0cm}
\centering
\subfloat[TimeSformer.]{\label{fig:timesformer}
\includegraphics[trim=0cm 0cm 0cm 0cm,clip=true, width=0.48\linewidth]{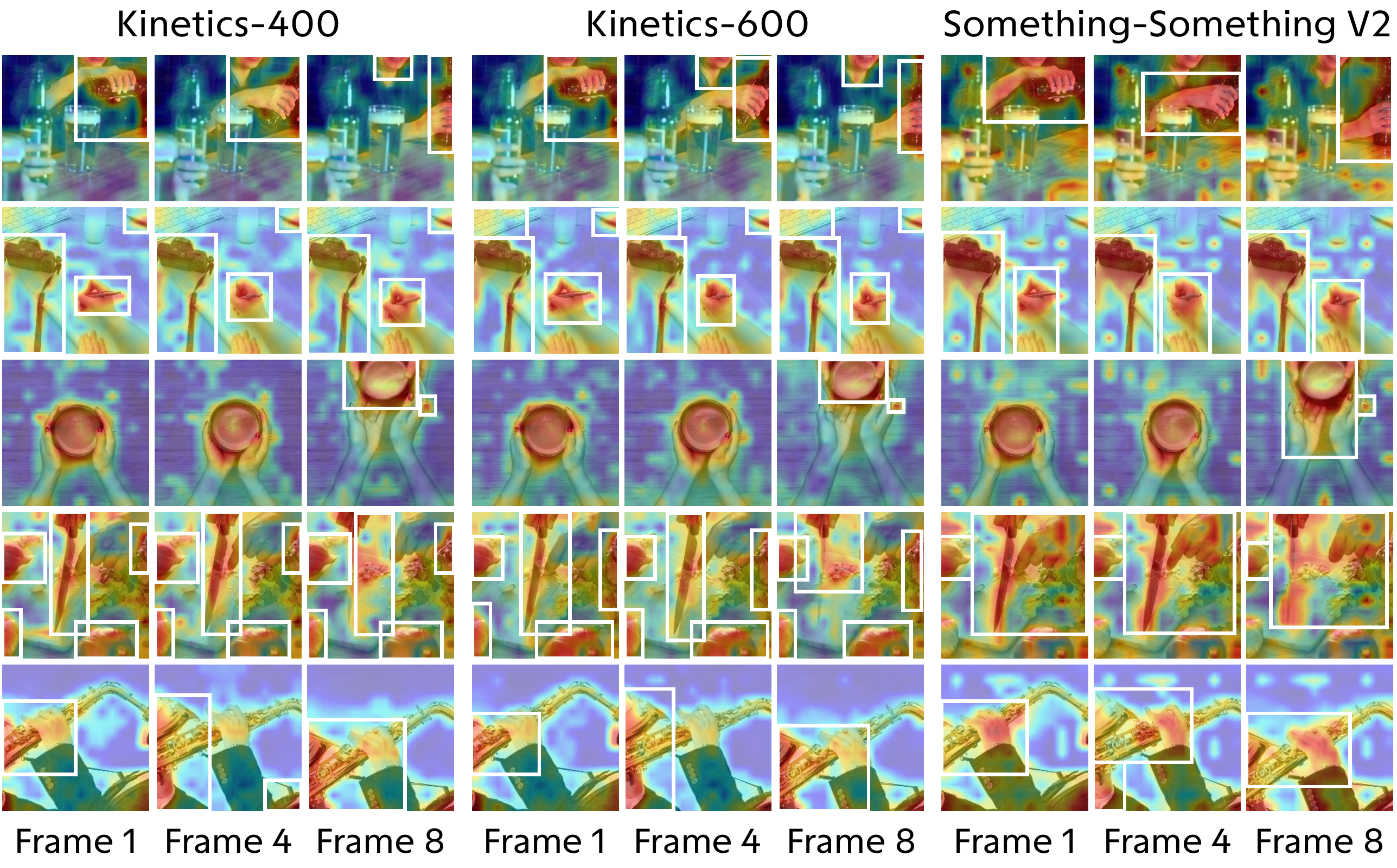}} \hfill
\subfloat[ViViT.]{\label{fig:vivit}
\includegraphics[trim=0cm 0cm 0cm 0cm,clip=true, width=0.472\linewidth]{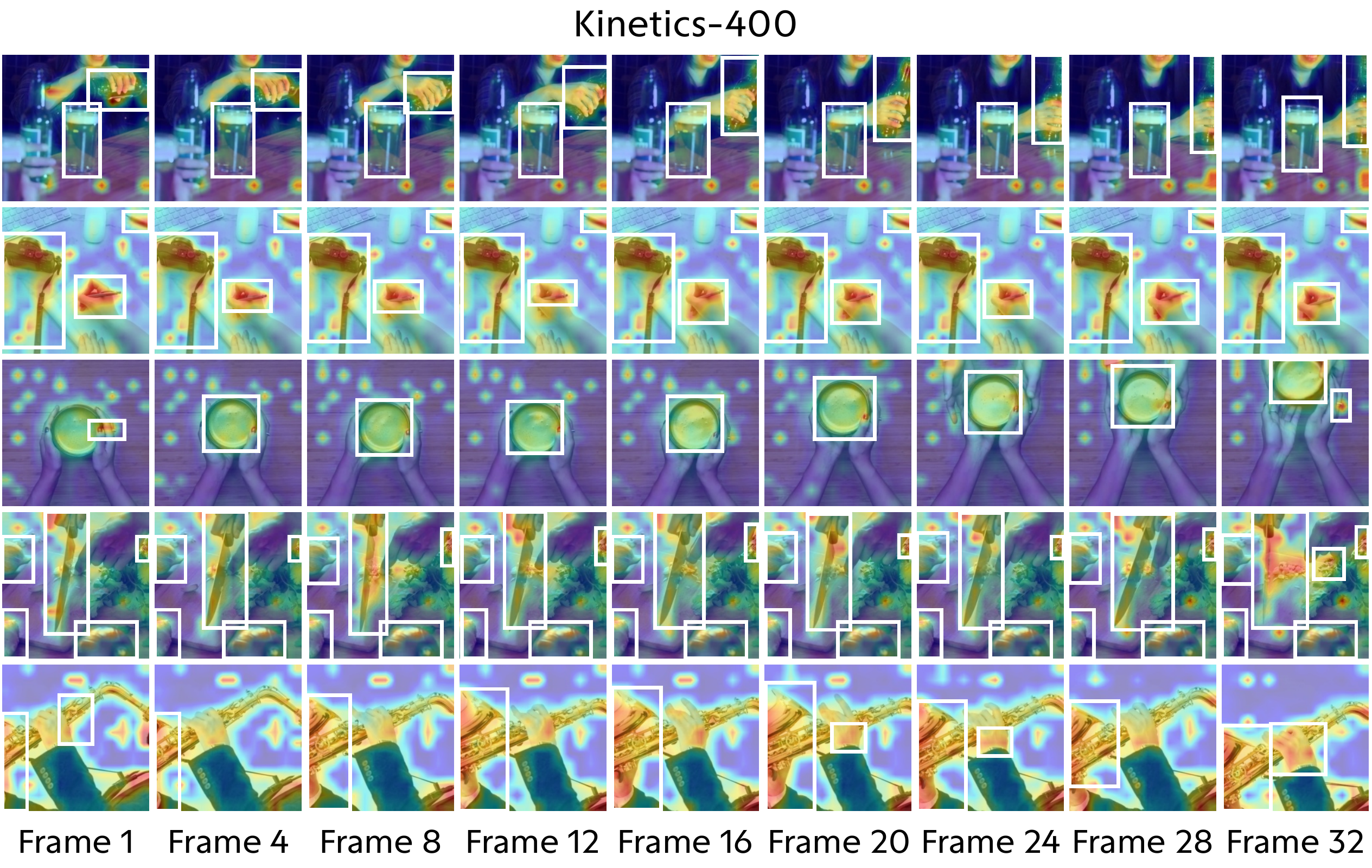}}\\
\vspace{-0.3cm}
\subfloat[Motionformer.]{\label{fig:motionformer}
\includegraphics[trim=0cm 0cm 0cm 0cm,clip=true, width=0.48\linewidth]{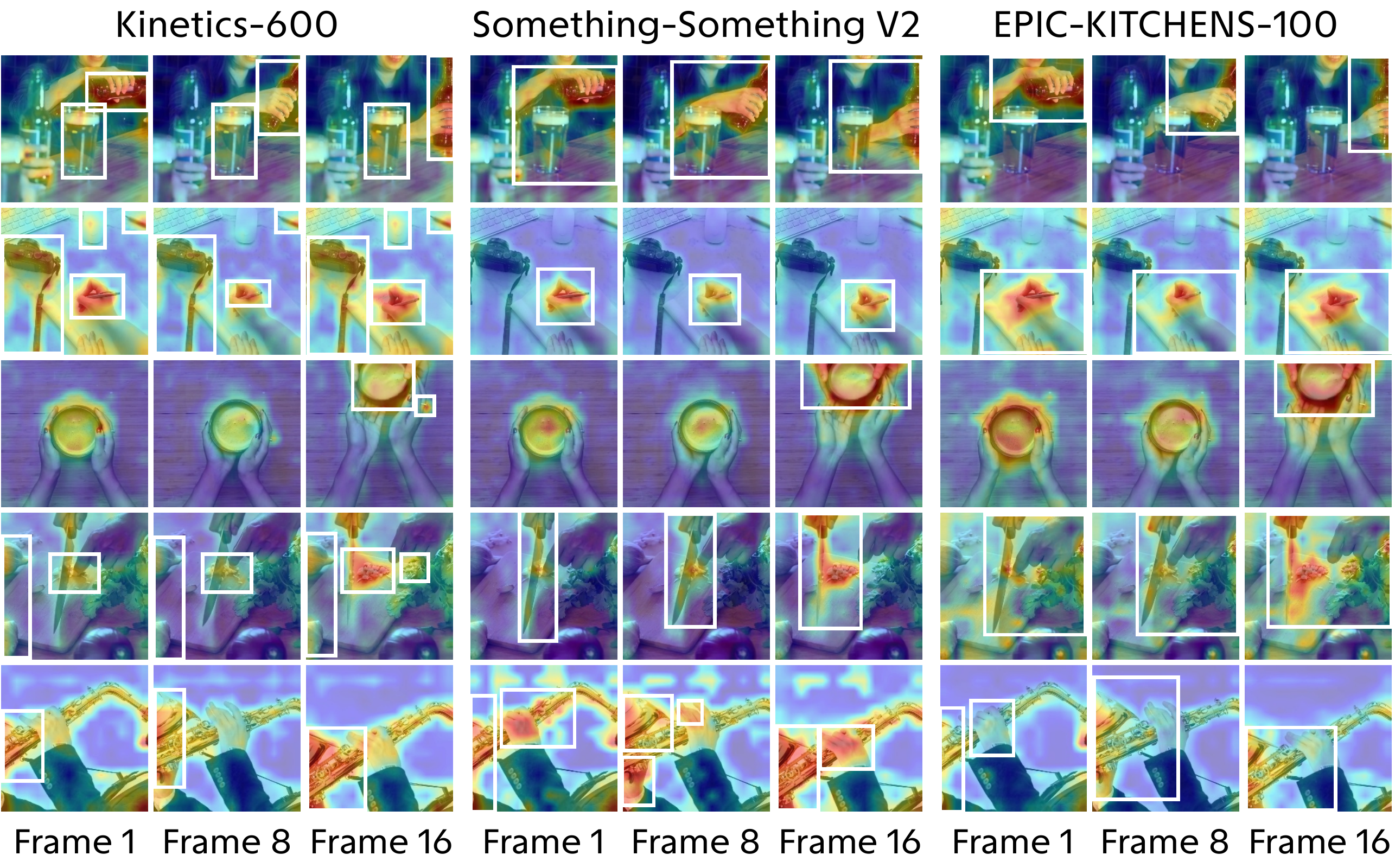}} \hfill
\subfloat[X-CLIP.]{\label{fig:xclip}
\includegraphics[trim=0cm 0cm 0cm 0cm,clip=true, width=0.415\linewidth]{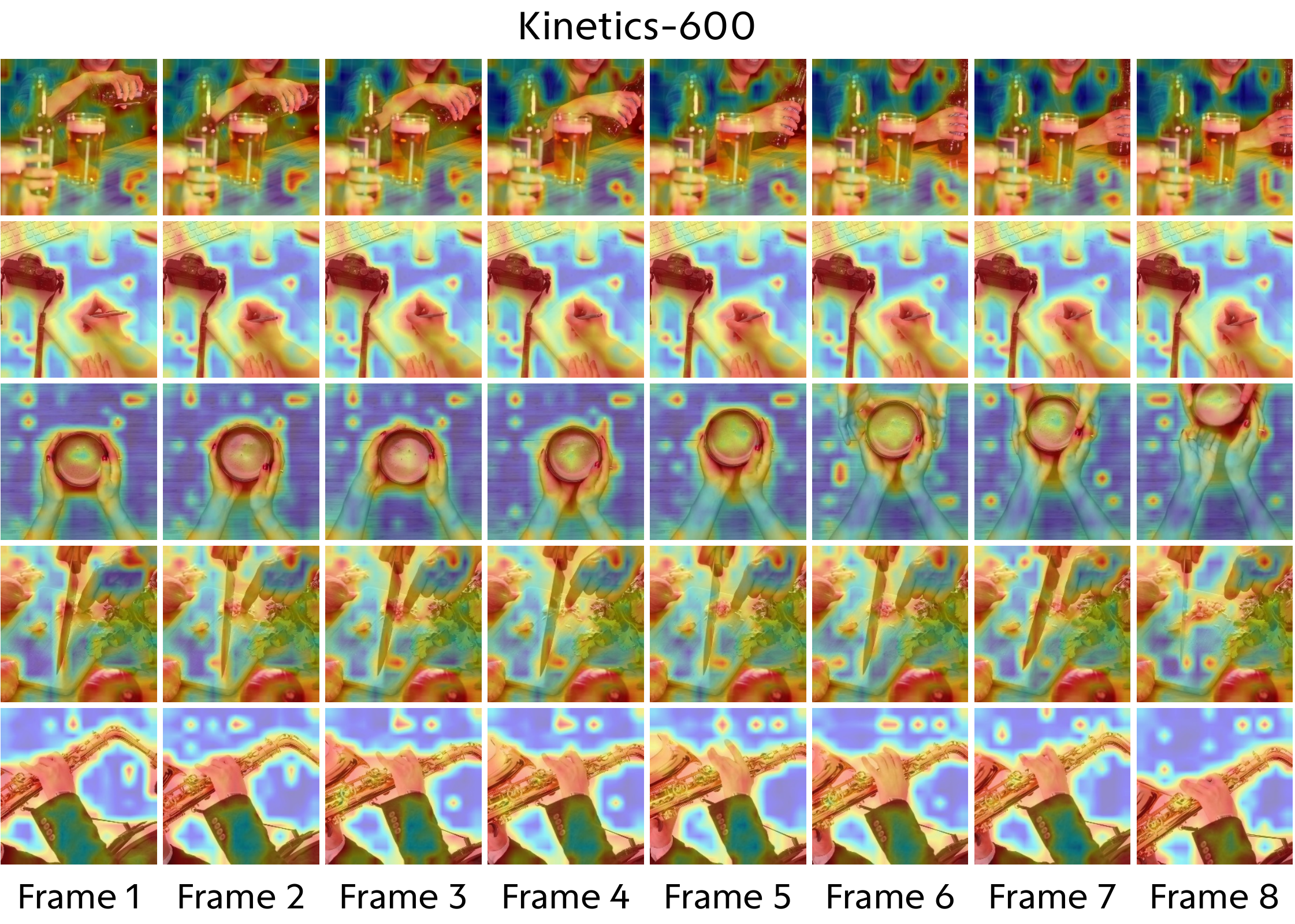}}\\
\vspace{-0.3cm}
\subfloat[VideoMAE.]{\label{fig:videomae}
\includegraphics[trim=0cm 0cm 0cm 0cm,clip=true, width=\linewidth]{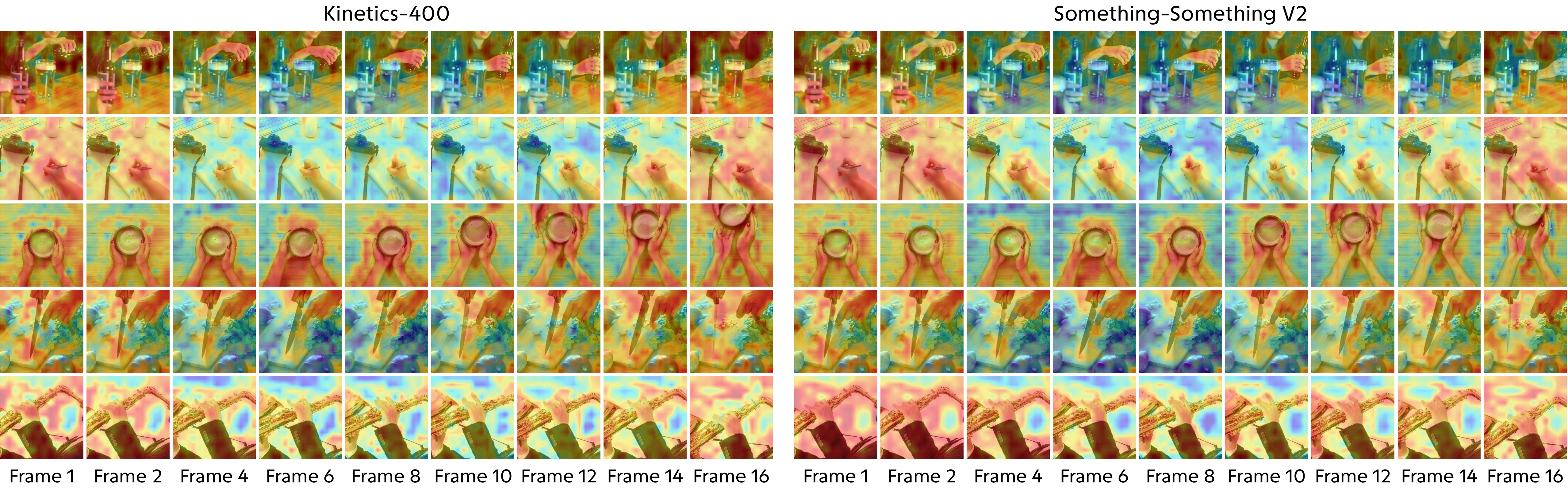}}\\
\vspace{-0.3cm}
\subfloat[VideoMAE V2.]{\label{fig:videomaev2}
\includegraphics[trim=0cm 0cm 0cm 0cm,clip=true, width=0.48\linewidth]{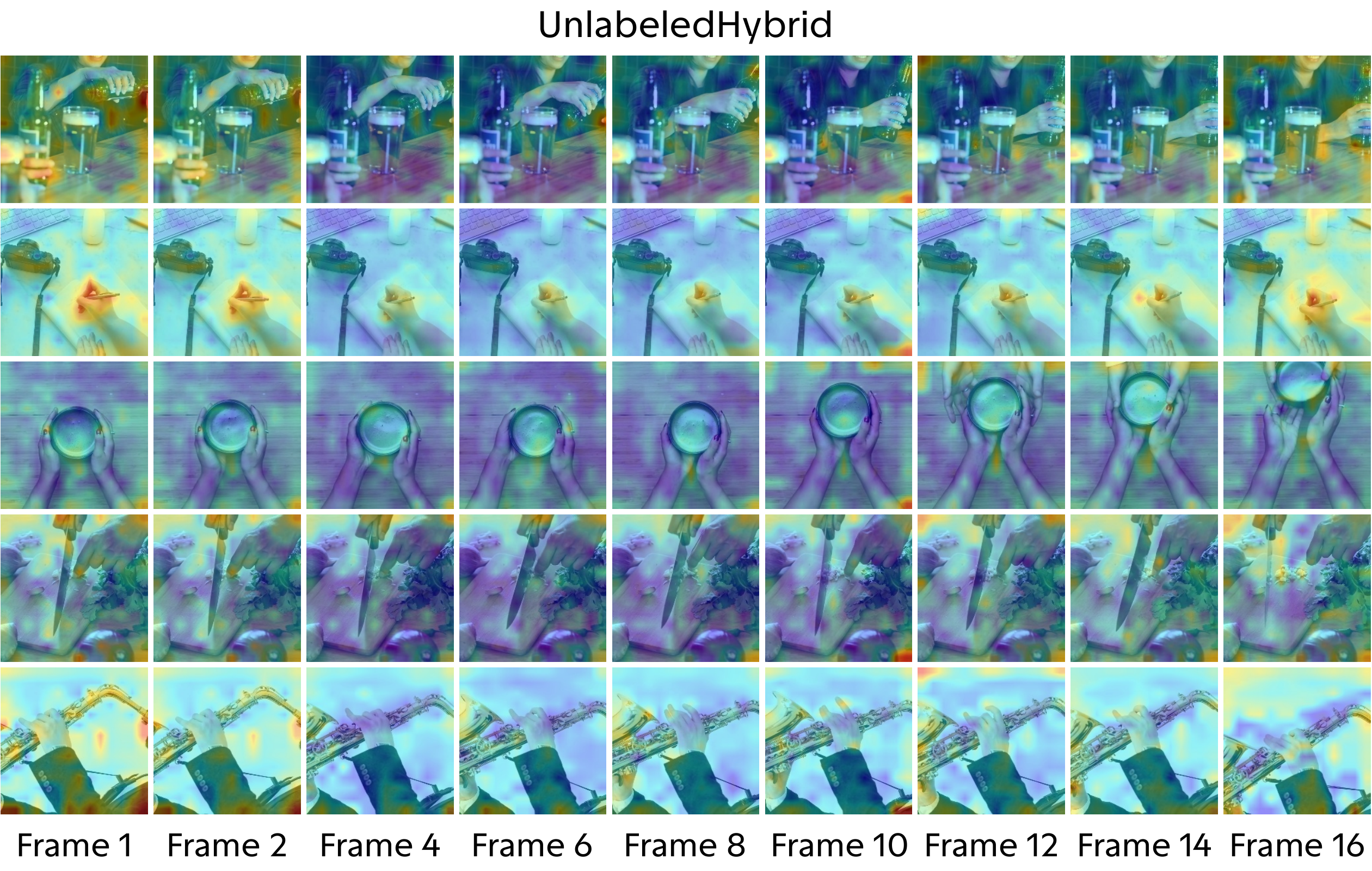}} \hfill
\subfloat[V-JEPA 2.]{\label{fig:vjepa2}
\includegraphics[trim=0cm 0cm 0cm 0cm,clip=true, width=0.48\linewidth]{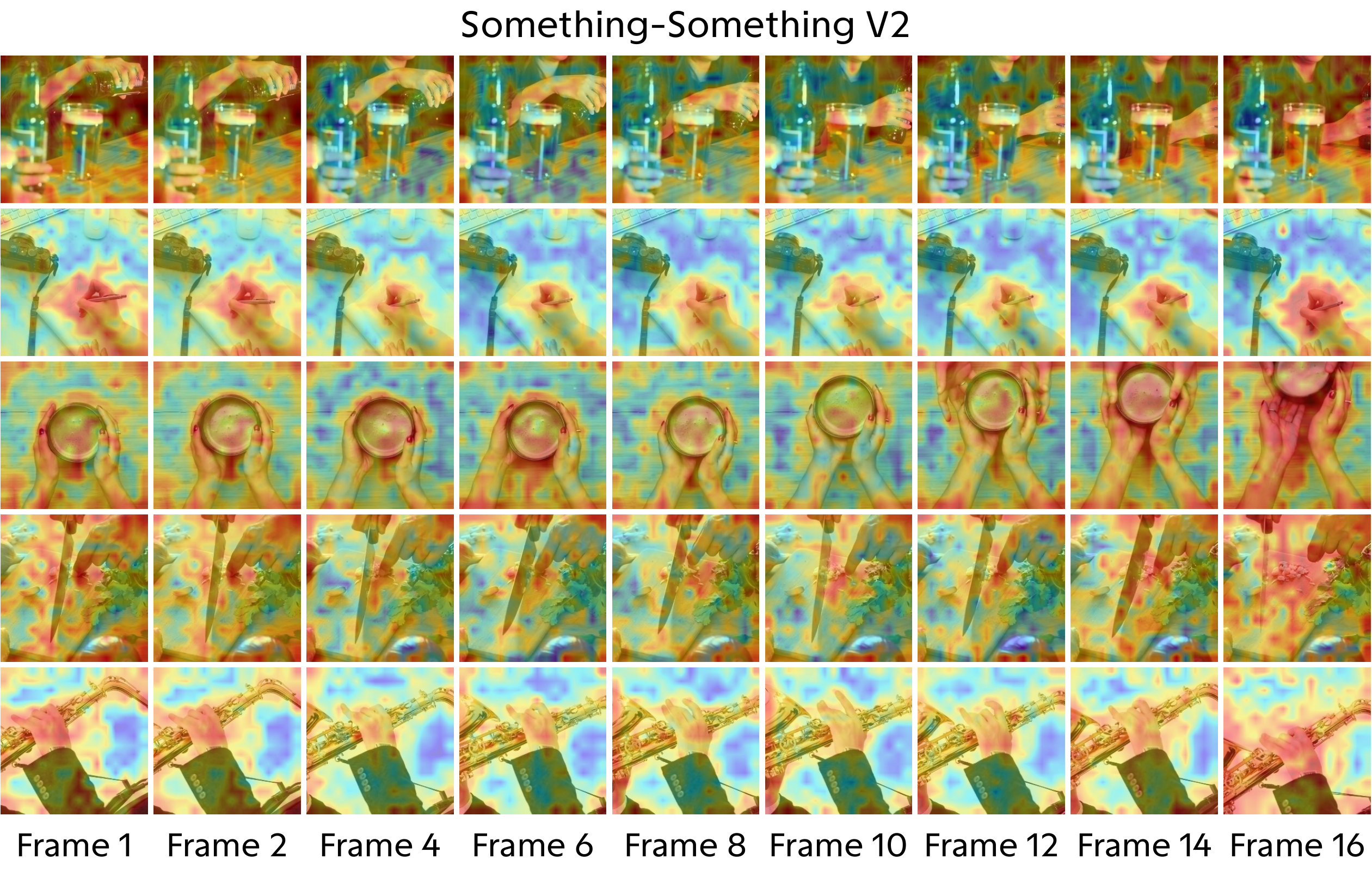}}\\
\vspace{-0.2cm}
\caption{
Attention visualizations of representative video models pretrained on different datasets and evaluated on the same video. 
Dataset-specific attention patterns are consistently observed, although their strength varies across architectures.
}
\label{fig:architectures}
\end{figure*}


\subsection{Dataset-Induced Representational Biases}

A central implication of the proposed framework is that datasets influence not only model performance but also the representations that models learn. To visualize dataset-induced representational biases, we use videos from OWM~\cite{baumann2026envisioning}, which contain rich object manipulations, temporal state transitions, and human-object interactions within a unified setting. Such videos simultaneously expose context, motion, and interaction cues, making them particularly suitable for probing how different pretraining datasets influence learned representations. The selected pretraining datasets are among the most widely adopted benchmarks in video understanding and capture diverse learning requirements, including scene context, temporal dynamics, fine-grained motion, and human-object interaction. Because publicly released checkpoints are not uniformly available across architectures and datasets, the visualizations are intended as qualitative evidence of dataset-induced biases rather than exhaustive architecture-dataset comparisons. By evaluating models on identical inputs that contain multiple reasoning signals, differences in attention patterns can be more directly attributed to dataset-induced representational biases rather than variations in scene content.

\textbf{Dataset biases across architectural families.} Fig. \ref{fig:architectures} 
reveals two important observations. First, dataset-induced biases generalize across diverse architectures. Across TimeSFormer~\cite{bertasius2021space}, ViViT~\cite{arnab2021vivit}, MotionFormer~\cite{patrick2021keeping}, VideoMAE~\cite{tong2022videomae}, VideoMAE V2~\cite{wang2023videomae}, and V-JEPA 2~\cite{assran2025v}, pretraining on different datasets consistently produces different attention patterns when processing the same video. Kinetics-based pretraining generally promotes context-oriented representations, Something-Something emphasizes manipulated objects and motion dynamics, and EPIC-KITCHENS focuses on hand-object interactions. These trends closely mirror the historical evolution of video datasets from action recognition toward temporal reasoning and interaction understanding.
Second, the strength of these biases is architecture-dependent. TimeSFormer and MotionFormer exhibit particularly clear transitions between context-, motion-, and interaction-centric representations, whereas VideoMAE, VideoMAE V2, and V-JEPA 2 produce more diffuse attention patterns with weaker separation among datasets. This suggests that dataset characteristics alone do not determine the resulting representations; rather, architectural design and pretraining objectives mediate how strongly dataset-specific learning requirements are reflected in the learned features. These observations provide additional evidence that architectural evolution is closely coupled with dataset evolution, while also highlighting that different architectural paradigms respond differently to the representational pressures imposed by successive generations of video datasets.

\begin{figure*}[tbp]
\centering
\includegraphics[trim=0 0 0 0, clip=true, width=\linewidth]{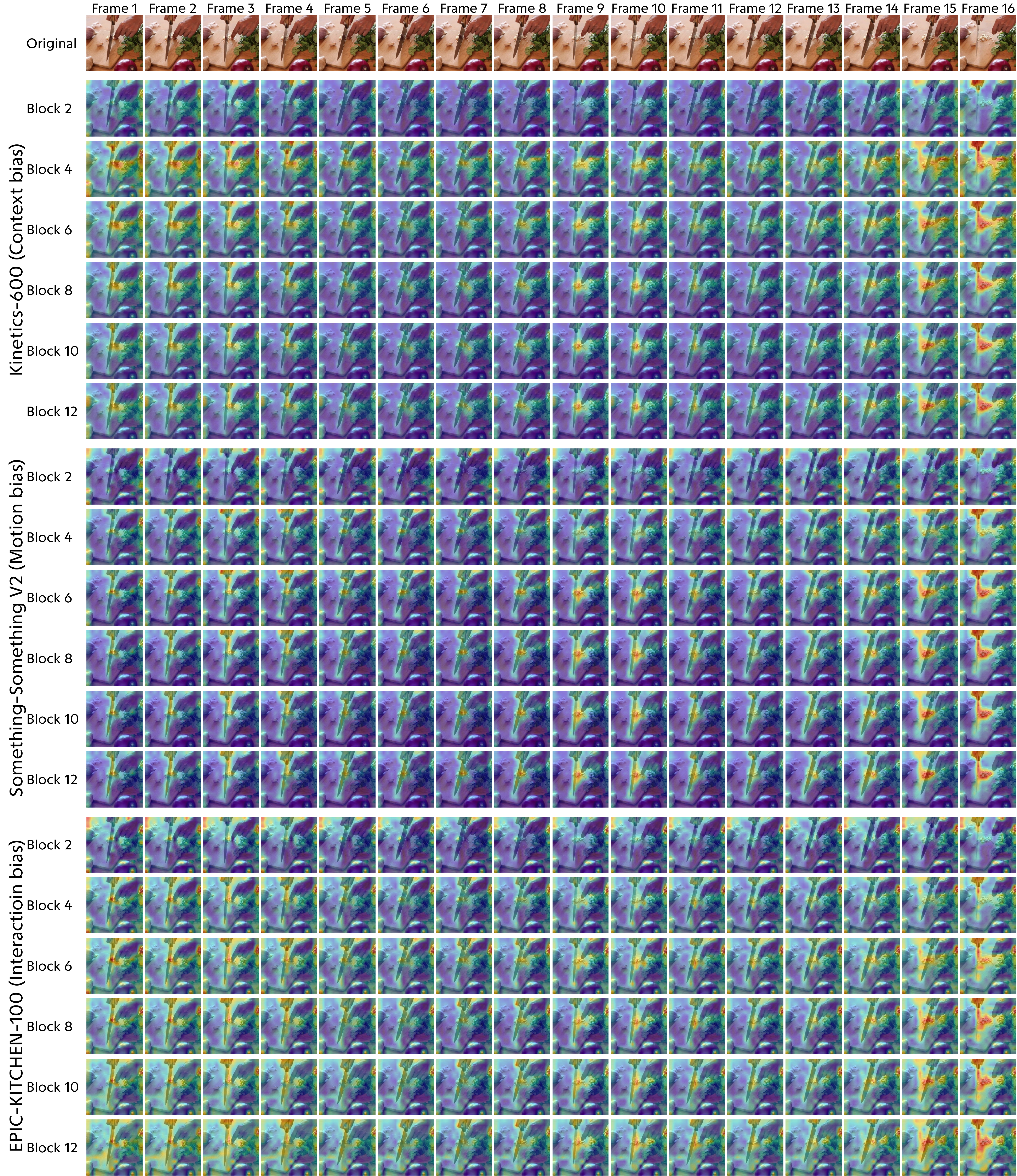}
\caption{
Dataset-induced biases across network depth. DAAM visualizations from MotionFormer blocks 2, 4, 6, 8, 10, and 12 pretrained on Kinetics-600, Something-Something V2, and EPIC-KITCHENS-100. Distinct context-, motion-, and interaction-centric attention patterns emerge early and remain visible across network depth, suggesting that dataset characteristics shape feature representations throughout the model hierarchy rather than only the final prediction.
}
\label{fig:motionformer-perblock}
\end{figure*}

\begin{figure*}[tbp]
\centering
\includegraphics[trim=0 0 0 0, clip=true, width=\linewidth]{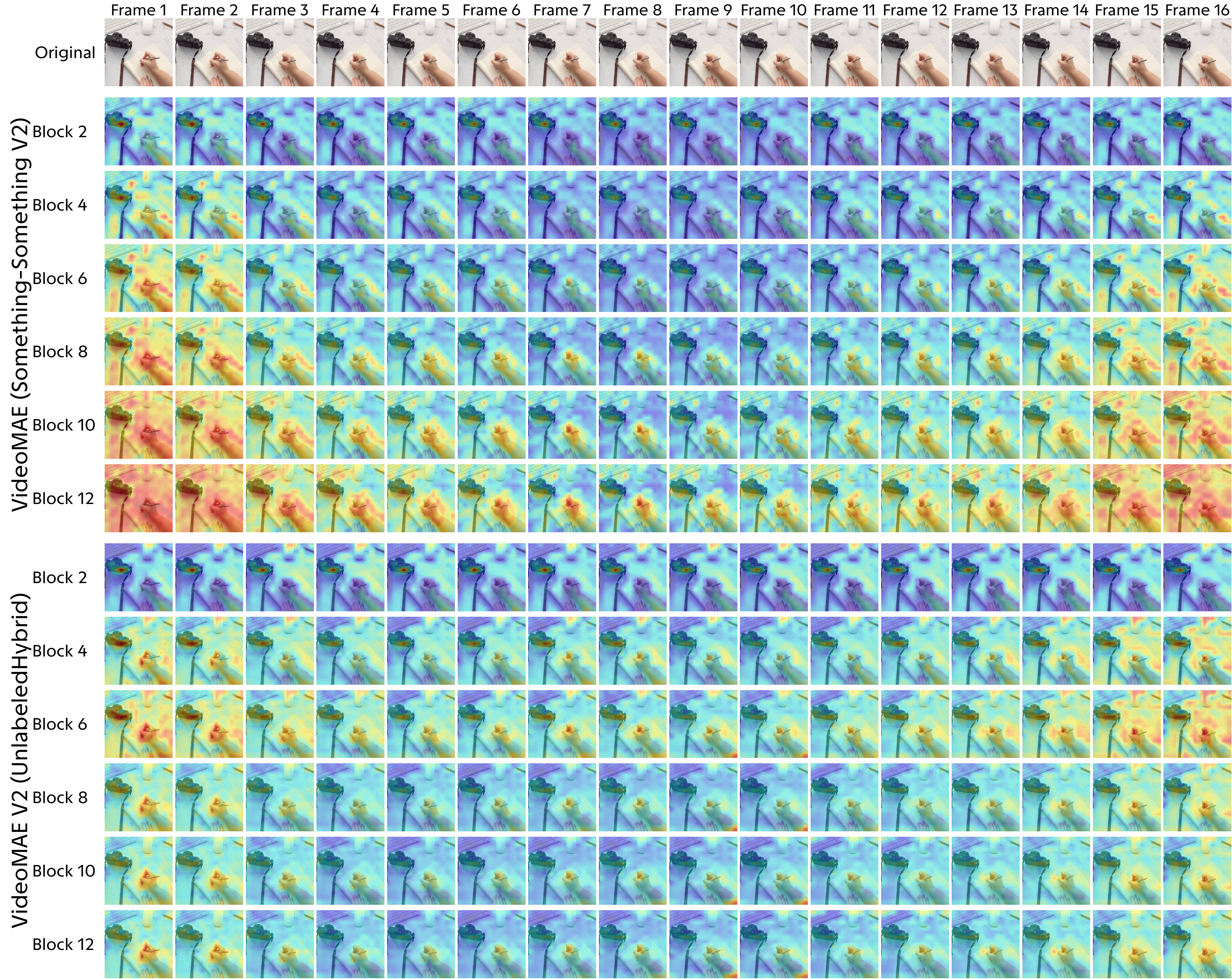}
\caption{
Dataset-induced biases across network depth. DAAM visualizations from VideoMAE blocks 2, 4, 6, 8, 10, and 12 pretrained on Something-Something V2 and UnlabeledHybrid. Differences in attention patterns become more pronounced in deeper layers, suggesting that dataset characteristics progressively shape learned representations throughout the model hierarchy.
}
\label{fig:videomae-perblock}
\end{figure*}

\begin{figure*}[tbp]
\centering
\includegraphics[trim=0 0 0 0, clip=true, width=\linewidth]{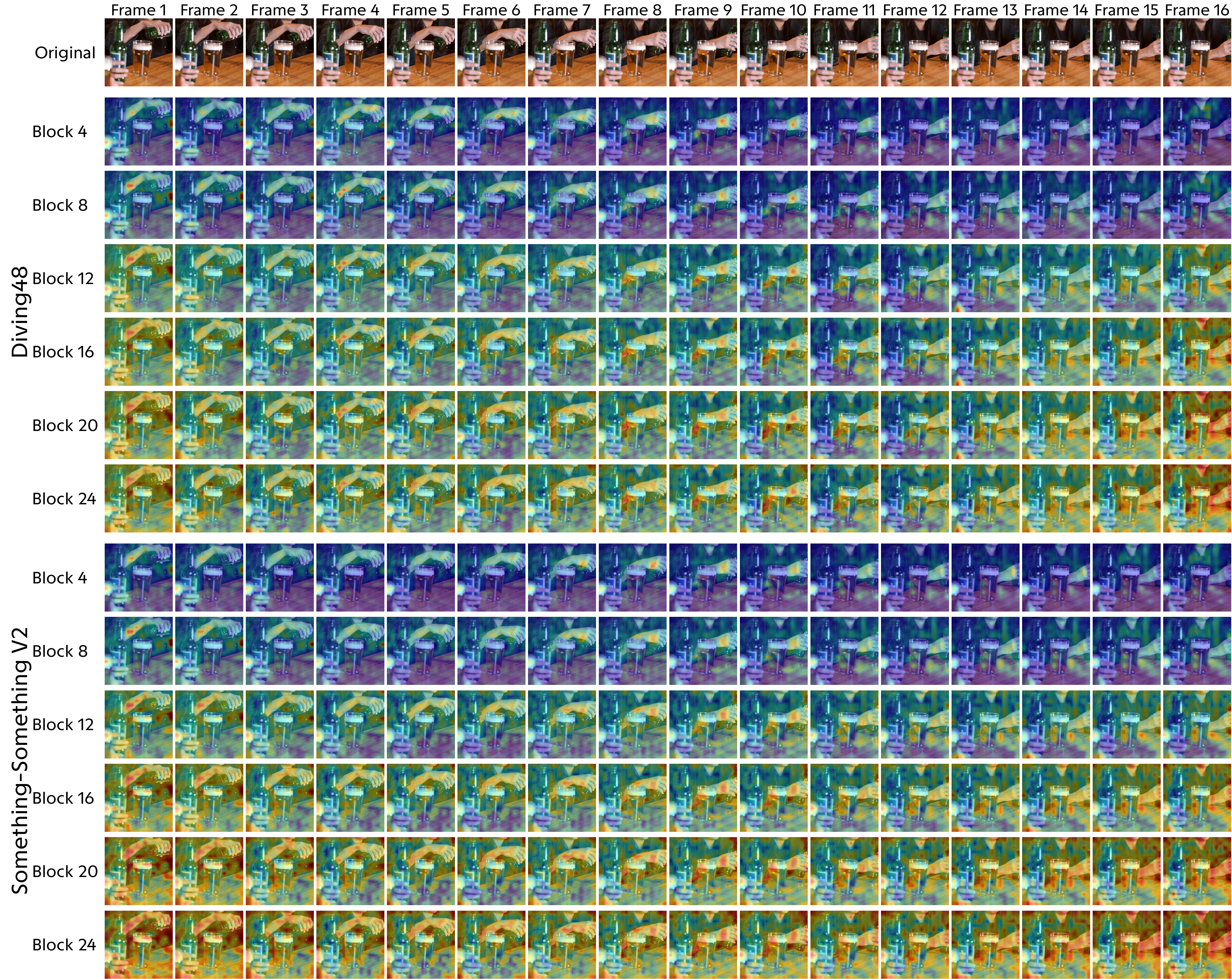}
\caption{
Dataset-induced biases across network depth.
DAAM visualizations from V-JEPA 2 blocks 4, 8, 12, 16, 20, and 24 pretrained on Diving48 and Something-Something V2. The relatively similar attention patterns across pretraining datasets suggest that large-scale self-supervised learning may reduce dataset-specific specialization, highlighting the joint influence of datasets, architectures, and learning objectives on learned representations.
}
\label{fig:vjepa-perblock}
\end{figure*}

\textbf{Dataset-induced biases across network depth.} Fig. \ref{fig:motionformer-perblock} extends the observations of Fig.~\ref{fig:data-driven} by visualizing attention maps across MotionFormer blocks pretrained on Kinetics-600, Something-Something V2, and EPIC-KITCHENS-100. A notable observation is that dataset-specific biases are visible throughout the network hierarchy rather than emerging only in the final layers. Despite sharing an identical architecture, the three models exhibit distinct attention patterns from the early blocks onward: Kinetics-600 consistently allocates attention to surrounding objects and scene context, Something-Something V2 emphasizes motion trajectories and temporal dynamics, and EPIC-KITCHENS-100 concentrates on hand-object interaction regions.
More importantly, these differences become progressively refined across depth rather than disappearing. The separation among the three pretrained models is particularly evident in the intermediate layers, suggesting that dataset characteristics influence not only the final representation but also the hierarchical organization of features learned throughout the network. This observation supports a central premise of the proposed dataset-bias-architecture framework: datasets do not merely affect model outputs; they shape the representational biases learned across the entire feature hierarchy. Consequently, architectural evolution can be viewed not only as a search for more expressive models, but also as a response to the representational pressures imposed by different dataset regimes.
Figure~\ref{fig:videomae-perblock} shows that VideoMAE pretrained on Something-Something V2 develops more localized attention around the hand-object interaction, whereas VideoMAE V2 pretrained on UnlabeledHybrid exhibits broader attention patterns. The separation becomes more evident in deeper layers, indicating that dataset characteristics continue to shape learned representations under self-supervised learning.
Unlike MotionFormer, V-JEPA 2 exhibits weaker separation between datasets, with both pretrained models focusing on similar action-centric regions (Fig.~\ref{fig:vjepa-perblock}). This observation suggests that the manifestation of dataset-induced biases depends not only on dataset characteristics but also on architectural design and pretraining objectives. The relatively 
similar attention patterns indicate that large-scale self-supervised predictive learning may reduce dataset-specific specialization, implying that the influence of datasets on learned representations is mediated by the underlying learning paradigm.

Although the visualizations are qualitative, the observed trends are consistent across multiple architectures
, 
network depths
, and pretraining datasets. The goal is not to establish a new quantitative bias metric, but to provide interpretable evidence that complements the broader historical and benchmark analyses presented throughout this survey. 

The implications extend beyond attention visualization. Dataset characteristics determine the dominant learning requirements of a task, which subsequently influence the representations that successful models acquire. Motion-centric datasets encourage sensitivity to temporal dynamics and state transitions; interaction-centric datasets promote representations focused on agents, objects, and their relationships; multimodal datasets encourage semantic grounding and cross-modal correspondence. As these representational biases become increasingly important, new architectural mechanisms emerge to exploit them more effectively.
This perspective provides a useful bridge between datasets and architectures. Rather than viewing architectural innovation as an independent process, we argue that changes in dataset design first alter the representations required for success, which subsequently create demand for new architectural capabilities. In this sense, representational evolution precedes architectural evolution. Understanding dataset-induced representational biases therefore provides insight into why particular architectural paradigms become effective, why certain models generalize better across data regimes, and why major shifts in video understanding often coincide with the emergence of new benchmark datasets.

\subsection{From Dataset Evolution to Architectural Evolution}

Rather than treating representative benchmarks as isolated evaluation resources, we interpret them as successive shifts in the dominant learning requirements imposed on video models. 

Through this lens, architectural evolution emerges not as a sequence of independent innovations, but as a response to increasingly demanding dataset characteristics and the representational biases they induce.
Several trends emerge from Table \ref{tab:datasets_structural_lenses}. First, the dominant challenge of video understanding has shifted from motion recognition toward temporal reasoning, relational understanding, and multimodal grounding. Early datasets such as KTH, UCF101, and HMDB51 are characterized by high motion saliency but limited temporal span and compositional structure. In contrast, datasets such as Breakfast, ActivityNet, EPIC-KITCHENS, and Ego4D increasingly emphasize long-horizon activities, procedural dependencies, and human-object interactions. More recent multimodal corpora, including HowTo100M, WebVid, InternVid, and Panda, introduce large-scale video-language alignment as a primary learning objective. Architectural trends broadly mirror these shifts in learning requirements, suggesting that changes in dataset structure systematically reshape the capabilities required for successful video understanding.
These trends reveal a consistent relationship
between dataset evolution and architectural innovation.
As dataset characteristics evolve from short-term motion
recognition to long-horizon reasoning, relational interactions,
and multimodal grounding, the dominant bottlenecks of video
understanding shift accordingly. Architectural advances can
therefore be interpreted as responses to changing learning
requirements rather than isolated algorithmic innovations.

\textbf{Role of datasets relative to compute and scaling.} It is important to note that architectural evolution is influenced by multiple factors, including computational resources, optimization techniques, self-supervised learning, and scaling laws. Our dataset-bias-architecture framework does not claim that datasets are the sole driver of progress. Rather, it argues that datasets determine the dominant learning requirements that successful models should satisfy. Advances in compute and optimization influence how effectively these requirements can be addressed, whereas datasets influence which requirements become important in the first place. The emergence of temporal reasoning architectures, relational models, and multimodal foundation models consistently coincides with the introduction of datasets that expose corresponding challenges, suggesting that dataset evolution acts as a major source of architectural pressure throughout the history of video understanding.

\begin{table*}[tbp]
\caption{Top-1 performance of representative video models on action recognition and detection datasets. 
For each dataset, the best-performing model variant is reported. Datasets: HMDB51 and UCF101, action classification (top-1 accuracy); Sports-1M, with Clip Hit@1 (C@1), Video Hit@1 (V@1), and Video Hit@5 (V@5) metrics; AVA v2.1 and v2.2, action detection (mAP); Diving48, Moments in Time (Moments), Kinetics-400/600/700 (K400/K600/K700), Something-Something v1/v2 (SSv1/SSv2), and ActivityNet (ANet), top-1 classification accuracy; Charades, action detection (mAP); EPIC-KITCHENS-100, top-1 accuracy reported separately for Action, Verb, and Noun.
Model abbreviations: 2S indicates Two-Stream, F-ST-ConvNet indicates Factorized Spatiotemporal ConvNet, and NL indicates Non-Local. 
A dash (-) indicates results not reported for a dataset.
Early two-stream and 3D CNNs excel on short clips, sequential models (TSN, TRN, TSM, SlowFast) capture long-range and compositional actions, and transformer-based models dominate procedural, multi-agent, and relational tasks.
}
\centering
\setlength{\tabcolsep}{0.15em}
\renewcommand{\arraystretch}{0.70}
\resizebox{\linewidth}{!}{%
\begin{tabular}{lcccccccccccccccccccc}
\toprule
\multirow{2}{*}{\textbf{Model}} & \multirow{2}{*}{\textbf{HMDB51}} & \multirow{2}{*}{\textbf{UCF101}} & \multicolumn{3}{c}{\textbf{Sports-1M}} & \multirow{2}{*}{\textbf{AVA}\tiny{v2.1}} & \multirow{2}{*}{\textbf{AVA}\tiny{v2.2}} & \multirow{2}{*}{\textbf{Diving48}} & \multirow{2}{*}{\textbf{Moments}} & \multirow{2}{*}{\textbf{K400}} & \multirow{2}{*}{\textbf{K600}} & \multirow{2}{*}{\textbf{K700}} & \multirow{2}{*}{\textbf{SSv1}} & \multirow{2}{*}{\textbf{SSv2}} & \multicolumn{3}{c}{\textbf{EK100}} & \multirow{2}{*}{\textbf{ANet}} & \multirow{2}{*}{\textbf{Charades}} \\
\addlinespace[0.3ex]
\cline{4-6}
\cline{16-18}
\addlinespace[0.3ex]

& & & C@1 & V@1 & V@5 & & & & & & & & & & Action & Verb & Noun & & & \\
\midrule

Slow Fusion \cite{karpathy2014large} & 
\cellcolor{motion}- & \cellcolor{motion}65.4 & 
\cellcolor{motion}41.9 & \cellcolor{motion}60.9 & \cellcolor{motion}80.2 & 
\cellcolor{motion}- & \cellcolor{motion}- & 
\cellcolor{motion}- & \cellcolor{motion}- & 
\cellcolor{motion}- & \cellcolor{motion}- & \cellcolor{motion}- & 
\cellcolor{motionprocedural}- & \cellcolor{motionprocedural}- & 
\cellcolor{temporalprocedural}- & \cellcolor{temporalprocedural}- & \cellcolor{temporalprocedural}- & 
\cellcolor{temporal}- & \cellcolor{temporal}- \\

Two-stream'14 \cite{simonyan2014two} & 
\cellcolor{motion}59.4 & \cellcolor{motion}88.0 & 
\cellcolor{motion}- & \cellcolor{motion}- & \cellcolor{motion}- & 
\cellcolor{motion}- & \cellcolor{motion}- & 
\cellcolor{motion}- & \cellcolor{motion}- & 
\cellcolor{motion}- & \cellcolor{motion}- & \cellcolor{motion}- & 
\cellcolor{motionprocedural}- & \cellcolor{motionprocedural}- & 
\cellcolor{temporalprocedural}- & \cellcolor{temporalprocedural}- & \cellcolor{temporalprocedural}- & 
\cellcolor{temporal}71.9 & \cellcolor{temporal}- \\

Two-stream'16 \cite{feichtenhofer2016convolutional} & 
\cellcolor{motion}69.2 & \cellcolor{motion}93.5 & 
\cellcolor{motion}- & \cellcolor{motion}- & \cellcolor{motion}- & 
\cellcolor{motion}- & \cellcolor{motion}- & 
\cellcolor{motion}- & \cellcolor{motion}- & 
\cellcolor{motion}- & \cellcolor{motion}- & \cellcolor{motion}- & 
\cellcolor{motionprocedural}- & \cellcolor{motionprocedural}- & 
\cellcolor{temporalprocedural}- & \cellcolor{temporalprocedural}- & \cellcolor{temporalprocedural}- & 
\cellcolor{temporal}- & \cellcolor{temporal}- \\

F-ST-ConvNet~\cite{sun2015human} & 
\cellcolor{motion}59.1 & \cellcolor{motion}88.1 & 
\cellcolor{motion}- & \cellcolor{motion}- & \cellcolor{motion}- & 
\cellcolor{motion}- & \cellcolor{motion}- & 
\cellcolor{motion}- & \cellcolor{motion}- & 
\cellcolor{motion}- & \cellcolor{motion}- & \cellcolor{motion}- & 
\cellcolor{motionprocedural}- & \cellcolor{motionprocedural}- & 
\cellcolor{temporalprocedural}- & \cellcolor{temporalprocedural}- & \cellcolor{temporalprocedural}- & 
\cellcolor{temporal}- & \cellcolor{temporal}- \\

Conv Pooling~\cite{yue2015beyond} & 
\cellcolor{motion}- & \cellcolor{motion}88.6 & 
\cellcolor{motion}70.8 & \cellcolor{motion}72.4 & \cellcolor{motion}90.8 & 
\cellcolor{motion}- & \cellcolor{motion}- & 
\cellcolor{motion}- & \cellcolor{motion}- & 
\cellcolor{motion}- & \cellcolor{motion}- & \cellcolor{motion}- & 
\cellcolor{motionprocedural}- & \cellcolor{motionprocedural}- & 
\cellcolor{temporalprocedural}- & \cellcolor{temporalprocedural}- & \cellcolor{temporalprocedural}- & 
\cellcolor{temporal}- & \cellcolor{temporal}- \\

C3D \cite{tran2015learning} & 
\cellcolor{motion}- & \cellcolor{motion}90.4 & 
\cellcolor{motion}46.1 & \cellcolor{motion}61.1 & \cellcolor{motion}85.2 & 
\cellcolor{motion}- & \cellcolor{motion}- & 
\cellcolor{motion}- & \cellcolor{motion}- & 
\cellcolor{motion}- & \cellcolor{motion}- & \cellcolor{motion}- & 
\cellcolor{motionprocedural}- & \cellcolor{motionprocedural}- & 
\cellcolor{temporalprocedural}- & \cellcolor{temporalprocedural}- & \cellcolor{temporalprocedural}- & 
\cellcolor{temporal}65.8 & \cellcolor{temporal}10.9 \\

RGB-I3D~\cite{carreira2017quo} & 
\cellcolor{motion}74.8 & \cellcolor{motion}95.6 & 
\cellcolor{motion}- & \cellcolor{motion}- & \cellcolor{motion}- & 
\cellcolor{motion}14.5 & \cellcolor{motion}- & 
\cellcolor{motion}- & \cellcolor{motion}29.5 & 
\cellcolor{motion}71.1 & \cellcolor{motion}71.9 & \cellcolor{motion}- & 
\cellcolor{motionprocedural}45.8 & \cellcolor{motionprocedural}- & 
\cellcolor{temporalprocedural}- & \cellcolor{temporalprocedural}- & \cellcolor{temporalprocedural}- & 
\cellcolor{temporal}- & \cellcolor{temporal}35.5 \\

Flow-I3D~\cite{carreira2017quo} & 
\cellcolor{motion}77.3 & \cellcolor{motion}96.7 & 
\cellcolor{motion}- & \cellcolor{motion}- & \cellcolor{motion}- & 
\cellcolor{motion}- & \cellcolor{motion}- & 
\cellcolor{motion}- & \cellcolor{motion}- & 
\cellcolor{motion}63.4 & \cellcolor{motion}- & \cellcolor{motion}- & 
\cellcolor{motionprocedural}- & \cellcolor{motionprocedural}- & 
\cellcolor{temporalprocedural}- & \cellcolor{temporalprocedural}- & \cellcolor{temporalprocedural}- & 
\cellcolor{temporal}- & \cellcolor{temporal}- \\

2S I3D~\cite{carreira2017quo} & 
\cellcolor{motion}80.9 & \cellcolor{motion}98.0 & 
\cellcolor{motion}- & \cellcolor{motion}- & \cellcolor{motion}- & 
\cellcolor{motion}15.6 & \cellcolor{motion}- & 
\cellcolor{motion}- & \cellcolor{motion}- & 
\cellcolor{motion}74.2 & \cellcolor{motion}75.7 & \cellcolor{motion}- & 
\cellcolor{motionprocedural}- & \cellcolor{motionprocedural}- & 
\cellcolor{temporalprocedural}- & \cellcolor{temporalprocedural}- & \cellcolor{temporalprocedural}- & 
\cellcolor{temporal}- & \cellcolor{temporal}- \\

P3D ResNet~\cite{qiu2017learning} & 
\cellcolor{motion}- & \cellcolor{motion}93.7 & 
\cellcolor{motion}47.9 & \cellcolor{motion}66.4 & \cellcolor{motion}87.4 & 
\cellcolor{motion}- & \cellcolor{motion}- & 
\cellcolor{motion}- & \cellcolor{motion}- & 
\cellcolor{motion}- & \cellcolor{motion}- & \cellcolor{motion}- & 
\cellcolor{motionprocedural}- & \cellcolor{motionprocedural}- & 
\cellcolor{temporalprocedural}- & \cellcolor{temporalprocedural}- & \cellcolor{temporalprocedural}- & 
\cellcolor{temporal}75.1 & \cellcolor{temporal}- \\

R(2+1)D RGB~\cite{tran2018closer} & 
\cellcolor{motion}74.5 & \cellcolor{motion}96.8 & 
\cellcolor{motion}57.0 & \cellcolor{motion}73.0 & \cellcolor{motion}91.5 & 
\cellcolor{motion}- & \cellcolor{motion}- & 
\cellcolor{motion}- & \cellcolor{motion}- & 
\cellcolor{motion}74.3 & \cellcolor{motion}- & \cellcolor{motion}- & 
\cellcolor{motionprocedural}- & \cellcolor{motionprocedural}- & 
\cellcolor{temporalprocedural}- & \cellcolor{temporalprocedural}- & \cellcolor{temporalprocedural}- & 
\cellcolor{temporal}- & \cellcolor{temporal}- \\

R(2+1)D Flow~\cite{tran2018closer} & 
\cellcolor{motion}76.4 & \cellcolor{motion}95.5 & 
\cellcolor{motion}46.4 & \cellcolor{motion}68.4 & \cellcolor{motion}88.7 & 
\cellcolor{motion}- & \cellcolor{motion}- & 
\cellcolor{motion}- & \cellcolor{motion}- & 
\cellcolor{motion}68.5 & \cellcolor{motion}- & \cellcolor{motion}- & 
\cellcolor{motionprocedural}- & \cellcolor{motionprocedural}- & 
\cellcolor{temporalprocedural}- & \cellcolor{temporalprocedural}- & \cellcolor{temporalprocedural}- & 
\cellcolor{temporal}- & \cellcolor{temporal}- \\

2S R(2+1)D~\cite{tran2018closer} & 
\cellcolor{motion}78.7 & \cellcolor{motion}97.3 & 
\cellcolor{motion}- & \cellcolor{motion}73.3 & \cellcolor{motion}91.9 & 
\cellcolor{motion}- & \cellcolor{motion}- & 
\cellcolor{motion}- & \cellcolor{motion}- & 
\cellcolor{motion}75.4 & \cellcolor{motion}- & \cellcolor{motion}- & 
\cellcolor{motionprocedural}- & \cellcolor{motionprocedural}- & 
\cellcolor{temporalprocedural}- & \cellcolor{temporalprocedural}- & \cellcolor{temporalprocedural}- & 
\cellcolor{temporal}- & \cellcolor{temporal}- \\

S3D~\cite{xie2018rethinking} & 
\cellcolor{motion}75.9 & \cellcolor{motion}96.8 & 
\cellcolor{motion}- & \cellcolor{motion}- & \cellcolor{motion}- & 
\cellcolor{motion}- & \cellcolor{motion}- & 
\cellcolor{motion}- & \cellcolor{motion}- & 
\cellcolor{motion}77.2 & \cellcolor{motion}- & \cellcolor{motion}- & 
\cellcolor{motionprocedural}48.2 & \cellcolor{motionprocedural}- & 
\cellcolor{temporalprocedural}- & \cellcolor{temporalprocedural}- & \cellcolor{temporalprocedural}- & 
\cellcolor{temporal}- & \cellcolor{temporal}- \\

X3D~\cite{feichtenhofer2020x3d} & 
\cellcolor{motion}- & \cellcolor{motion}- & 
\cellcolor{motion}- & \cellcolor{motion}- & \cellcolor{motion}- & 
\cellcolor{motion}- & \cellcolor{motion}27.4 & 
\cellcolor{motion}- & \cellcolor{motion}- & 
\cellcolor{motion}79.1 & \cellcolor{motion}81.9 & \cellcolor{motion}- & 
\cellcolor{motionprocedural}- & \cellcolor{motionprocedural}- & 
\cellcolor{temporalprocedural}- & \cellcolor{temporalprocedural}- & \cellcolor{temporalprocedural}- & 
\cellcolor{temporal}- & \cellcolor{temporal}47.1 \\

NL RGB-I3D~\cite{wang2018non} & 
\cellcolor{motion}- & \cellcolor{motion}- & 
\cellcolor{motion}- & \cellcolor{motion}- & \cellcolor{motion}- & 
\cellcolor{motion}- & \cellcolor{motion}- & 
\cellcolor{motion}- & \cellcolor{motion}- & 
\cellcolor{motion}77.7 & \cellcolor{motion}- & \cellcolor{motion}- & 
\cellcolor{motionprocedural}- & \cellcolor{motionprocedural}- & 
\cellcolor{temporalprocedural}- & \cellcolor{temporalprocedural}- & \cellcolor{temporalprocedural}- & 
\cellcolor{temporal}- & \cellcolor{temporal}37.5 \\
\addlinespace[0.3ex]
\hline
\addlinespace[0.3ex]
TSN~\cite{wang2016temporal} & 
\cellcolor{motion}71.0 & \cellcolor{motion}94.9 & 
\cellcolor{motion}- & \cellcolor{motion}- & \cellcolor{motion}- & 
\cellcolor{motion}- & \cellcolor{motion}- & 
\cellcolor{motion}- & \cellcolor{motion}50.1 & 
\cellcolor{motion}- & \cellcolor{motion}- & \cellcolor{motion}- & 
\cellcolor{motionprocedural}30.0 & \cellcolor{motionprocedural}33.2 & 
\cellcolor{temporalprocedural}60.2 & \cellcolor{temporalprocedural}46.0 & \cellcolor{temporalprocedural}89.6 & 
\cellcolor{temporal}- & \cellcolor{temporal}- \\

TRN~\cite{zhou2018temporal} & 
\cellcolor{motion}- & \cellcolor{motion}83.8 & 
\cellcolor{motion}- & \cellcolor{motion}- & \cellcolor{motion}- & 
\cellcolor{motion}- & \cellcolor{motion}- & 
\cellcolor{motion}- & \cellcolor{motion}28.3 & 
\cellcolor{motion}- & \cellcolor{motion}- & \cellcolor{motion}- & 
\cellcolor{motionprocedural}42.0 & \cellcolor{motionprocedural}55.5 & 
\cellcolor{temporalprocedural}35.3 & \cellcolor{temporalprocedural}65.9 & \cellcolor{temporalprocedural}45.4 & 
\cellcolor{temporal}- & \cellcolor{temporal}25.2 \\

TSM~\cite{lin2019tsm} & 
\cellcolor{motion}73.6 & \cellcolor{motion}95.9 & 
\cellcolor{motion}- & \cellcolor{motion}- & \cellcolor{motion}- & 
\cellcolor{motion}- & \cellcolor{motion}- & 
\cellcolor{motion}- & \cellcolor{motion}- & 
\cellcolor{motion}74.3 & \cellcolor{motion}- & \cellcolor{motion}- & 
\cellcolor{motionprocedural}52.6 & \cellcolor{motionprocedural}66.0 & 
\cellcolor{temporalprocedural}38.3 & \cellcolor{temporalprocedural}67.9 & \cellcolor{temporalprocedural}49.0 & 
\cellcolor{temporal}- & \cellcolor{temporal}- \\

TPN~\cite{tpn_cvpr20} & 
\cellcolor{motion}- & \cellcolor{motion}- & 
\cellcolor{motion}- & \cellcolor{motion}- & \cellcolor{motion}- & 
\cellcolor{motion}- & \cellcolor{motion}- & 
\cellcolor{motion}- & \cellcolor{motion}- & 
\cellcolor{motion}78.9 & \cellcolor{motion}- & \cellcolor{motion}- & 
\cellcolor{motionprocedural}49.0 & \cellcolor{motionprocedural}62.0 & 
\cellcolor{temporalprocedural}- & \cellcolor{temporalprocedural}61.1 & \cellcolor{temporalprocedural}41.3 & 
\cellcolor{temporal}- & \cellcolor{temporal}- \\

TIN~\cite{tin_aaai20} & 
\cellcolor{motion}72.0 & \cellcolor{motion}93.6 & 
\cellcolor{motion}- & \cellcolor{motion}- & \cellcolor{motion}- & 
\cellcolor{motion}- & \cellcolor{motion}- & 
\cellcolor{motion}- & \cellcolor{motion}- & 
\cellcolor{motion}- & \cellcolor{motion}70.4 & \cellcolor{motion}- & 
\cellcolor{motionprocedural}49.6 & \cellcolor{motionprocedural}60.0 & 
\cellcolor{temporalprocedural}- & \cellcolor{temporalprocedural}- & \cellcolor{temporalprocedural}- & 
\cellcolor{temporal}- & \cellcolor{temporal}- \\

SlowFast~\cite{feichtenhofer2019slowfast} & 
\cellcolor{motion}- & \cellcolor{motion}- & 
\cellcolor{motion}- & \cellcolor{motion}- & \cellcolor{motion}- & 
\cellcolor{motion}27.3 & \cellcolor{motion}30.7 & 
\cellcolor{motion}77.6 & \cellcolor{motion}- & 
\cellcolor{motion}79.8 & \cellcolor{motion}81.8 & \cellcolor{motion}71.0 & 
\cellcolor{motionprocedural}- & \cellcolor{motionprocedural}63.1 & 
\cellcolor{temporalprocedural}38.5 & \cellcolor{temporalprocedural}65.6 & \cellcolor{temporalprocedural}50.0 & 
\cellcolor{temporal}- & \cellcolor{temporal}45.2 \\
TVN \cite{piergiovanni2022tiny} & 
\cellcolor{motion}75.5 & \cellcolor{motion}- & \cellcolor{motion}- & \cellcolor{motion}- & \cellcolor{motion}- & 
\cellcolor{motion}- & \cellcolor{motion}- & \cellcolor{motion}- & \cellcolor{motion}30.7 & 
\cellcolor{motion}- & \cellcolor{motion}- & \cellcolor{motion}- & 
\cellcolor{motionprocedural}- & \cellcolor{motionprocedural}- & \cellcolor{temporalprocedural}- & \cellcolor{temporalprocedural}- & \cellcolor{temporalprocedural}- & 
\cellcolor{temporal}- & \cellcolor{temporal}54.6 \\
MoViNet \cite{kondratyuk2021movinets} & 
\cellcolor{motion}- & \cellcolor{motion}- & \cellcolor{motion}- & \cellcolor{motion}- & \cellcolor{motion}- & 
\cellcolor{motion}- & \cellcolor{motion}- & \cellcolor{motion}- & \cellcolor{motion}40.2 & 
\cellcolor{motion}81.5 & \cellcolor{motion}84.8 & \cellcolor{motion}72.3 & 
\cellcolor{motionprocedural}- & \cellcolor{motionprocedural}64.1 & 
\cellcolor{temporalprocedural}47.7 & \cellcolor{temporalprocedural}72.2 & \cellcolor{temporalprocedural}57.3 & 
\cellcolor{temporal}- & \cellcolor{temporal}63.2 \\

ECO \cite{zolfaghari2018eco} & 
\cellcolor{motion}72.4 & \cellcolor{motion}94.8 & \cellcolor{motion}- & \cellcolor{motion}- & \cellcolor{motion}- & 
\cellcolor{motion}- & \cellcolor{motion}- & \cellcolor{motion}- & \cellcolor{motion}- & 
\cellcolor{motion}70.0 & \cellcolor{motion}- & \cellcolor{motion}- & 
\cellcolor{motionprocedural}49.5 & \cellcolor{motionprocedural}- & \cellcolor{temporalprocedural}- & \cellcolor{temporalprocedural}- & \cellcolor{temporalprocedural}- & 
\cellcolor{temporal}- & \cellcolor{temporal}- \\

VTN \cite{neimark2021video} & 
\cellcolor{motion}- & \cellcolor{motion}- & \cellcolor{motion}- & \cellcolor{motion}- & \cellcolor{motion}- & 
\cellcolor{motion}- & \cellcolor{motion}- & \cellcolor{motion}- & \cellcolor{motion}37.4 & 
\cellcolor{motion}79.8 & \cellcolor{motion}- & \cellcolor{motion}- & 
\cellcolor{motionprocedural}- & \cellcolor{motionprocedural}- & \cellcolor{temporalprocedural}- & \cellcolor{temporalprocedural}- & \cellcolor{temporalprocedural}- & 
\cellcolor{temporal}- & \cellcolor{temporal}- \\

AssembleNet \cite{ryoo2019assemblenet} & 
\cellcolor{motion}- & \cellcolor{motion}- & \cellcolor{motion}- & \cellcolor{motion}- & \cellcolor{motion}- & 
\cellcolor{motion}- & \cellcolor{motion}- & \cellcolor{motion}- & \cellcolor{motion}34.3 & 
\cellcolor{motion}- & \cellcolor{motion}- & \cellcolor{motion}- & 
\cellcolor{motionprocedural}- & \cellcolor{motionprocedural}- & \cellcolor{temporalprocedural}- & \cellcolor{temporalprocedural}- & \cellcolor{temporalprocedural}- & 
\cellcolor{temporal}- & \cellcolor{temporal}58.6 \\
\addlinespace[0.3ex]
\hline
\addlinespace[0.3ex]
TimeSformer~\cite{bertasius2021space} & 
\cellcolor{motion}- & \cellcolor{motion}- & \cellcolor{motion}- & \cellcolor{motion}- & \cellcolor{motion}- & 
\cellcolor{motion}- & \cellcolor{motion}- & 
\cellcolor{motion}81.0 & \cellcolor{motion}- & 
\cellcolor{motion}80.7 & \cellcolor{motion}82.2 & \cellcolor{motion}- & 
\cellcolor{motionprocedural}- &  
\cellcolor{motionprocedural}62.5 & \cellcolor{temporalprocedural}- & \cellcolor{temporalprocedural}- & \cellcolor{temporalprocedural}- & 
\cellcolor{temporal}- & \cellcolor{temporal}- \\

ViViT~\cite{arnab2021vivit} & 
\cellcolor{motion}- & \cellcolor{motion}- & \cellcolor{motion}- & \cellcolor{motion}- & \cellcolor{motion}- & 
\cellcolor{motion}- & \cellcolor{motion}- & \cellcolor{motion}- & \cellcolor{motion}38.5 & 
\cellcolor{motion}84.9 & \cellcolor{motion}85.8 & \cellcolor{motion}- & 
\cellcolor{motionprocedural}- & \cellcolor{motionprocedural}65.9 & 
\cellcolor{temporalprocedural}44.0 & \cellcolor{temporalprocedural}66.4 & \cellcolor{temporalprocedural}56.8 & 
\cellcolor{temporal}- & \cellcolor{temporal}- \\

MViT~\cite{fan2021multiscale} & 
\cellcolor{motion}- & \cellcolor{motion}- & \cellcolor{motion}- & \cellcolor{motion}- & \cellcolor{motion}- & 
\cellcolor{motion}- & \cellcolor{motion}28.7 & \cellcolor{motion}- & \cellcolor{motion}- & 
\cellcolor{motion}81.2 & \cellcolor{motion}83.8 & \cellcolor{motion}- & 
\cellcolor{motionprocedural}- & \cellcolor{motionprocedural}68.7 & \cellcolor{temporalprocedural}- & \cellcolor{temporalprocedural}- & \cellcolor{temporalprocedural}- & 
\cellcolor{temporal}- & \cellcolor{temporal}47.7 \\

Motionformer \cite{patrick2021keeping} & 
\cellcolor{motion}- & \cellcolor{motion}- & \cellcolor{motion}- & \cellcolor{motion}- & \cellcolor{motion}- & 
\cellcolor{motion}- & \cellcolor{motion}- & \cellcolor{motion}- & \cellcolor{motion}- & 
\cellcolor{motion}81.1 & \cellcolor{motion}82.7 & \cellcolor{motion}- & 
\cellcolor{motionprocedural}- & \cellcolor{motionprocedural}68.1 & \cellcolor{temporalprocedural}44.5 & \cellcolor{temporalprocedural}67.0 & \cellcolor{temporalprocedural}58.5 & 
\cellcolor{temporal}- & \cellcolor{temporal}- \\

Swin~\cite{liu2022video} & 
\cellcolor{motion}- & \cellcolor{motion}- & \cellcolor{motion}- & \cellcolor{motion}- & \cellcolor{motion}- & 
\cellcolor{motion}- & \cellcolor{motion}- & \cellcolor{motion}- & \cellcolor{motion}- & 
\cellcolor{motion}84.9 & \cellcolor{motion}86.1 & \cellcolor{motion}- & 
\cellcolor{motionprocedural}- & \cellcolor{motionprocedural}69.6 & \cellcolor{temporalprocedural}- & \cellcolor{temporalprocedural}- & \cellcolor{temporalprocedural}- & 
\cellcolor{temporal}- & \cellcolor{temporal}- \\

MaskFeat \cite{wei2022masked} & 
\cellcolor{motion}- & \cellcolor{motion}- & \cellcolor{motion}- & \cellcolor{motion}- & \cellcolor{motion}- & 
\cellcolor{motion}37.8 & \cellcolor{motion}38.8 & \cellcolor{motion}- & \cellcolor{motion}- & 
\cellcolor{motion}87.0 & \cellcolor{motion}88.3 & \cellcolor{motion}80.4 & 
\cellcolor{motionprocedural}- & \cellcolor{motionprocedural}75.0 & \cellcolor{temporalprocedural}- & \cellcolor{temporalprocedural}- & \cellcolor{temporalprocedural}- & 
\cellcolor{temporal}- & \cellcolor{temporal}- \\

MViTv2~\cite{li2022mvitv2} & 
\cellcolor{motion}- & \cellcolor{motion}- & \cellcolor{motion}- & \cellcolor{motion}- & \cellcolor{motion}- & 
\cellcolor{motion}- & \cellcolor{motion}33.5 & \cellcolor{motion}- & \cellcolor{motion}- & 
\cellcolor{motion}86.1 & \cellcolor{motion}87.9 & \cellcolor{motion}79.4 & 
\cellcolor{motionprocedural}- & \cellcolor{motionprocedural}73.3 & \cellcolor{temporalprocedural}- & \cellcolor{temporalprocedural}- & \cellcolor{temporalprocedural}- & 
\cellcolor{temporal}- & \cellcolor{temporal}- \\

X-CLIP \cite{xclip_eccv} & 
\cellcolor{motion}- & \cellcolor{motion}- & \cellcolor{motion}- & \cellcolor{motion}- & \cellcolor{motion}- & 
\cellcolor{motion}- & \cellcolor{motion}- & \cellcolor{motion}- & \cellcolor{motion}- & 
\cellcolor{motion}87.1 & \cellcolor{motion}88.3 & \cellcolor{motion}- & 
\cellcolor{motionprocedural}- & \cellcolor{motionprocedural}- & \cellcolor{temporalprocedural}- & \cellcolor{temporalprocedural}- & \cellcolor{temporalprocedural}- & 
\cellcolor{temporal}- & \cellcolor{temporal}- \\

\addlinespace[0.3ex]
\hline
\addlinespace[0.3ex]

VideoMAE \cite{tong2022videomae} & 
\cellcolor{motion}73.3 & \cellcolor{motion}96.1 & \cellcolor{motion}- & \cellcolor{motion}- & \cellcolor{motion}- & 
\cellcolor{motion}- & \cellcolor{motion}39.3 & \cellcolor{motion}- & \cellcolor{motion}- & 
\cellcolor{motion}87.4 & \cellcolor{motion}- & \cellcolor{motion}- & 
\cellcolor{motionprocedural}- & \cellcolor{motionprocedural}75.4 & \cellcolor{temporalprocedural}- & \cellcolor{temporalprocedural}- & \cellcolor{temporalprocedural}- & 
\cellcolor{temporal}- & \cellcolor{temporal}- \\

VideoMAE V2 \cite{wang2023videomae} & 
\cellcolor{motion}88.1 & \cellcolor{motion}99.6 & \cellcolor{motion}- & \cellcolor{motion}- & \cellcolor{motion}- & 
\cellcolor{motion}- & \cellcolor{motion}42.6 & \cellcolor{motion}- & \cellcolor{motion}- & 
\cellcolor{motion}90.0 & \cellcolor{motion}89.9 & \cellcolor{motion}- & 
\cellcolor{motionprocedural}68.7 & \cellcolor{motionprocedural}77.0 & \cellcolor{temporalprocedural}- & \cellcolor{temporalprocedural}- & \cellcolor{temporalprocedural}- & 
\cellcolor{temporal}- & \cellcolor{temporal}- \\

V-JEPA 2 \cite{assran2025v} & 
\cellcolor{motion}- & \cellcolor{motion}- & \cellcolor{motion}- & \cellcolor{motion}- & \cellcolor{motion}- & 
\cellcolor{motion}- & \cellcolor{motion}- & \cellcolor{motion}90.2 & \cellcolor{motion}- & 
\cellcolor{motion}87.3 & \cellcolor{motion}- & \cellcolor{motion}- & 
\cellcolor{motionprocedural}- & \cellcolor{motionprocedural}77.3 & \cellcolor{temporalprocedural}39.7 & \cellcolor{temporalprocedural}63.6 & \cellcolor{temporalprocedural}57.1 & 
\cellcolor{temporal}- & \cellcolor{temporal}- \\

InternVideo \cite{wang2022internvideo} & 
\cellcolor{motion}89.3 & \cellcolor{motion}- & \cellcolor{motion}- & \cellcolor{motion}- & \cellcolor{motion}- & 
\cellcolor{motion}- & \cellcolor{motion}- & \cellcolor{motion}- & \cellcolor{motion}- & 
\cellcolor{motion}91.1 & \cellcolor{motion}91.3 & \cellcolor{motion}84.0 & 
\cellcolor{motionprocedural}70.0 & \cellcolor{motionprocedural}77.2 & \cellcolor{temporalprocedural}- & \cellcolor{temporalprocedural}- & \cellcolor{temporalprocedural}- & 
\cellcolor{temporal}94.3 & \cellcolor{temporal}- \\

InternVideo2 \cite{wang2024internvideo2} & 
\cellcolor{motion}80.7 & \cellcolor{motion}97.3 & \cellcolor{motion}- & \cellcolor{motion}- & \cellcolor{motion}- & 
\cellcolor{motion}- & \cellcolor{motion}- & \cellcolor{motion}- & \cellcolor{motion}51.2 & 
\cellcolor{motion}92.1 & \cellcolor{motion}91.9 & \cellcolor{motion}85.9 & 
\cellcolor{motionprocedural}- & \cellcolor{motionprocedural}77.5 & \cellcolor{temporalprocedural}- & \cellcolor{temporalprocedural}- & \cellcolor{temporalprocedural}- & 
\cellcolor{temporal}95.9 & \cellcolor{temporal}- \\

\bottomrule
\end{tabular}}
\label{tab:video_models}
\end{table*}

\begin{table*}[tbp]
\caption{
Comparison of representative video models across tasks. Dataset abbreviations are as follows: Video retrieval (MSR=MSR-VTT, MSVD=MSVD, LSMDC=LSMDC, ANet=ActivityNet, DiDeMo=DiDeMo, VATEX=VATEX), Temporal action localization (TH'14=THUMOS'14, ANet=ActivityNet, HACS=HACS, FineAct=FineAction), Video question answering (MSRVTT-QA, MSVD-QA, TGIF-QA, ANet-QA), and Multiple-choice (MSR-VTT, LSMDC).
Reported metrics: Recall@1 (R@1) for retrieval, average mAP for localization, top-1 accuracy for QA, and zero-shot performance for multiple-choice. 
Early backbones 
excel at localization but have limited reported performance in retrieval and QA; video-language pretrained models 
show strong retrieval and QA; instruction-tuned models (Video-ChatGPT, Grounding-GPT, Valley) 
excel at zero-shot QA.
}

\centering
\setlength{\tabcolsep}{0.15em}
\renewcommand{\arraystretch}{0.70}
\resizebox{\linewidth}{!}{%
\begin{tabular}{lccccccccccccccccccc}
\toprule
\multirow{2}{*}{\textbf{Model}} & \multicolumn{6}{c}{ Video retrieval} && \multicolumn{4}{c}{Action localization} && \multicolumn{4}{c}{Video question answering} && \multicolumn{2}{c}{Multiple-choice} \\
\addlinespace[0.3ex]
\cline{2-7}
\cline{9-12}
\cline{14-17}
\cline{19-20}
\addlinespace[0.3ex]
& \textbf{MSR} & \textbf{MSVD} & \textbf{LSMDC} & \textbf{ANet} & \textbf{DiDeMo} & \textbf{VATEX} && \textbf{TH'14} & \textbf{ANet} & \textbf{HACS} & \textbf{FineAct} && \textbf{MSRVTT} & \textbf{MSVD} & \textbf{TGIF} & \textbf{ANet} && \textbf{MSR} & \textbf{LSMDC} \\
\midrule
I3D \cite{carreira2017quo}+Flow & \cellcolor{vlm}- &\cellcolor{vlm}- &\cellcolor{vlm}- &\cellcolor{vlm} -& \cellcolor{vlm}-&\cellcolor{vlm}- & \cellcolor{temporal}& \cellcolor{temporal}66.8 & \cellcolor{temporal}35.6 &\cellcolor{temporal} - & \cellcolor{temporal}- &\cellcolor{vlm} & \cellcolor{vlm}-&\cellcolor{vlm}-  &\cellcolor{vlm}- &\cellcolor{vlm}- & \cellcolor{vlm}&\cellcolor{vlm}- &\cellcolor{vlm}- \\
R(2+1)D \cite{tran2018closer} &\cellcolor{vlm}- &\cellcolor{vlm}- &\cellcolor{vlm}- &\cellcolor{vlm}- &\cellcolor{vlm}- &\cellcolor{vlm}- &\cellcolor{temporal} & \cellcolor{temporal}55.6 & \cellcolor{temporal}36.6 & \cellcolor{temporal}- & \cellcolor{temporal}- &\cellcolor{vlm} & \cellcolor{vlm}-&\cellcolor{vlm}- &\cellcolor{vlm}- &\cellcolor{vlm}- & \cellcolor{vlm}&\cellcolor{vlm}- &\cellcolor{vlm}- \\
SlowFast~\cite{feichtenhofer2019slowfast} &\cellcolor{vlm}- &\cellcolor{vlm}- &\cellcolor{vlm}- &\cellcolor{vlm}- &\cellcolor{vlm}- &\cellcolor{vlm}- & \cellcolor{temporal}& \cellcolor{temporal}- & \cellcolor{temporal}- & \cellcolor{temporal}38.7 & \cellcolor{temporal}- & \cellcolor{vlm}& \cellcolor{vlm}-& \cellcolor{vlm}-& \cellcolor{vlm}-& \cellcolor{vlm}-&\cellcolor{vlm} &\cellcolor{vlm} -&\cellcolor{vlm}- \\
\addlinespace[0.3ex]
\hline
\addlinespace[0.3ex]
Heterogeneous \cite{fan2019heterogeneous} & \cellcolor{vlm}- & \cellcolor{vlm}- & \cellcolor{vlm}- &\cellcolor{vlm} - &\cellcolor{vlm} - &\cellcolor{vlm} - & \cellcolor{temporal}& \cellcolor{temporal}- &\cellcolor{temporal}- &\cellcolor{temporal}- &\cellcolor{temporal}- &\cellcolor{vlm} &  \cellcolor{vlm}33.0 & \cellcolor{vlm}33.7 & \cellcolor{vlm}53.8 & \cellcolor{vlm}- & \cellcolor{vlm}& \cellcolor{vlm}- & \cellcolor{vlm}- \\
VideoCLIP \cite{xu2021videoclip} & \cellcolor{vlm}30.9 & \cellcolor{vlm}- & \cellcolor{vlm}-& \cellcolor{vlm}-& \cellcolor{vlm}- & \cellcolor{vlm}- & \cellcolor{temporal}&\cellcolor{temporal}- &\cellcolor{temporal}- &\cellcolor{temporal}- &\cellcolor{temporal}- &\cellcolor{vlm} & \cellcolor{vlm}92.1 & \cellcolor{vlm}-&\cellcolor{vlm}- &\cellcolor{vlm}- & \cellcolor{vlm}& \cellcolor{vlm}-&\cellcolor{vlm}- \\
ClipBERT \cite{lei2021less} & \cellcolor{vlm}22.0  & \cellcolor{vlm}- &\cellcolor{vlm}-  &\cellcolor{vlm}21.3 & \cellcolor{vlm}20.4 & \cellcolor{vlm}- & \cellcolor{temporal}& \cellcolor{temporal}- & \cellcolor{temporal}- &\cellcolor{temporal} - & \cellcolor{temporal}- & \cellcolor{vlm}& \cellcolor{vlm}37.4 & \cellcolor{vlm}- & \cellcolor{vlm}60.3 & \cellcolor{vlm}- & \cellcolor{vlm}& \cellcolor{vlm}88.2 & \cellcolor{vlm}- \\
VIOLET \cite{fu2021violet} & \cellcolor{vlm}34.5 & \cellcolor{vlm}- & \cellcolor{vlm}16.1 & \cellcolor{vlm}-  & \cellcolor{vlm}32.6 & \cellcolor{vlm}- & \cellcolor{temporal}& \cellcolor{temporal}- & \cellcolor{temporal}- & \cellcolor{temporal}- &\cellcolor{temporal}- & \cellcolor{vlm}&\cellcolor{vlm}43.9 & \cellcolor{vlm}47.9 & \cellcolor{vlm}68.9 & \cellcolor{vlm}- & \cellcolor{vlm}& \cellcolor{vlm}- &  \cellcolor{vlm}82.9 \\
Frozen \cite{bain2021frozen} & \cellcolor{vlm}32.5 & \cellcolor{vlm}33.7 & \cellcolor{vlm}15.0 & \cellcolor{vlm}28.8 & \cellcolor{vlm}34.6 & \cellcolor{vlm}- & \cellcolor{temporal}&\cellcolor{temporal}- &\cellcolor{temporal}- &\cellcolor{temporal}- &\cellcolor{temporal} - &\cellcolor{vlm} &\cellcolor{vlm}- &\cellcolor{vlm}- &\cellcolor{vlm}- & \cellcolor{vlm}- & \cellcolor{vlm}& \cellcolor{vlm}- & \cellcolor{vlm}- \\
CLIP4Clip \cite{luo2021clip4clip} & \cellcolor{vlm}42.1 & \cellcolor{vlm}46.2 & \cellcolor{vlm}22.6 & \cellcolor{vlm}40.5 & \cellcolor{vlm}43.4 & \cellcolor{vlm}- & \cellcolor{temporal}&\cellcolor{temporal}- &\cellcolor{temporal}- &\cellcolor{temporal}- & \cellcolor{temporal}- & \cellcolor{vlm}&\cellcolor{vlm}- &\cellcolor{vlm}- &\cellcolor{vlm}- & \cellcolor{vlm}- & \cellcolor{vlm}&\cellcolor{vlm} - & \cellcolor{vlm}- \\
FrozenBiLM \cite{yang2022zero} & \cellcolor{vlm}- & \cellcolor{vlm}- & \cellcolor{vlm}- & \cellcolor{vlm}- & \cellcolor{vlm}- & \cellcolor{vlm}- & \cellcolor{temporal}& \cellcolor{temporal}- &\cellcolor{temporal}- &\cellcolor{temporal}- &\cellcolor{temporal}- & \cellcolor{vlm}&  \cellcolor{vlm}16.8 & \cellcolor{vlm}32.2 & \cellcolor{vlm}41.0 & \cellcolor{vlm}24.7 &\cellcolor{vlm} & \cellcolor{vlm}-&\cellcolor{vlm}- \\
ALPRO \cite{li2022align}& \cellcolor{vlm}33.9 & \cellcolor{vlm}- &\cellcolor{vlm}-  &\cellcolor{vlm}- & \cellcolor{vlm}35.9 & \cellcolor{vlm}- & \cellcolor{temporal}& \cellcolor{temporal}-&\cellcolor{temporal}- &\cellcolor{temporal}- &\cellcolor{temporal}- & \cellcolor{vlm}& \cellcolor{vlm}42.1 & \cellcolor{vlm}45.9 & \cellcolor{vlm}- & \cellcolor{vlm}- & \cellcolor{vlm}& \cellcolor{vlm}- & \cellcolor{vlm}-\\
InternVideo \cite{wang2022internvideo} & \cellcolor{vlm}55.2 & \cellcolor{vlm}58.4 & \cellcolor{vlm}34.0 & \cellcolor{vlm}62.2 & \cellcolor{vlm}57.9 & \cellcolor{vlm}71.1 &\cellcolor{temporal}& \cellcolor{temporal}71.6 & \cellcolor{temporal}39.0 & \cellcolor{temporal}41.6 & \cellcolor{temporal}17.6 &\cellcolor{vlm}& \cellcolor{vlm}47.1 & \cellcolor{vlm}55.5 & \cellcolor{vlm}72.2 & \cellcolor{vlm}- &\cellcolor{vlm}& \cellcolor{vlm}93.4 & \cellcolor{vlm}77.3\\
VideoMAE V2 \cite{wang2023videomae} &\cellcolor{vlm}- &\cellcolor{vlm}- &\cellcolor{vlm}- &\cellcolor{vlm}- &\cellcolor{vlm}- & \cellcolor{vlm}-&\cellcolor{temporal}& \cellcolor{temporal}69.6 & \cellcolor{temporal}- & \cellcolor{temporal}- & \cellcolor{temporal}18.2 &\cellcolor{vlm}& \cellcolor{vlm}-& \cellcolor{vlm}-& \cellcolor{vlm}-& \cellcolor{vlm}-&\cellcolor{vlm}& \cellcolor{vlm}-& \cellcolor{vlm}-\\
All-in-one \cite{wang2023all} & \cellcolor{vlm}37.3 & \cellcolor{vlm}- & \cellcolor{vlm}- & \cellcolor{vlm}22.4 & \cellcolor{vlm}32.7 & \cellcolor{vlm}- & \cellcolor{temporal}& \cellcolor{temporal}-& \cellcolor{temporal}- & \cellcolor{temporal}- & \cellcolor{temporal}- & \cellcolor{vlm}& \cellcolor{vlm}46.8 & \cellcolor{vlm}48.3 & \cellcolor{vlm}67.3 & \cellcolor{vlm}- & \cellcolor{vlm}& \cellcolor{vlm}91.9 & \cellcolor{vlm}83.9\\
Video Chat \cite{li2023videochat} & \cellcolor{vlm}- & \cellcolor{vlm}- & \cellcolor{vlm}- & \cellcolor{vlm}- & \cellcolor{vlm}- & \cellcolor{vlm}- & \cellcolor{temporal}& \cellcolor{temporal}- &\cellcolor{temporal}- &\cellcolor{temporal}- &\cellcolor{temporal}- & \cellcolor{vlm}& \cellcolor{vlm}45.0 & \cellcolor{vlm}56.3 & \cellcolor{vlm}34.4 & \cellcolor{vlm}26.5 & \cellcolor{vlm}& \cellcolor{vlm}-&\cellcolor{vlm}- \\
LLaMA Adapter \cite{zhang2023llama} & \cellcolor{vlm}- & \cellcolor{vlm}- & \cellcolor{vlm}- & \cellcolor{vlm}- & \cellcolor{vlm}- & \cellcolor{vlm}- & \cellcolor{temporal}&\cellcolor{temporal} - &\cellcolor{temporal}- &\cellcolor{temporal}- &\cellcolor{temporal}- & \cellcolor{vlm}& \cellcolor{vlm}43.8 & \cellcolor{vlm}54.9 & \cellcolor{vlm}- & \cellcolor{vlm}34.2 & \cellcolor{vlm}& \cellcolor{vlm}-&\cellcolor{vlm}- \\
Video LLaMA \cite{zhang2023video}  & \cellcolor{vlm}- & \cellcolor{vlm}- & \cellcolor{vlm}- & \cellcolor{vlm}- & \cellcolor{vlm}- & \cellcolor{vlm}- &\cellcolor{temporal} & \cellcolor{temporal}- &\cellcolor{temporal}- &\cellcolor{temporal}- &\cellcolor{temporal}- &\cellcolor{vlm} & \cellcolor{vlm}29.6 & \cellcolor{vlm}51.6 & \cellcolor{vlm}- & \cellcolor{vlm}12.4 & \cellcolor{vlm}& \cellcolor{vlm}-&\cellcolor{vlm}- \\
Video-ChatGPT \cite{maaz2023video} & \cellcolor{vlm}- & \cellcolor{vlm}- & \cellcolor{vlm}- & \cellcolor{vlm}- & \cellcolor{vlm}- & \cellcolor{vlm}- &\cellcolor{temporal} & \cellcolor{temporal}- &\cellcolor{temporal}- &\cellcolor{temporal}- &\cellcolor{temporal}- &\cellcolor{vlm} & \cellcolor{vlm}49.3 & \cellcolor{vlm}64.9 & \cellcolor{vlm}51.4 & \cellcolor{vlm}35.2 & \cellcolor{vlm}& \cellcolor{vlm}-&\cellcolor{vlm}- \\
Valley \cite{luo2023valley} & \cellcolor{vlm}- & \cellcolor{vlm}- & \cellcolor{vlm}- & \cellcolor{vlm}- & \cellcolor{vlm}- & \cellcolor{vlm}- & \cellcolor{temporal}& \cellcolor{temporal}- &\cellcolor{temporal}- &\cellcolor{temporal}- &\cellcolor{temporal}- & \cellcolor{vlm}& \cellcolor{vlm}50.8 & \cellcolor{vlm}69.2 & \cellcolor{vlm}- & \cellcolor{vlm}44.9 & \cellcolor{vlm}& \cellcolor{vlm}- &\cellcolor{vlm}- \\
InternVideo2 \cite{wang2025internvideo2} & \cellcolor{vlm}62.8 & \cellcolor{vlm}61.4 & \cellcolor{vlm}46.4 & \cellcolor{vlm}74.1 & \cellcolor{vlm}74.2 & \cellcolor{vlm}75.5 &\cellcolor{temporal}& \cellcolor{temporal}72.0 & \cellcolor{temporal}41.2 & \cellcolor{temporal}43.3 & \cellcolor{temporal}27.7 & \cellcolor{vlm} &\cellcolor{vlm}-  &\cellcolor{vlm}- &\cellcolor{vlm}- & \cellcolor{vlm}- &\cellcolor{vlm}& \cellcolor{vlm}-&\cellcolor{vlm}- \\

Video Prism \cite{videoprism_icml} & \cellcolor{vlm}52.7 & \cellcolor{vlm}- & \cellcolor{vlm}- & \cellcolor{vlm}- & \cellcolor{vlm}- & \cellcolor{vlm}62.5 &\cellcolor{temporal}& \cellcolor{temporal}- & \cellcolor{temporal}37.8 & \cellcolor{temporal}- & \cellcolor{temporal}- & \cellcolor{vlm} &\cellcolor{vlm}32.0  &\cellcolor{vlm}47.1 &\cellcolor{vlm}- & \cellcolor{vlm}- &\cellcolor{vlm}& \cellcolor{vlm}-&\cellcolor{vlm}- \\
Grounding-GPT \cite{li2024groundinggpt} & \cellcolor{vlm}- & \cellcolor{vlm}- & \cellcolor{vlm}- & \cellcolor{vlm}- & \cellcolor{vlm}- & \cellcolor{vlm}- & \cellcolor{temporal}& \cellcolor{temporal}- &\cellcolor{temporal}- &\cellcolor{temporal}- &\cellcolor{temporal}- & \cellcolor{vlm}& \cellcolor{vlm}51.6 & \cellcolor{vlm}67.8 & \cellcolor{vlm}- & \cellcolor{vlm}44.7 & \cellcolor{vlm}&\cellcolor{vlm}- &\cellcolor{vlm}- \\

\bottomrule
\end{tabular}}
\label{tab:video_models2}
\end{table*}

\section{Benchmark Evidence for Architectural Evolution}
\label{sec:benchmark}

The preceding section provides a conceptual explanation
of architectural evolution through the proposed
dataset-bias-architecture framework.
We now examine whether benchmark evidence supports
this perspective.
Tables~\ref{tab:video_models} and~\ref{tab:video_models2} show that architectural rankings vary
substantially across datasets, indicating that model success
depends strongly on the learning requirements imposed by
different benchmark regimes. Rather than interpreting these
results as isolated performance comparisons, we view them
as evidence of the relationship between dataset characteristics
and architectural suitability.

To highlight this relationship, we organize the analysis according to the dominant learning requirements exhibited by representative benchmarks (additional benchmark analysis is provided in Appendix C). Section~\ref{sec:reg-det} focuses on recognition and detection datasets, examining how motion complexity, temporal span, compositional structure, and relational interactions influence architectural performance. Section~\ref{sec:retri-reason} analyzes retrieval, localization, and question-answering benchmarks, where multimodal alignment, semantic grounding, and cross-modal reasoning become increasingly important. These results provide empirical support for the central thesis of this survey: architectural evolution in video understanding is closely coupled with the changing characteristics of the datasets on which models are developed and evaluated.

\subsection{Spatiotemporal Modeling for Recognition and Detection}
\label{sec:reg-det}

The performance trends in Table~\ref{tab:video_models} provide empirical support for the dataset-bias-architecture framework introduced in Section~\ref{sec:dataset}. Rather than revealing a universally superior architecture, the results suggest that architectural success depends strongly on the learning requirements imposed by different dataset regimes. Across benchmarks, performance improvements consistently coincide with shifts in motion complexity, temporal span, compositional structure, and relational interactions.

\textbf{Motion-centric benchmarks.}
Early action-recognition datasets such as HMDB51 and UCF101 primarily require distinguishing actions through local appearance and short-term motion cues. Consequently, architectures that explicitly model spatiotemporal dynamics achieve strong performance. Two-Stream networks already surpass 93\% accuracy on UCF101, while RGB-I3D further exceeds 95\%, demonstrating the effectiveness of motion-sensitive representations when discriminative information is concentrated within short temporal windows. These results suggest that for coarse, high-amplitude actions, modeling local motion and appearance remains the dominant source of predictive power. The strong performance of convolution-based architectures on these benchmarks also explains why two-stream networks and 3D CNNs dominated the early development of video understanding.

\textbf{Temporal and compositional benchmarks.}
The transition to datasets such as Something-Something, Charades, ActivityNet, and Breakfast introduces fundamentally different challenges. In these benchmarks, actions are often defined by object manipulations, temporal ordering, and procedural context rather than appearance alone. As a result, architectures designed to preserve temporal structure consistently outperform models relying primarily on local spatiotemporal cues. For example, TRN demonstrated substantial gains on Something-Something by explicitly modeling temporal relations between frames, while later architectures such as TSM and temporal transformers further improved performance through more effective long-range dependency modeling. The gap between performance on UCF101 and Something-Something is particularly revealing: architectures that achieve near-saturated accuracy on UCF101 often experience substantial performance drops on Something-Something, indicating that temporal reasoning rather than motion recognition becomes the primary bottleneck. These observations suggest that increasing temporal span and compositional complexity fundamentally reshape the representations required for successful prediction.

\textbf{Relational and interaction-centric benchmarks.}
Benchmarks such as AVA, EPIC-KITCHENS, and Ego4D further expand the scope of video understanding by emphasizing human-object interactions, multi-agent activities, and egocentric perception. Success in these settings depends less on recognizing actions in isolation and more on understanding relationships among actors, objects, and surrounding context. Consistent with this shift, transformer-based architectures and large-scale pretrained models generally outperform traditional convolutional approaches. Models such as MotionFormer, VideoMAE, and InternVideo achieve strong performance because attention mechanisms naturally support interaction modeling across space and time. The benchmark trends are consistent with the representation-level evidence presented in Section~\ref{sec:dataset}, where identical architectures pretrained on different datasets develop markedly different attention patterns. These observations suggest that interaction-centric datasets not only require different architectural capabilities but also encourage distinct representations that may transfer imperfectly across benchmark regimes.

\textbf{Large-scale recognition datasets and foundation pretraining.}
On large-scale benchmarks such as Kinetics-400, Kinetics-600, and Kinetics-700, performance increasingly correlates with representation quality, pretraining scale, and model capacity. Architectures including Swin Transformer, MViT, VideoMAE, InternVideo, and InternVideo2 consistently outperform earlier CNN-based approaches. Importantly, these gains are not solely attributable to larger architectures. Large-scale pretraining enables models to acquire transferable motion, temporal, and semantic representations that generalize across diverse downstream tasks. The progression from supervised architectures such as I3D and SlowFast to self-supervised approaches such as VideoMAE and foundation-scale models such as InternVideo reflects a growing emphasis on transferable representations acquired through large-scale pretraining. This trend foreshadows the emergence of video foundation models capable of supporting a broad range of downstream tasks using a shared representation space.

These benchmarks demonstrate that different recognition datasets favor substantially different representational and architectural properties. Performance gains are therefore best understood in relation to the specific challenges emphasized by each benchmark rather than as evidence of universally superior architectures.

\subsection{Multimodal Alignment for Retrieval and Reasoning}
\label{sec:retri-reason}

The results in Table~\ref{tab:video_models2} reveal a second major trajectory in video understanding: the transition from spatiotemporal modeling toward multimodal alignment and, ultimately, language-guided reasoning. Unlike action recognition benchmarks, where success is primarily determined by visual and temporal representations, retrieval and question-answering tasks introduce additional learning requirements centered on semantic grounding, cross-modal correspondence, and reasoning over language. The benchmark trends reveal a clear progression from temporal modeling, to multimodal alignment, to general-purpose multimodal reasoning.

\textbf{Temporal representations as a foundation.}
Early architectures such as I3D, R(2+1)D, and SlowFast consistently achieve strong performance on temporal localization benchmarks, including THUMOS'14, ActivityNet, and HACS. Their strong localization accuracy demonstrates an ability to capture fine-grained spatiotemporal dynamics and identify action boundaries effectively. However, these architectures are largely absent from retrieval and question-answering leaderboards. The contrast is revealing: while temporal modeling is sufficient for recognizing and localizing actions, it does not automatically provide the semantic representations required for aligning videos with natural language. These results suggest that strong visual and temporal representations constitute a necessary foundation for video understanding, but are insufficient for multimodal reasoning tasks that require semantic grounding.

\textbf{Cross-modal alignment through video-language pretraining.}
The emergence of large-scale video-language datasets fundamentally changes the dominant learning requirement. Benchmarks such as MSR-VTT, MSVD, VATEX, ActivityNet Retrieval, and DiDeMo reward models that align visual content with textual descriptions rather than merely recognize actions. Consistent with this shift, architectures such as CLIP4Clip, Frozen, VideoCLIP, VIOLET, and ALPRO achieve substantial gains in retrieval and question answering through large-scale video-language pretraining. For example, retrieval performance on MSR-VTT improves markedly, with representative video-language models achieving R@1 scores exceeding 30--40\%, while VideoCLIP and VIOLET also demonstrate strong performance on question-answering benchmarks. These improvements indicate that paired video-text data induces representational biases centered on semantic correspondence and language grounding, enabling transfer across retrieval, captioning, and reasoning tasks. The results therefore provide strong evidence that multimodal datasets encourage fundamentally different representations from those learned on conventional action-recognition benchmarks.

\textbf{Foundation models and unified multimodal representations.}
A further transition occurs with foundation-scale models such as InternVideo and InternVideo2. Unlike earlier video-language architectures that primarily target retrieval, these models achieve strong performance across retrieval, localization, and reasoning benchmarks simultaneously. InternVideo and InternVideo2 obtain state-of-the-art results on retrieval datasets including MSR-VTT, ActivityNet, and VATEX while also maintaining competitive performance on localization benchmarks such as THUMOS'14. This broad improvement suggests that large-scale multimodal pretraining does more than strengthen cross-modal alignment; it enables the emergence of transferable representations that generalize across tasks with substantially different objectives. The ability to perform well on retrieval, localization, and question answering using a shared representation suggests that large-scale multimodal pretraining enables increasingly general representations that transfer across tasks with substantially different objectives.

\textbf{Instruction tuning and reasoning-centric representations.}
Recent instruction-tuned architectures, including Video-ChatGPT, Valley, and Grounding-GPT, introduce yet another learning requirement: open-ended reasoning over video content. Rather than optimizing exclusively for retrieval or classification, these models are trained to follow natural language instructions and generate contextually grounded responses. Their strongest gains are typically observed on question-answering benchmarks, where they outperform earlier retrieval-oriented models despite limited task-specific supervision. This behavior suggests that instruction tuning encourages representations that integrate perception, semantic grounding, and reasoning within a unified framework. More broadly, it mirrors developments across multimodal AI, where instruction alignment increasingly serves as a bridge between representation learning and general-purpose reasoning.
The benchmark evidence highlights the increasing importance of semantic grounding and multimodal alignment in modern video understanding. Models trained on large-scale video-language corpora consistently outperform architectures optimized solely for visual recognition, indicating that cross-modal representations have become a key determinant of performance across retrieval and reasoning tasks.

\subsection{Cross-Dataset Trends and Design Implications}

Several consistent trends emerge from the benchmark evidence summarized in Tables~\ref{tab:video_models} and~\ref{tab:video_models2}.
The benchmark evidence demonstrates that architectural progress cannot be fully understood through model design alone. Performance trends consistently reflect the interaction between dataset characteristics, learned representations, and architectural capabilities. Viewed from this perspective, benchmark datasets function not merely as evaluation tools, but as forces that shape the directions of representation learning and architectural development.

\textbf{Architectural success is dataset-dependent.}
The results provide little evidence for a universally optimal video architecture. Models that perform strongly on motion-centric benchmarks such as UCF101, HMDB51, and THUMOS'14 do not necessarily maintain the same advantage on temporally complex datasets such as Something-Something and ActivityNet, nor on multimodal benchmarks such as MSR-VTT, VATEX, and MSVD-QA. For example, architectures optimized for short-term motion modeling achieve near-saturated performance on classical action-recognition benchmarks yet often experience substantial performance degradation when temporal reasoning, semantic grounding, or multimodal alignment become the dominant challenge. These 
suggest that architectural success depends primarily on how well a model's inductive biases align with the learning requirements imposed by the target dataset.

\textbf{Video understanding bottlenecks have evolved.}
Across successive generations of benchmarks, the factors limiting performance have changed. 
Early benchmarks primarily rewarded accurate motion recognition, whereas later datasets increasingly emphasize temporal dependencies, interaction modeling, semantic grounding, and cross-modal reasoning. As new challenges emerge, architectures evolve to address the dominant sources of error exposed by each generation of datasets.

\textbf{Representations increasingly matter more than architecture.}
A notable trend across modern benchmarks is the growing importance of transferable representations. Earlier architectures often relied on carefully engineered temporal operators tailored to specific tasks, whereas recent foundation-scale models achieve competitive performance across recognition, retrieval, localization, and reasoning benchmarks using a shared pretrained representation. The strong cross-task performance of models such as VideoMAE, InternVideo, and InternVideo2 suggests that large-scale pretraining increasingly serves as the primary mechanism for acquiring motion, temporal, semantic, and multimodal knowledge. As a result, representation learning is becoming a more important determinant of performance than task-specific architectural modifications alone.

\textbf{Multimodal alignment is emerging as a unifying principle.}
The strongest-performing systems on modern retrieval, question-answering, and reasoning benchmarks consistently integrate temporal understanding with language grounding. Video-language pretraining improves retrieval and semantic correspondence, while instruction tuning further extends these capabilities toward open-ended reasoning and interaction. This progression suggests that future advances in video understanding will depend increasingly on architectures capable of jointly modeling visual dynamics, semantic knowledge, language, and reasoning within a unified framework. In this sense, multimodal alignment is no longer merely an auxiliary capability but is becoming a central organizing principle of modern video understanding systems.

\section{A Dataset-Centric Roadmap}

The analyses presented in Sections \ref{sec:dataset} and \ref{sec:benchmark} suggest that progress in video understanding is fundamentally driven by the co-evolution of datasets, representations, and architectures. Rather than emerging as isolated algorithmic advances, successful architectures consistently arise in response to new learning requirements introduced by increasingly complex data regimes. 
Below we organize the discussion into two complementary parts. First, we synthesize how successive generations of datasets have driven architectural innovation throughout the evolution of video understanding. Second, we discuss emerging dataset regimes and the architectural opportunities they create for future video understanding systems. The resulting perspective not only explains the historical trajectory of the field, but also provides practical guidance for future dataset design, model development, and the construction of general-purpose video understanding systems. 
Additional discussion can be found in Appendix D.

\subsection{Datasets as Engines of Architectural Innovation}

A central argument of this survey is that architectural innovation in video understanding is rarely driven by model design alone. Instead, major architectural advances consistently emerge when new datasets introduce learning requirements that existing representations cannot adequately satisfy. Viewed through the dataset-bias-architecture framework, advances in video understanding are shaped not only by larger models or improved optimization, but also by the structural characteristics of the datasets on which these models are developed and evaluated.
These observations reveal a recurring pattern: dataset characteristics define learning requirements, learning requirements induce representational biases, and architectures evolve to exploit these biases more effectively.
Table~\ref{tab:datasets_structural_lenses} reveals a clear historical pattern. Early benchmarks such as KTH, Weizmann, HMDB51, and UCF101 primarily consisted of short, trimmed clips in which discriminative information was concentrated in local appearance and motion cues. These datasets imposed relatively limited temporal requirements and naturally favored handcrafted descriptors, two-stream networks, and 3D convolutional architectures optimized for short-range spatiotemporal patterns. As larger benchmarks such as Kinetics expanded visual diversity, category coverage, and scale, architectural progress increasingly depended on richer spatiotemporal representations and larger model capacity.

The emergence of datasets such as Something-Something, Breakfast, ActivityNet, and Charades exposed the limitations of appearance-centric recognition. In these benchmarks, actions are often defined by object manipulations, temporal ordering, and procedural context rather than static visual content alone. Consequently, the dominant learning requirement shifted from motion recognition to temporal reasoning. Models were required not only to identify actions but also to understand how actions evolve over time and how individual steps contribute to broader activities. This transition motivated temporal aggregation networks, memory-based architectures, and transformers capable of modeling long-range dependencies and compositional structure. More broadly, it illustrates a recurring principle: when dataset characteristics change, the dominant architectural bottleneck changes accordingly.

Relational and interaction-centric datasets introduced a further stage of evolution. Benchmarks such as AVA, EPIC-KITCHENS, and Ego4D require reasoning about human-object interactions, multiple agents, egocentric viewpoints, and evolving environmental context. Success in these settings depends less on recognizing actions in isolation and more on understanding relationships among entities across space and time. As a result, graph-based representations, attention mechanisms, and interaction-centric transformers have become increasingly important. The attention visualizations discussed in Section~\ref{sec:dataset} provide qualitative evidence for this phenomenon. Despite sharing the same model architecture, models pretrained on Kinetics, Something-Something, and EPIC-KITCHENS develop markedly different attention patterns, focusing on actors and scene context, object state transitions, and hand-object interactions, respectively. These observations suggest that dataset evolution first reshapes learned representations before ultimately driving architectural evolution.

The most recent generation of datasets extends these requirements into the multimodal domain. Large-scale corpora such as HowTo100M, WebVid, InternVid, and related video-language datasets introduce semantic grounding, cross-modal correspondence, and open-ended reasoning as primary learning objectives. These datasets have accelerated the emergence of video-language pretraining, video foundation models, and video-LLMs, where success increasingly depends on aligning visual dynamics with language, audio, and external knowledge. The benchmark trends in Table~\ref{tab:video_models2} reflect this transition, with multimodal pretraining and instruction tuning producing 
gains in retrieval, question answering, and zero-shot transfer.

These developments suggest a practical principle for model design. Motion-centric datasets favor models with strong temporal representations; datasets characterized by long temporal span and procedural structure benefit from memory and attention mechanisms; interaction-rich datasets require relational reasoning capabilities; and multimodal datasets increasingly reward architectures capable of aligning visual, linguistic, and auditory information. Consequently, architectural selection should be guided not only by computational considerations, but also by the dominant learning requirements encoded in the target dataset.

The historical trajectory of video understanding shows that datasets do not merely evaluate progress. By defining the learning requirements that successful models should satisfy, they shape the representational biases that emerge and ultimately determine the architectural paradigms that dominate the field. Future advances are therefore likely to be driven as much by new dataset regimes as by new model architectures.

\subsection{Emerging Dataset Regimes and Future Directions}

The central argument of this survey is that architectural evolution follows dataset evolution. Historically, each major generation of video datasets introduced new learning requirements that exposed limitations in existing representations and ultimately motivated new architectural paradigms. Motion-centric benchmarks accelerated spatiotemporal convolutions, procedural datasets stimulated temporal reasoning architectures, interaction-rich corpora encouraged relational representations, and multimodal datasets catalyzed video-language pretraining and foundation models. From this perspective, forecasting the future of video understanding requires forecasting the characteristics of future datasets.

A first emerging trend is the \emph{convergence of dataset dimensions}. Most existing benchmarks emphasize only a subset of the structural characteristics identified in this survey. Motion-centric datasets focus primarily on temporal dynamics, procedural datasets emphasize temporal span and compositional structure, interaction-centric datasets prioritize relational reasoning, and video-language corpora focus on multimodal grounding. Future datasets are likely to combine these requirements simultaneously, requiring models to reason about fine-grained motion, long-range temporal dependencies, hierarchical procedures, multi-agent interactions, and multimodal information within a unified environment. Such datasets will place increasing pressure on architectures to move beyond specialized solutions and develop unified representations that support multiple forms of reasoning concurrently.

A second trend is the \emph{expansion of temporal horizons}. Although recent benchmarks have substantially increased temporal coverage, most current models continue to operate at the clip level or over relatively short temporal windows. Real-world activities frequently unfold over minutes or hours, involving delayed dependencies, evolving intentions, and hierarchical goals. Future datasets will likely expose richer long-range temporal structure, creating demand for architectures with persistent memory, hierarchical temporal abstraction, event-centric representations, and scalable mechanisms for reasoning across extended durations. In this regime, memory may become as important as perception.

A third trend is the \emph{growing importance of compositional and relational generalization}. Many current benchmarks contain recurring combinations of actions, objects, and environments that enable models to exploit statistical regularities without learning deeper structure. Future datasets are likely to place greater emphasis on novel compositions, unseen action sequences, and unfamiliar interaction patterns. Such benchmarks will reward architectures capable of modeling relationships, abstractions, and reusable behavioral primitives rather than memorizing frequent correlations. This shift may accelerate interest in structured representations, neuro-symbolic reasoning, and models that explicitly capture compositional structure.

A fourth trend is the \emph{evolution of multimodal supervision}. Existing video-language datasets have enabled substantial progress in retrieval, captioning, and question answering, but many remain weakly aligned, temporally coarse, and susceptible to language shortcuts. Future datasets will increasingly provide dense temporal grounding, richer multimodal annotations, and stronger supervision of cross-modal correspondences. These developments are likely to encourage architectures that integrate visual, linguistic, auditory, and contextual information within unified representations rather than treating multimodal fusion as a downstream component. Such datasets may further accelerate the emergence of general-purpose video foundation models capable of transferring across diverse tasks and modalities.

These emerging dataset regimes suggest \emph{a corresponding shift in architectural priorities}. Future video understanding systems will likely require four complementary capabilities: temporal memory for long-horizon reasoning, relational reasoning for interaction understanding, compositional representations for generalization, and multimodal grounding for semantic interpretation. Rather than being implemented as independent modules, these capabilities may increasingly be unified within foundation-model architectures that learn shared representations across tasks, modalities, and environments.
Equally important is the evolution of evaluation protocols. If future datasets are designed to expose richer learning requirements, evaluation should evolve accordingly. Beyond benchmark-specific accuracy, future protocols should emphasize cross-dataset transfer, long-horizon reasoning, compositional generalization, robustness under distribution shift, multimodal consistency, and efficiency. Such evaluations would provide a more faithful assessment of whether models have acquired the capabilities required for real-world video understanding.
Viewed through the dataset-bias-architecture framework, the future trajectory of video understanding is ultimately determined by dataset design. New datasets introduce new learning requirements; these requirements reshape representational biases; and new biases create demand for new architectures. Future benchmarks should therefore be viewed not merely as evaluation platforms, but as instruments for steering the next generation of video understanding systems toward increasingly general, adaptive, and multimodal intelligence.

\section{Conclusion}

Video understanding has evolved from recognizing isolated actions to supporting temporal reasoning, relational understanding, multimodal grounding, and increasingly general video intelligence. In this survey, we adopted a dataset-centric perspective, arguing that progress in the field is best understood as a co-evolution of datasets, representations, and architectures.
To formalize this relationship, we introduced a \emph{dataset-bias-architecture} framework, showing how dataset characteristics shape learning requirements, induce representational biases, and ultimately drive architectural innovation. Through analyses of representative datasets, models, and benchmarks, we demonstrated that major architectural transitions, from two-stream networks and 3D CNNs to transformers, video-language models, and video foundation models, closely follow changes in dataset structure and complexity.
Building on these insights, we presented a dataset-centric roadmap for future research. We argue that future progress will be driven by datasets that jointly expose fine-grained motion, long-range temporal dependencies, compositional activities, relational interactions, and multimodal grounding, motivating architectures that integrate these capabilities within unified video foundation models.
Ultimately, datasets do not merely evaluate video understanding systems; they shape the representations they learn and the architectures that emerge. Understanding this relationship provides a principled framework for explaining past advances and guiding the development of future video understanding systems.

\appendices

\input{appendix}

\bibliographystyle{IEEEtran}
\bibliography{research}


\begin{IEEEbiography}
[{\includegraphics[width=1in,height=1.25in,clip,keepaspectratio]{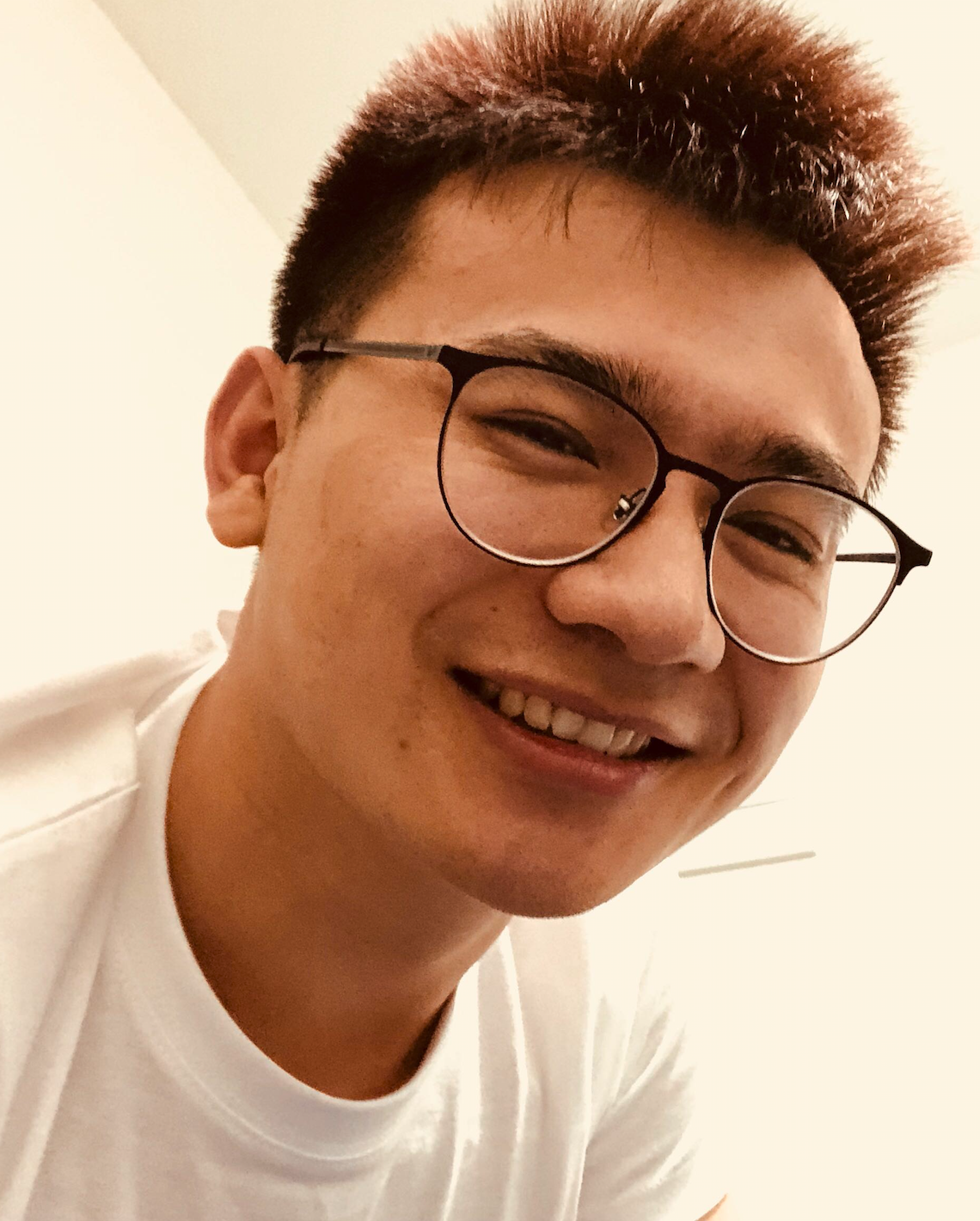}}]{Lei Wang} received his M.E. in Software Engineering from the University of Western Australia (UWA) in 2018 and his Ph.D. in Engineering and Computer Science from the Australian National University (ANU) in 2023. He is a Research Fellow in the School of Electrical and Electronic Engineering at Griffith University and a Visiting Scientist with Data61/CSIRO. He leads the Temporal Intelligence and Motion Extraction (TIME) Lab at Griffith University. He previously held research positions at ANU, UWA, and Data61/CSIRO. His research focuses on motion-, data-, and model-centric approaches to action recognition and anomaly detection. 
He has authored numerous first-author papers in top-tier venues, including CVPR, ICCV, ECCV, ACM Multimedia, NeurIPS, ICLR, ICML, AAAI, TPAMI, IJCV, and TIP, and received the Sang Uk Lee Best Student Paper Award at ACCV 2022. He serves as an Area Chair for NeurIPS 2026, ACM Multimedia 2024-2026, ICASSP 2025, and ICPR 2024, and was recognized as an Outstanding Area Chair at ACM Multimedia 2024.
\end{IEEEbiography}


\begin{IEEEbiography}
[{\includegraphics[trim=0 0 0 300, width=1in,height=1.25in,clip,keepaspectratio]{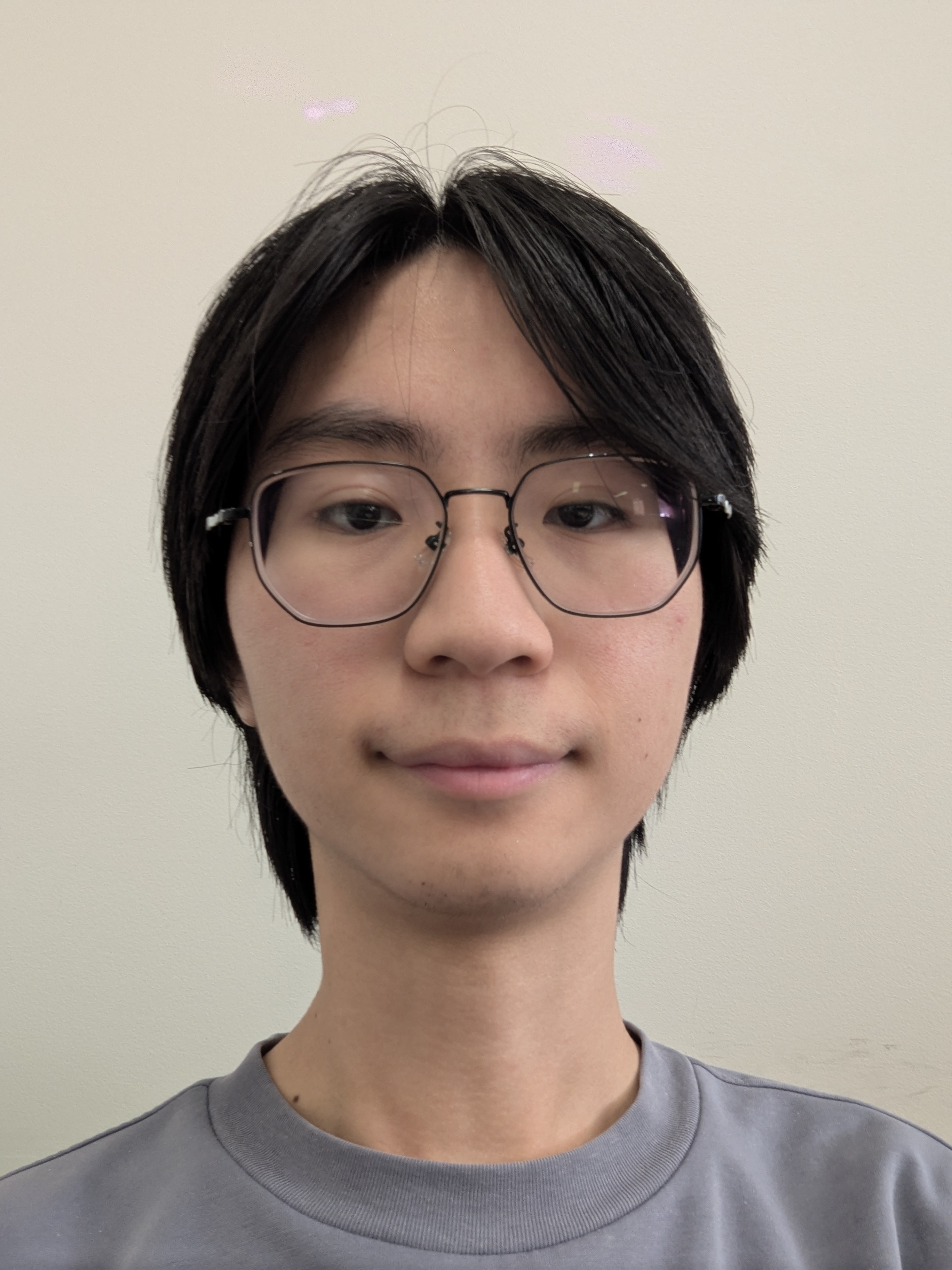}}
]{Syuan-Hao Li} received the B.S. degree in Computer Science, National Taitung University (NTTU), Taiwan, in 2025. He is currently a Ph.D. pathway student at Griffith University and a research intern at the Temporal Intelligence and Motion Extraction (TIME) Lab. He serves as a workshop coordinator for TIME 2026: the 2nd International Workshop on Transformative Insights in Multi-faceted Evaluation, hosted at the Web Conference (WWW 2026). His research interests include temporal modeling, multimodal intelligence, and fine- and ultra-fine-grained visual understanding.
\end{IEEEbiography}


\begin{IEEEbiography}[{\includegraphics[trim=80 80 30 0,width=1in,height=1.25in,clip,keepaspectratio]{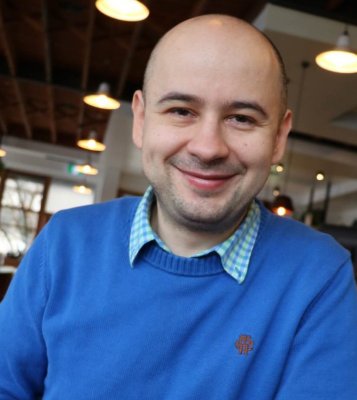}}]{Piotr Koniusz} received the BSc degree in Telecommunications and Software Engineering from Warsaw University of Technology, Poland, in 2004, and the PhD degree in Computer Vision from CVSSP, University of Surrey, U.K., in 2013. He is an Associate Professor in Theoretical ML at the University of New South Wales (UNSW) and a Principal (now Visiting) Researcher with the Machine Learning Research Group, Data61/CSIRO. He was previously a postdoctoral researcher with the LEAR team at INRIA, France. 
His research interests include representation learning (contrastive and self-supervised learning, unlearning), vision-language models, MLLMs, and deep and graph neural networks, as well as Machine Learning Safety. He has received awards including the Sang Uk Lee Best Student Paper Award (ACCV 2022), Runner-up APRS/IAPR Best Student Paper Award (DICTA 2022), and Outstanding Area Chair recognition (ICLR 2021--2023). He served as a Program Chair for NeurIPS 2025 and serves as a Senior Workshop Program Chair for NeurIPS 2026, and a Journal Track Chair for ACML 2026.\end{IEEEbiography}


\begin{IEEEbiography}[{\includegraphics[trim=0 0 0 0, width=1in,height=1.25in,clip,keepaspectratio]{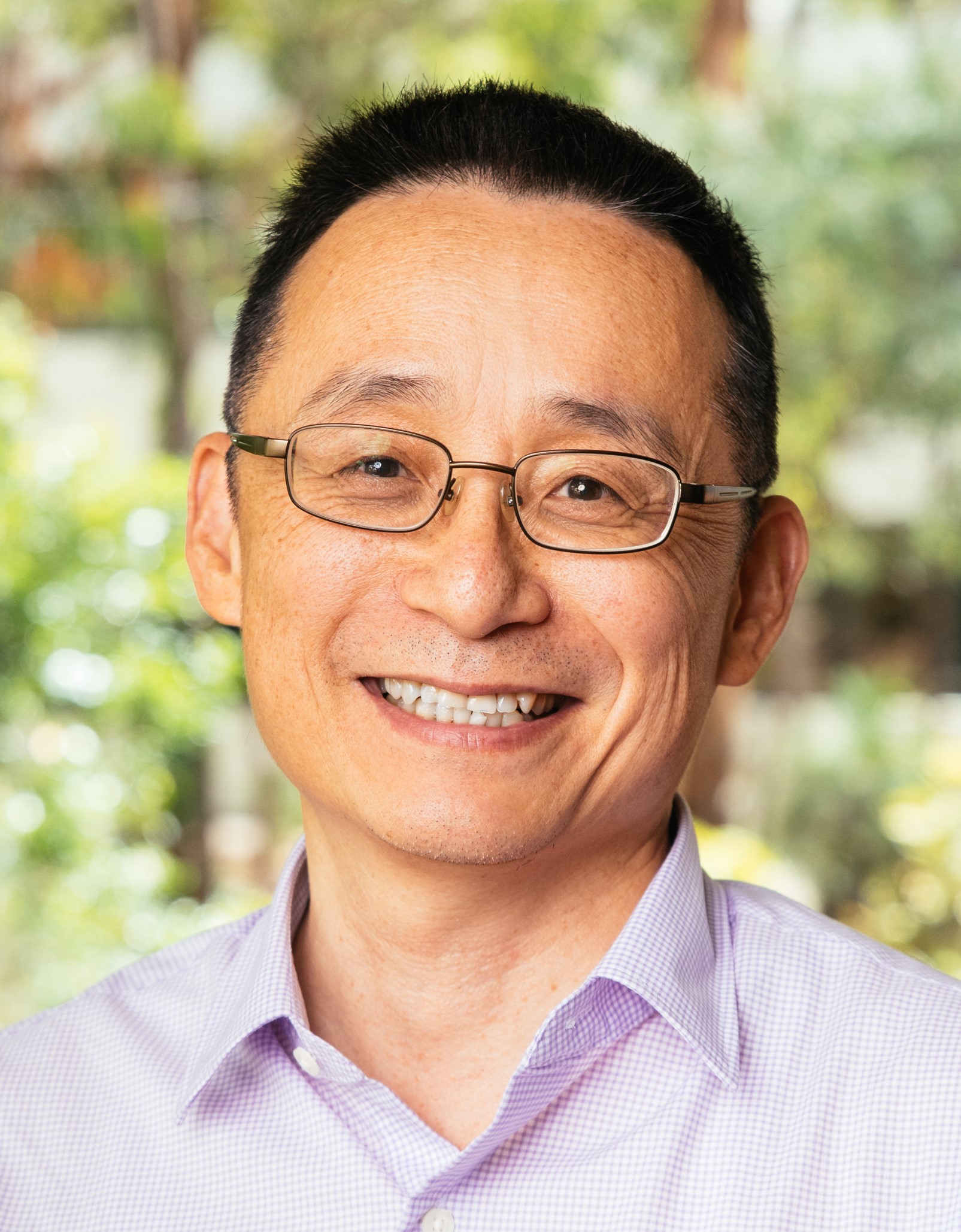}}
]{Yongsheng Gao} (Senior Member, IEEE) received the BSc and MSc degrees in Electronic Engineering from Zhejiang University, China, in 1985 and 1988, respectively, and the PhD degree in Computer Engineering from Nanyang Technological University, Singapore. He is currently a Professor with the School of Engineering and Built Environment, Griffith University, and Director of the ARC Research Hub for Driving Farming Productivity and Disease Prevention, Australia. He was previously the Leader of the Biosecurity Group at the Queensland Research Laboratory, National ICT Australia (ARC Centre of Excellence), a consultant at Panasonic Singapore Laboratories, and an Assistant Professor at Nanyang Technological University. His research interests include smart farming, machine vision for agriculture, biosecurity, face recognition, biometrics, image retrieval, computer vision, pattern recognition, environmental informatics, and medical imaging. He is a recipient of the 2025 ARC Industry Laureate Fellow.\end{IEEEbiography}

\end{document}

%% file: appendix.tex
\section{Discussion of Existing Surveys}

The literature on video understanding is vast and rapidly expanding. Existing survey efforts can broadly be grouped into four complementary strands: (i) general surveys summarizing methods for action recognition and video analysis (\eg, \cite{shih2017survey, herath2017going, pareek2021survey, madan2024foundation, tang2025video, ding2025journey}), (ii) focused reviews on specific architectural families (\eg, CNNs, RNNs, transformers) or modalities (\eg, skeleton, audio-visual, vision-language) (\eg, \cite{ahmad2005video, aafaq2019video, wang2019comparative,sun2022human, selva2023video, ding2025language,tang2025video}), (iii) evaluations and benchmark-driven comparative analyses (\eg, \cite{liu2016benchmarking, carreira2017quo, singh2019video, yao2019review, guo2024benchmarking, liurepresentation}), and (iv) surveys on emerging paradigms such as self-supervision, generative modeling, and reinforcement/continual learning (\eg, \cite{wang2019generative, wang2021actionclip, kong2022human, aldausari2022video, xing2024survey, zhao2023search, schiappa2023self}). 
In this section, we group prior surveys into coherent categories and highlight how our work extends their scope by adopting a dataset-centric perspective. Our goal is not to exhaustively catalog prior work but to position representative survey efforts relative to the unique contributions of this article.

\textbf{Classical and deep-learning surveys.} Early surveys focused on handcrafted features and classical pipelines (\eg, spatio-temporal interest points \cite{laptev2005space}, HOG3D \cite{dalal2005histograms}, dense trajectories \cite{wang2013dense}) before tracing the transition to deep learning. Representative examples include Aggarwal and Ryoo \cite{aggarwal2011human} (pre-deep era) and Herath \etal \cite{herath2017going}, which reviewed the evolution from 2D CNNs to 3D CNNs and two-stream architectures. These works documented historical backbone evolution and early benchmark evaluations (\eg, \cite{kuehne2011hmdb, soomro2012ucf101, karpathy2014large}).
While foundational, these surveys were largely architecture-centric and focused on classification tasks. They overlooked how dataset structural properties (\eg, temporal span, compositional complexity, multimodal richness) induce dataset-driven inductive biases in architectures and paradigms, and did not account for the influence of large-scale multimodal pretraining, VLMs, or LLMs on model evolution. In contrast, our survey systematically links dataset characteristics to both architectural and paradigm choices, showing the structural pressures that drive model innovation and task-specific performance.

\textbf{Transformer and modern architecture surveys.} With the rise of attention-based models, several surveys and tutorials have examined transformers in vision and video (\eg, Video Transformers and ViT extension \cite{khan2022transformers, ulhaq2022vision, selva2023video}), analyzing design patterns (\eg, space-time factorization) and computational trade-offs with benchmark comparisons.
While indispensable for understanding architectural design, these surveys remain narrow. They seldom consider (i) dataset-aligned inductive biases beyond attention, such as graph-based structures or generative modeling priors, (ii) interactions between architectures and learning paradigms, including how masked or multimodal objectives reshape transformers' effective biases, or (iii) dataset-driven motivations, such as which dataset properties catalyzed the shift toward transformers. Our work situates transformers within a broader taxonomy and explicitly analyzes how dataset characteristics influence their downstream utility and task-specific performance.

\textbf{Multimodal and VLM/LLM surveys.} As video understanding increasingly incorporates language and audio, surveys have emerged on multimodal learning (\eg, video captioning, retrieval, cross-modal pretraining \cite{miech2020end, xu2023multimodal, madan2024foundation, tang2025video}). These works largely catalogue datasets, alignment objectives, and evaluation protocols without explaining how dataset characteristics shape multimodal model design and paradigm selection.
We advance this perspective by (i) treating multimodal alignment as a dataset-informed paradigm, reflecting invariances induced by different data regimes, (ii) analyzing modality imbalance and modality-dominance failure modes as consequences of dataset properties, and (iii) situating VLM- and LLM-augmented models within hybrid pipelines integrating generative priors and self-supervision, highlighting how dataset structure guides paradigm and architectural choices.

\textbf{Structured representations.} Many surveys cover skeleton-based action recognition and graph-based approaches for modeling human-object interactions, multi-agent dynamics, and relational reasoning~\cite{subetha2016survey, yan2018spatial, wang2019comparative, cheng2020skeleton, ahmad2021graph, mourot2022survey, ren2024survey, liurepresentation}. While valuable, these works largely remain modality-specific and rarely consider how dataset properties influence the choice of representational paradigms.
Our survey situates graph-based representations within a broader architecture and paradigm design space, showing how dataset characteristics, \eg, agent density, relational complexity, compositional depth, drive the selection of graph, attention, or self-supervised representations for effective spatiotemporal and relational reasoning.

\textbf{Self-supervision, generative, and pretraining surveys.} Surveys on self-supervised and generative modeling~\cite{jing2020self,liu2021self, ericsson2022self, oussidi2018deep,suzuki2022survey, cho2024sora, xing2024survey, wang2025feature} summarize objectives, augmentations, and downstream transfer. While informative, they seldom analyze how dataset characteristics guide pretraining paradigm selection.
In contrast, our survey emphasizes alignment between dataset properties and learning paradigms, highlighting when contrastive, masked, or generative objectives are most effective, and how hybridization or curriculum strategies can be guided by structural and temporal dataset properties.

\textbf{Benchmarks and dataset-centered studies.} Several surveys focus on benchmarks \cite{liu2016benchmarking, carreira2017quo, singh2019video, yao2019review, guo2024benchmarking, plizzari2024outlook, liurepresentation}, evaluation practices, and dataset analyses (\eg, Kinetics \cite{kay2017kinetics}, EPIC-KITCHENS \cite{damen2018scaling, damen2022rescaling}). While useful, they often treat datasets as static resources.
In contrast, we adopt datasets as structural lenses, analyzing properties such as temporal span, compositionality, annotation granularity, multimodal richness, and agent density, and show how these drive paradigm and architectural choices, enabling more principled guidance for model design, pretraining, and dataset construction.

\textbf{Reinforcement, continual, and privacy-aware learning.} Some surveys review reinforcement learning, continual learning, and federated/privacy-preserving approaches, highlighting challenges such as catastrophic forgetting, non-i.i.d. data, and sparse rewards~\cite{li2017deep, arulkumaran2017deep, parisi2019continual, yang2019federated, de2021continual, yin2021comprehensive}. While valuable, these works are often disconnected from mainstream video understanding.
From a dataset-centric perspective, we highlight how specific dataset properties, such as long-horizon egocentric sequences or distributed data collection, can influence the applicability of these strategies, providing guidance on when they may complement standard video understanding pipelines.

\textbf{Classical and deep-learning surveys.} Early surveys focused on handcrafted features and classical pipelines (\eg, spatio-temporal interest points \cite{laptev2005space}, HOG3D \cite{dalal2005histograms}, dense trajectories \cite{wang2013dense}) before tracing the transition to deep learning. Representative examples include Aggarwal and Ryoo \cite{aggarwal2011human} (pre-deep era) and Herath \etal \cite{herath2017going}, which reviewed the evolution from 2D CNNs to 3D CNNs and two-stream architectures. These works documented historical backbone evolution and early benchmark evaluations (\eg, \cite{kuehne2011hmdb, soomro2012ucf101, karpathy2014large}).
While foundational, these surveys were largely architecture-centric and focused on classification tasks. They overlooked how dataset structural properties (\eg, temporal span, compositional complexity, multimodal richness) induce dataset-driven inductive biases in architectures and paradigms, and did not account for the influence of large-scale multimodal pretraining, VLMs, or LLMs on model evolution. In contrast, our survey systematically links dataset characteristics to both architectural and paradigm choices, showing the structural pressures that drive model innovation and task-specific performance.

\textbf{Transformer and modern architecture surveys.} With the rise of attention-based models, several surveys and tutorials have examined transformers in vision and video (\eg, Video Transformers and ViT extension \cite{khan2022transformers, ulhaq2022vision, selva2023video}), analyzing design patterns (\eg, space-time factorization) and computational trade-offs with benchmark comparisons.
While indispensable for understanding architectural design, these surveys remain narrow. They seldom consider (i) dataset-aligned inductive biases beyond attention, such as graph-based structures or generative modeling priors, (ii) interactions between architectures and learning paradigms, including how masked or multimodal objectives reshape transformers' effective biases, or (iii) dataset-driven motivations, such as which dataset properties catalyzed the shift toward transformers. Our work situates transformers within a broader taxonomy and explicitly analyzes how dataset characteristics influence their downstream utility and task-specific performance.

\textbf{Multimodal and VLM/LLM surveys.} As video understanding increasingly incorporates language and audio, surveys have emerged on multimodal learning (\eg, video captioning, retrieval, cross-modal pretraining \cite{miech2020end, xu2023multimodal, madan2024foundation, tang2025video}). These works largely catalogue datasets, alignment objectives, and evaluation protocols without explaining how dataset characteristics shape multimodal model design and paradigm selection.
We advance this perspective by (i) treating multimodal alignment as a dataset-informed paradigm, reflecting invariances induced by different data regimes, (ii) analyzing modality imbalance and modality-dominance failure modes as consequences of dataset properties, and (iii) situating VLM- and LLM-augmented models within hybrid pipelines integrating generative priors and self-supervision, highlighting how dataset structure guides paradigm and architectural choices.

\textbf{Structured representations.} Many surveys cover skeleton-based action recognition and graph-based approaches for modeling human-object interactions, multi-agent dynamics, and relational reasoning~\cite{subetha2016survey, yan2018spatial, wang2019comparative, cheng2020skeleton, ahmad2021graph, mourot2022survey, ren2024survey, liurepresentation}. While valuable, these works largely remain modality-specific and rarely consider how dataset properties influence the choice of representational paradigms.
Our survey situates graph-based representations within a broader architecture and paradigm design space, showing how dataset characteristics, \eg, agent density, relational complexity, compositional depth, drive the selection of graph, attention, or self-supervised representations for effective spatiotemporal and relational reasoning.

\textbf{Self-supervision, generative, and pretraining surveys.} Surveys on self-supervised and generative modeling~\cite{jing2020self,liu2021self, ericsson2022self, oussidi2018deep,suzuki2022survey, cho2024sora, xing2024survey, wang2025feature} summarize objectives, augmentations, and downstream transfer. While informative, they seldom analyze how dataset characteristics guide pretraining paradigm selection.
In contrast, our survey emphasizes alignment between dataset properties and learning paradigms, highlighting when contrastive, masked, or generative objectives are most effective, and how hybridization or curriculum strategies can be guided by structural and temporal dataset properties.

\textbf{Benchmarks and dataset-centered studies.} Several surveys focus on benchmarks \cite{liu2016benchmarking, carreira2017quo, singh2019video, yao2019review, guo2024benchmarking, plizzari2024outlook, liurepresentation}, evaluation practices, and dataset analyses (\eg, Kinetics \cite{kay2017kinetics}, EPIC-KITCHENS \cite{damen2018scaling, damen2022rescaling}). While useful, they often treat datasets as static resources.
In contrast, we adopt datasets as structural lenses, analyzing properties such as temporal span, compositionality, annotation granularity, multimodal richness, and agent density, and show how these drive paradigm and architectural choices, enabling more principled guidance for model design, pretraining, and dataset construction.

\textbf{Reinforcement, continual, and privacy-aware learning.} Some surveys review reinforcement learning, continual learning, and federated/privacy-preserving approaches, highlighting challenges such as catastrophic forgetting, non-i.i.d. data, and sparse rewards~\cite{li2017deep, arulkumaran2017deep, parisi2019continual, yang2019federated, de2021continual, yin2021comprehensive}. While valuable, these works are often disconnected from mainstream video understanding.
From a dataset-centric perspective, we highlight how specific dataset properties, such as long-horizon egocentric sequences or distributed data collection, can influence the applicability of these strategies, providing guidance on when they may complement standard video understanding pipelines.

\section{Dataset Characteristics and Their Architectural Implications}
\label{app:dataset_characteristics}

\subsection{Motion \& Fine-Grained}

\begin{figure}[tbp]
\centering
\includegraphics[trim=1cm 1.5cm 0.5cm 2cm, clip=true, width=\linewidth]{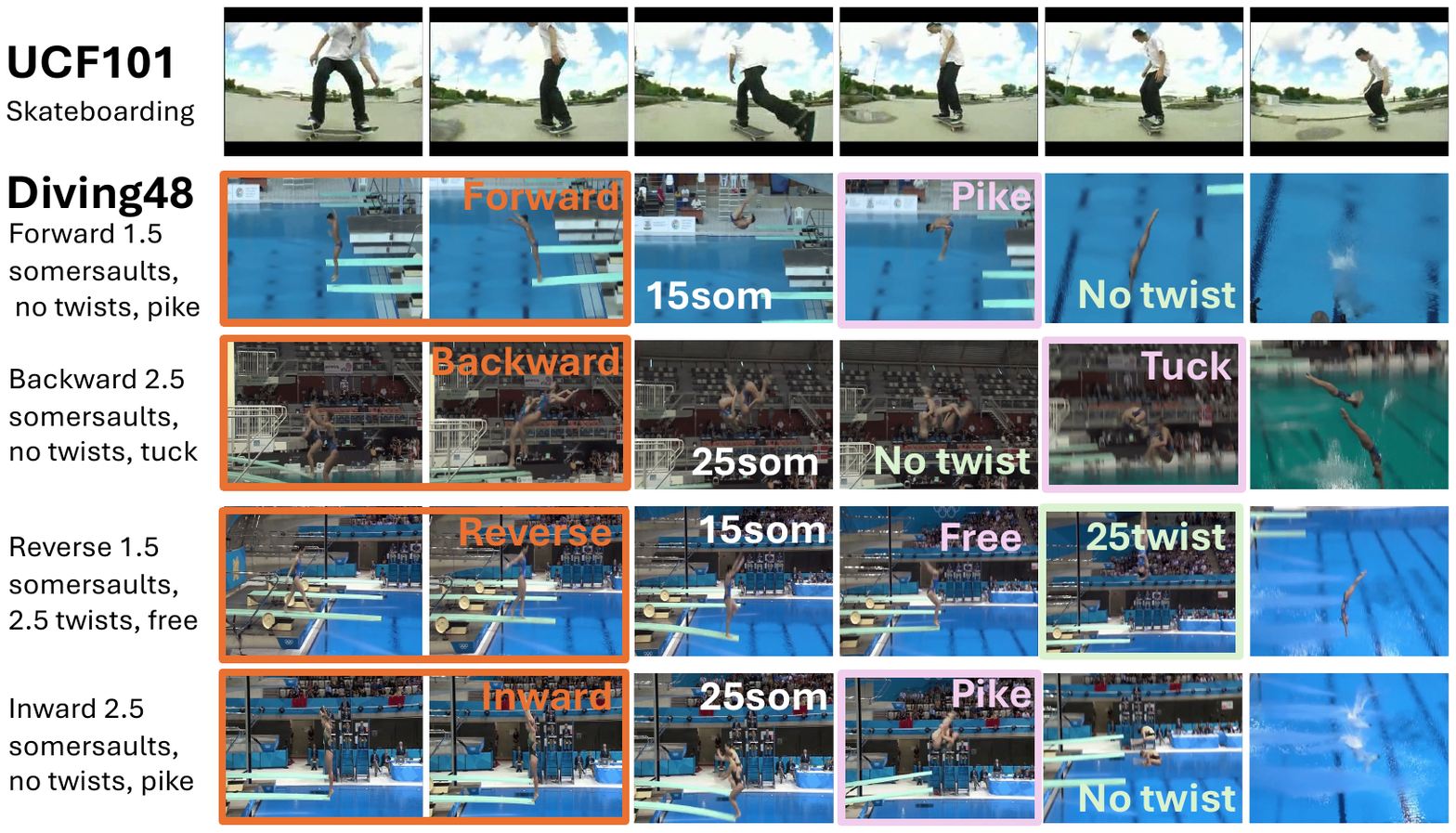}
\caption{Motion complexity across datasets. UCF101 (top) illustrates high-amplitude actions such as skateboarding, where global body motion dominates. Diving48 (bottom) contains four dive categories distinguished by subtle variations in rotation, twist, and posture, requiring fine-grained motion modeling. The contrast reflects the shift from early benchmarks focused on salient whole-body movements to modern datasets that demand recognition of nuanced micro-motions.
}
\label{fig:motion}
\end{figure}

\textbf{Motion complexity.} Motion represents the foundational signal in video understanding, providing critical cues for action discrimination, intent inference, and interaction modeling. Across datasets, the manner in which motion is captured, emphasized, and structured varies dramatically, shaping both model development and evaluation strategies. Early benchmarks, including KTH \cite{schuldt2004recognizing}, Weizmann \cite{blank2005actions}, HMDB51 \cite{kuehne2011hmdb}, and UCF101 \cite{soomro2012ucf101}, illustrate the initial stage of dataset design, where the focus was on clearly perceivable human movements in relatively controlled environments. These datasets primarily feature high-amplitude, visually salient motions, such as running, basketball dunking, or swinging, often captured from third-person viewpoints with limited background clutter. Such design enabled early architectures, \eg, 3D ConvNets, two-stream networks, and optical-flow-based models, to effectively learn coarse global motion patterns, yet provided limited challenge for modeling subtle, context-dependent, or overlapping actions \cite{simonyan2014two, carreira2017quo, feichtenhofer2016convolutional, wang2016temporal}. 

Progressing beyond coarse motion, intermediate datasets introduced more realistic variability in camera angles, scene context, and actor appearance. Hollywood \cite{laptev2008learning}, Hollywood2 \cite{marszalek2009actions}, and Collective Activity \cite{choi2009they} expose models to complex backgrounds, multi-person interactions, and social context, while datasets like MSRAction3D \cite{li2010action} and UTKinect-Action3D \cite{xia2012view} provide depth and skeleton information to capture 3D body dynamics. These datasets necessitate models capable of disentangling actor motion from background variations, handling viewpoint changes, and interpreting relational motion among multiple agents. Similarly, group-activity and social interaction datasets, including Volleyball \cite{ibrahim2016hierarchical} and DALY \cite{weinzaepfel2016human}, require tracking multiple agents simultaneously, highlighting the importance of relational reasoning and graph-structured representations for motion modeling \cite{girdhar2021anticipative, feichtenhofer2019slowfast}.

Recent datasets further emphasize fine-grained, subtle, and context-dependent motion, presenting new challenges for model design. FineGym \cite{shao2020finegym} and Diving48 \cite{kanojia2019attentive} exemplify actions where minor differences in rotational velocity, limb alignment, or entry angle define entirely distinct classes, requiring temporal precision and spatial fidelity. Here, global motion representations are insufficient; models should capture micro-motion, joint trajectories, and nuanced temporal dependencies, motivating architectures such as multi-scale 3D convolutions, temporal transformers, and pose-based representations \cite{wang2020video, sudhakaran2020gate, chen2021sportscap, kim2024learning}. Figure~\ref{fig:motion} illustrates this progression: UCF101 exemplifies high-amplitude, global body motions, whereas Diving48 demonstrates fine-grained distinctions in rotation, twist, and posture, highlighting the evolution from early salient-motion datasets to benchmarks that require nuanced micro-motion modeling. Similarly, AVA \cite{gu2018ava} demonstrates the challenges of multi-person, occluded scenarios where low-amplitude gestures such as handshake or object manipulation should be disambiguated within complex visual contexts. In egocentric datasets like EPIC-KITCHENS \cite{damen2018scaling, damen2022rescaling}, motion complexity is further amplified by camera ego-motion and hand-object interactions, requiring models to decouple actor-induced motion from environmental and self-motion dynamics, a task largely absent in early third-person datasets.

Motion complexity in modern datasets is inherently multi-dimensional, capturing variations in amplitude, temporal coherence, spatial coverage, and multi-agent interactions. High-amplitude actions, such as jumping or running, contrast with low-amplitude, localized gestures like typing or stirring. Temporal coherence distinguishes cyclic patterns, such as dribbling or walking, from discrete, isolated movements. Spatial coverage differentiates whole-body motions from localized actions involving hands, facial expressions, or small objects. Multi-agent interactions introduce additional layers of complexity, requiring models to capture interdependent motions, relational dynamics, and social context. Extended temporal datasets such as Breakfast \cite{kuehne2014language} and Charades \cite{sigurdsson2016hollywood} further demand disentanglement of overlapping motion streams, where concurrent actions like chopping vegetables while attending to boiling water should be understood as temporally and semantically distinct yet relationally dependent \cite{farha2019ms, sener2020temporal, wang2023selective}. Collectively, these complexities underscore that motion is not a single attribute but a multi-faceted property that directly informs dataset design, feature representation, and model architecture.

\textbf{Action concepts.} Beyond motion itself, the semantic conceptualization of actions within datasets significantly shapes model learning, generalization, and reasoning. Early large-scale datasets such as Sports-1M \cite{karpathy2014large} and Kinetics-400 \cite{kay2017kinetics} provide general action categories that emphasize robust recognition of prominent spatiotemporal patterns, including running, jumping, or playing instruments. These coarse labels enable models to capture broad motion trends and object-affordance cues but often fail to discriminate subtle, context-dependent differences, limiting fine-grained generalization. In contrast, specialized benchmarks such as FineGym and Diving48 demand precise discrimination of nuanced variants, requiring models to attend to detailed body alignments, rotations, and micro-temporal cues \cite{shao2020finegym, kanojia2019attentive}. First-person video datasets further integrate object semantics and environmental context, as shown in EPIC-KITCHENS, where differentiating cut tomato from cut cucumber relies not only on hand motion but also on object identity, affordances, and interaction dynamics. Atomic action datasets, including AVA, extend this challenge to multi-agent and overlapping actions, highlighting the necessity of spatiotemporal attention, multi-scale feature extraction, and graph-based relational modeling \cite{feichtenhofer2022masked, girdhar2022omnivore, ryali2023hiera}.

Through this lens, the evolution of motion-focused and fine-grained datasets reflects a broader trajectory in video understanding: from coarse, single-agent, high-amplitude actions to multi-agent, context-rich, and micro-motion-sensitive activities. This progression has directly shaped model design, encouraging the development of architectures capable of disentangling overlapping motion streams, reasoning over relational and temporal hierarchies, and integrating multi-modal cues. By situating model evaluation within this nuanced understanding of motion complexity and action conceptualization, researchers can better assess generalization, robustness, and zero-shot reasoning, providing insights that directly inform both dataset curation and the design of next-generation video models.

\begin{figure*}[tbp]
\centering
\subfloat[Ball sport actions in Kinetics-400, grouped by category (\eg, dribbling, dunking, shooting under Basketball), illustrating taxonomic hierarchies that support generalization and compositional reasoning.]{
\includegraphics[trim=0 0 0 0, clip=true, width=\textwidth]{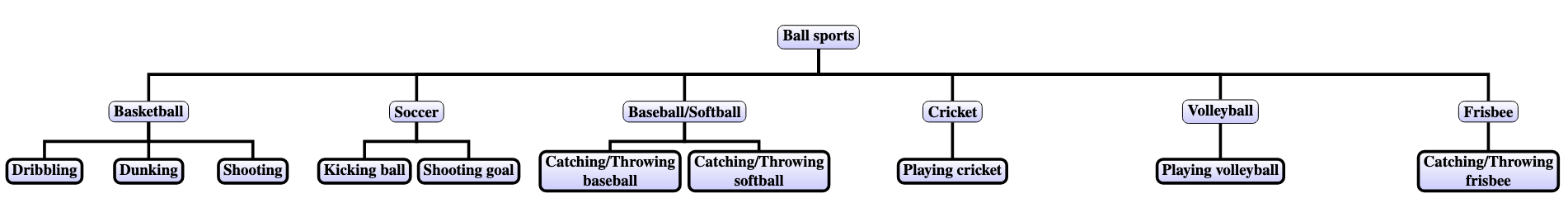}


  \label{fig:ballsports}
}

\subfloat[Making cereal in Breakfast, showing sub-actions (take bowl, pour cereals, pour milk, stir) that form the overall task.]{
  \includegraphics[trim=0 0 0 0, clip=true, width=\textwidth]{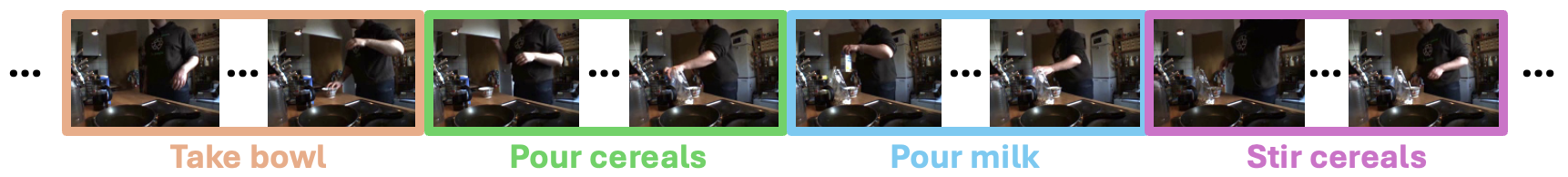}
  \label{fig:breakfast}
}

\caption{Hierarchical and compositional structures in video datasets. (a) Kinetics-400: taxonomic hierarchy of ball sport actions showing semantic groupings that support generalization and compositional reasoning. (b) Breakfast: procedural activity sequence illustrating sub-actions that compose a complete task. 
}
\label{fig:combined}
\end{figure*}

\subsection{Procedural \& Compositional}

Hierarchical structure represents a fundamental dimension of video datasets that is often underexplored. Real-world actions rarely occur in isolation; they are organized both semantically and procedurally, and understanding these relationships is crucial for models that aim to generalize, perform compositional reasoning, and handle multi-step activities. Datasets differ in how they capture these hierarchies, and analyzing these differences reveals structural pressures that shape model design, evaluation, and generalization.

One form of hierarchy is \textbf{taxonomic}, which groups semantically related actions under broader categories. Datasets such as Sports-1M \cite{karpathy2014large} and Kinetics \cite{kay2017kinetics} illustrate this approach: fine-grained classes like soccer, basketball, and tennis are unified under ball sports, providing shared features that models can exploit for recognition and zero-shot generalization. Hierarchically grounded taxonomies enable models to transfer knowledge across semantically similar actions, for example, recognizing handball after learning soccer and basketball, by using structural similarities in objects, motion patterns, and context \cite{yue2015beyond, tran2015learning, qiu2017learning, carreira2017quo, tran2018closer}. Figure~\ref{fig:ballsports} visualizes a taxonomic hierarchy for ball sports in Kinetics-400, grouping related actions such as basketball dribbling, dunking, and shooting under a broader category, which supports semantic generalization and compositional reasoning.

Compositional hierarchies encode \textbf{procedural dependencies} among sub-actions that constitute complex tasks. Instructional datasets, including CAD-60 \cite{sung2012unstructured}, GTEA Gaze \cite{fathi2012learning}, CAD-120 \cite{koppula2013learning}, 50 Salads \cite{stein2013combining}, Breakfast \cite{kuehne2014language}, HowTo100M \cite{miech2019howto100m}, and YouCook2 \cite{zhou2018towards}, capture sequences such as \emph{open milk carton, pour milk, add cereal, stir}. By representing activities as sequences of sub-actions, these datasets encourage models to learn action primitives, reason over temporal structure, and generalize to novel procedural combinations \cite{wu2019long, wu2019sequence, kazakos2019epic, farha2019ms, hussein2019timeception, sun2019videobert, munro2020multi, yang2020temporal, sener2020temporal, gabeur2020multi, girdhar2021anticipative, bertasius2021space, akbari2021vatt, wu2022memvit, islam2022long, wang2022language, wang2023selective, liu2023diffusion, yang2023vid2seq, li2024videomamba, he2024ma, li2024mamba}. Larger-scale egocentric datasets such as EPIC-KITCHENS \cite{damen2018scaling} extend this compositional reasoning to fine-grained hand-object interactions and continuous workflows, highlighting the need for hierarchical transformers, relational graphs, and attention mechanisms capable of capturing both short-term manipulations and long-range dependencies. Figure~\ref{fig:breakfast} illustrates a procedural hierarchy in Breakfast, showing how sub-actions like take bowl, pour cereals, and stir compose a complete task. 

A third type of hierarchy is \textbf{contextual}, which situates actions within their environment, interacting objects, or social context. Identical motions may correspond to different semantic roles depending on context. Datasets such as AVA \cite{gu2018ava} and EPIC-KITCHENS \cite{damen2018scaling} capture these distinctions: lifting a hand may mean picking up a cup, waving hello, or grabbing a dumbbell, depending on surrounding cues and task stage. 
Modeling contextual hierarchies requires integrating visual, temporal, and environmental information, motivating multimodal and context-aware representations \cite{girdhar2019video, feichtenhofer2019slowfast, wu2019long, feichtenhofer2020x3d, qian2021spatiotemporal, wang2022internvideo, wei2022masked, feichtenhofer2022masked, tong2022videomae, wang2023videomae, ryali2023hiera, li2023unmasked}.

Encoding hierarchical, procedural, and contextual relationships directly influences model performance and generalization. Without such hierarchies, models may confuse visually similar sub-actions with distinct semantic roles, \eg, cut carrot versus cut cucumber, or misinterpret multi-step tasks. Hierarchical annotations, whether explicit or inferred, provide the relational structure necessary to reason over atomic actions, their composition, and context. This facilitates zero-shot learning, cross-domain transfer, and the development of architectures that capture dependencies across scales, from sub-action primitives to complex, multi-agent procedures \cite{wu2019long, wu2019sequence, kazakos2019epic, farha2019ms, hussein2019timeception, munro2020multi, yang2020temporal, sener2020temporal, gabeur2020multi, girdhar2021anticipative, bertasius2021space, akbari2021vatt, wu2022memvit, islam2022long, wang2023selective, liu2023diffusion, yang2023vid2seq, li2024videomamba, he2024ma, li2024mamba}. Furthermore, hierarchical datasets enable graded evaluation, reflecting errors at different semantic or procedural levels, and inform the design of structured models such as graph networks, hierarchical transformers, and temporally-aware RNNs/TCNs. By systematically synthesizing procedural, compositional, and contextual hierarchies across datasets of varying scale, viewpoint, and modality, we illuminate recurring design patterns and dataset-driven pressures that have shaped modern video understanding architectures.

\begin{figure*}[tbp]
\centering

\subfloat[MPII Cooking 2: third-person static view showing procedural cooking activities.]{
  \includegraphics[trim=0 0 0 0, clip=true, width=\textwidth]{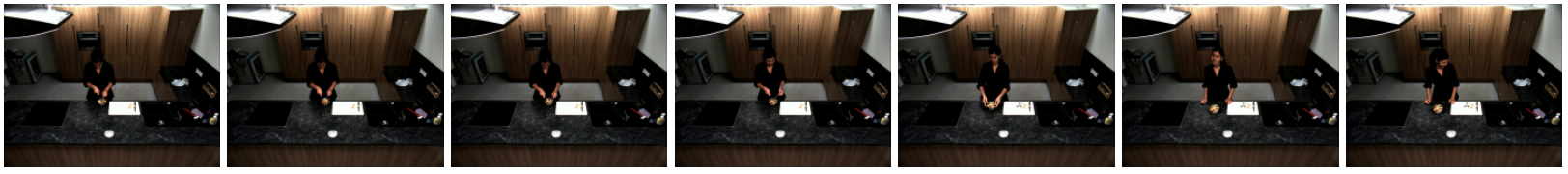}
  \label{fig:mpii}
}

\subfloat[EPIC-KITCHEN-100: egocentric, moving camera capturing continuous hand-object interactions, \eg, open bin, get tomato, put glass, highlighting long-horizon temporal dependencies.]{
  \includegraphics[trim=0 0.1cm 0 0.08cm, clip=true, width=0.653\textwidth]{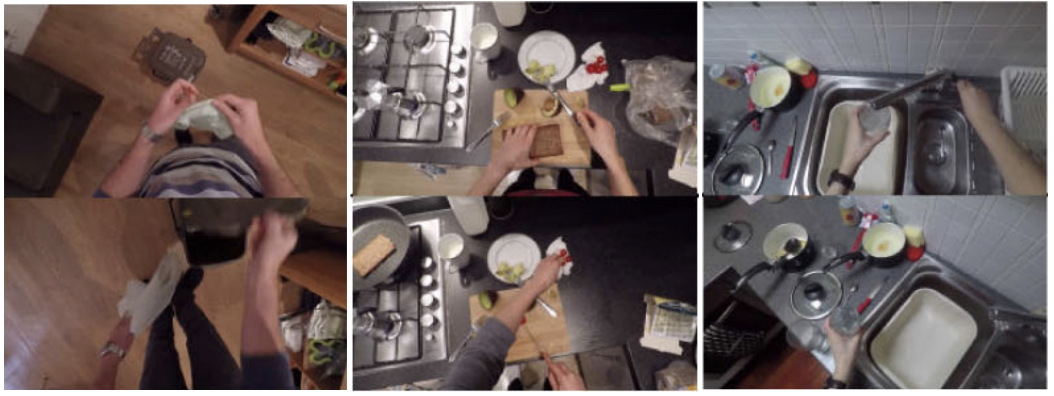}
  \label{fig:epic}
}\hspace{0.15cm}
\subfloat[Charades-Ego: combined third-person and egocentric views capturing overlapping everyday actions.]{
  \includegraphics[trim=0 0 0.01cm 0.05cm, clip=true, width=0.303\textwidth]{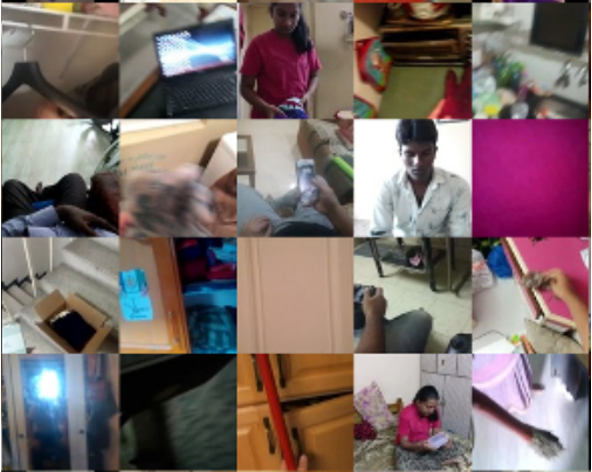}
  \label{fig:charades}
}
\caption{Examples of egocentric and long-horizon video datasets highlighting temporal and procedural complexity. (a) MPII Cooking 2 shows controlled procedural cooking actions from a third-person perspective. (b) EPIC-KITCHENS-100 captures continuous hand-object interactions from an egocentric perspective, emphasizing multi-step workflows and temporal dependencies. (c) Charades-Ego combines third-person and egocentric views, showcasing overlapping everyday activities, which challenge models to reason over concurrent and sequential actions. Collectively, these datasets demonstrate how varying viewpoints, temporal granularity, and action overlap shape model design for long-horizon video understanding.}
\label{fig:ego-combined}
\end{figure*}

\subsection{Egocentric \& Long-horizon}

Recent video datasets increasingly include long, continuous sequences where multiple actions unfold sequentially or concurrently, reflecting real-world activity complexity \cite{kuehne2014language, sigurdsson2016hollywood,sigurdsson2018charades,gu2018ava,damen2018scaling,bain2021frozen,damen2022rescaling,xue2022advancing,wang2023internvid,nan2024openvid,chen2024panda,ju2024miradata,wang2025koala}. For instance, Charades \cite{sigurdsson2016hollywood} captures everyday household activities with temporally overlapping actions, such as \emph{pick up cup while walking to the kitchen}, requiring models to disentangle concurrent motion streams and track multiple sub-actions. Procedural datasets like Breakfast \cite{kuehne2014language} provide dense temporal annotations across multi-step activities; a single breakfast video may sequentially include \emph{cut vegetables, boil water, add pasta, stir sauce}, embedding fine-grained sub-actions within a higher-level activity. Correct interpretation demands hierarchical temporal reasoning, from micro-level motions to macro-level procedural dependencies, motivating architectures with multi-scale temporal modeling \cite{gao2017tall,zhou2018temporal, wang2018non, wang2018videos, feichtenhofer2019slowfast, feichtenhofer2020x3d, fan2021multiscale}.

Egocentric datasets further intensify temporal and relational complexity. EPIC-KITCHENS \cite{damen2018scaling,damen2022rescaling} captures continuous hand-object interactions from a first-person perspective, where actions like \emph{slice cucumber} or \emph{pour milk} occur within fluid workflows including transitions, pauses, and context-dependent variations. Figure~\ref{fig:ego-combined} shows representative examples of such datasets, including third-person procedural recordings (MPII Cooking 2 \cite{rohrbach2012database}), egocentric continuous interactions (EPIC-KITCHENS-100 \cite{damen2018scaling,damen2022rescaling}), and combined third-person/egocentric views (Charades-Ego \cite{sigurdsson2018charades}), highlighting the challenges of long-horizon temporal reasoning, procedural complexity, and multi-view action overlap. Modeling such sequences requires capturing both short-term manipulations and long-range activity dependencies, highlighting the need for hierarchical transformers, recurrent attention mechanisms, or graph-based relational networks \cite{girdhar2019video, feichtenhofer2019slowfast, wu2019long, feichtenhofer2020x3d, qian2021spatiotemporal, wang2022internvideo, wei2022masked, feichtenhofer2022masked, tong2022videomae, wang2023videomae, ryali2023hiera, li2023unmasked}. Ego-motion introduces additional challenges, as camera movement is tightly coupled with the actor's motion, requiring disentanglement of self-motion from object-centric interactions.

Spatial grounding is equally critical. Datasets like AVA \cite{gu2018ava} provide bounding boxes for multiple actors along with temporally localized labels, enabling models to resolve overlapping or interacting actions such as \emph{talk to while pick up object}. Such dense spatiotemporal annotation supports multi-agent reasoning, concurrent action detection, and interaction understanding, which short-clip or flat-label datasets cannot address. Similarly, first-person and multi-view procedural datasets, \eg, CAD-60 \cite{sung2012unstructured}, GTEA Gaze \cite{fathi2012learning}, CAD-120 \cite{koppula2013learning}, 50 Salads \cite{stein2013combining}, YouCook2 \cite{zhou2018towards}, and MOMA \cite{luo2021moma,luo2022moma}, encourage architectures capable of integrating temporal hierarchies, attention to hand-object relations, and relational graph reasoning.

Temporal span, hierarchical structure, and egocentric perspective collectively introduce structural constraints that shape model design \cite{girdhar2019video, feichtenhofer2019slowfast,wu2019long, feichtenhofer2020x3d}. Short-clip or single-action datasets favor frame-based or 3D convolutional networks for local motion modeling, whereas long-horizon datasets like EPIC-KITCHENS-100 \cite{damen2022rescaling} or YouCook2-BoundingBox \cite{zhou2018weakly} demand temporal transformers, memory-aware modules, and graph-based relational networks to capture long-range dependencies, sub-action composition, and multi-agent interactions \cite{qian2021spatiotemporal, wang2022internvideo, wei2022masked, feichtenhofer2022masked, tong2022videomae, wang2023videomae,ryali2023hiera, li2023unmasked}. Procedural and instructional datasets, including Breakfast \cite{kuehne2014language}, COIN \cite{tang2019coin}, and CATER \cite{girdhar2019cater}, further drive hierarchical and step-aware reasoning for sequential task modeling. Egocentric datasets such as GTEA Gaze+ \cite{li2015delving,li2018eye} emphasize attention to gaze, hand-object relations, and perspective-specific dynamics, fostering relational modeling and context-sensitive feature integration. Temporal and spatiotemporal annotations across datasets like AVA \cite{gu2018ava}, Charades-Ego \cite{sigurdsson2018charades}, and PKU-MMD \cite{liu2017pku,liu2020benchmark} reinforce fine-grained localization capabilities, guiding architectures toward concurrent action detection and multi-agent reasoning. 

Collectively, these design pressures show recurring patterns linking dataset properties 
to model evolution, demonstrating how structural biases inherent in datasets actively guide innovation in video understanding architectures.

\subsection{Multimodal Corpora}

Modern video datasets increasingly recognize that visual information alone is often insufficient to fully capture the semantics of actions, particularly in complex, fine-grained, or procedural scenarios. Multimodal signals, including audio, text, and metadata, introduce complementary cues that not only improve performance but also impose structural pressures that guide model design. Audio tracks convey critical information about actions that may be visually subtle or partially occluded. For example, the sound of chopping, pouring, or clinking utensils allows models to disambiguate visually similar gestures, such as cut tomato versus cut cucumber, where visual cues alone are ambiguous. Environmental sounds, such as footsteps, machinery, or applause, provide context for temporal alignment and action recognition in real-world scenarios \cite{gemmeke2017audio,bain2021frozen,grauman2022ego4d,chen2020vggsound,nan2024openvid,wang2023internvid,xue2022advancing,chen2024panda,ju2024miradata,wang2025koala,huh2025epic}. 

Textual modalities enrich video understanding by providing semantic grounding that complements visual and audio information. Instructional and narrated datasets, such as HowTo100M \cite{miech2019howto100m}, include natural language descriptions aligned with video sequences, spanning detailed procedural steps to high-level activity summaries. These annotations enable models to map observed actions to semantic concepts, bridging low-level motion cues and high-level activity reasoning. For instance, a sequence labeled \emph{whisk eggs and add to pan} allows a model to disambiguate visually similar sub-actions such as stir ingredients versus mix batter. Temporal reasoning is also facilitated, as textual cues often describe action sequences that extend beyond the duration of individual visual clips \cite{zhou2018weakly,xu2021videoclip,yan2022videococa,akbari2021vatt}.

Contextual information further shapes understanding by situating actions within their environment. The same motion may correspond to different actions depending on surrounding objects or participants, for example, lifting a hand may indicate pick up cup in a kitchen, wave hello in a social scene, or grab dumbbell in a gym. AVA \cite{gu2018ava} captures such nuances via spatial and contextual annotations in crowded movie scenes, whereas EPIC-KITCHENS \cite{damen2018scaling,damen2022rescaling} records first-person hand-object interactions where object identity and placement are central to action semantics. Multi-agent scenarios emphasize relational reasoning, requiring models to track concurrent actions and interactions over space and time \cite{xu2021videoclip,fu2021violet,li2022align}.

Across datasets, these multimodal properties systematically shape architectural evolution \cite{zhang2023video,maaz2023video,ren2024timechat}. Small-scale captioning datasets, \eg, MSVD \cite{chen2011collecting} and MSR-VTT \cite{xu2016msr}, encourage cross-modal embedding learning. Temporally grounded corpora, including DiDeMo \cite{anne2017localizing} and ActivityNet Captions \cite{krishna2017dense}, enforce fine-grained alignment between visual frames and textual cues, which drives the adoption of frame-level attention and alignment mechanisms. Procedural and instructional datasets, \eg, HowTo100M \cite{miech2019howto100m}, TACOS \cite{regneri2013grounding}, VidChapters-7M \cite{yang2023vidchapters}, and compositional QA datasets like AGQA \cite{grunde2021agqa}, promote hierarchical and memory-aware architectures capable of capturing long-range dependencies and compositional action structure \cite{miech2020end, xu2023multimodal, selva2023video, madan2024foundation, xing2024survey,tang2025video, ding2025language, ding2025journey}. Egocentric and first-person datasets, including EPIC-KITCHENS and Ego4D \cite{grauman2022ego4d}, emphasize relational modeling and attention-based reasoning for hand-object and agent-object interactions. Large-scale pretraining datasets, \eg, WebVid-2M/10M \cite{bain2021frozen}, HD-VILA-100M \cite{xue2022advancing}, InternVid \cite{wang2023internvid}, Panda-70M \cite{chen2024panda}, MiraData \cite{ju2024miradata}, OpenVid \cite{nan2024openvid}, and Koala-36M \cite{wang2025koala}, enable models to generalize across tasks and support zero-shot reasoning in retrieval, captioning, temporal grounding, and question answering \cite{bain2021frozen,xue2022advancing,wang2023internvid,nan2024openvid,chen2024panda,ju2024miradata,wang2025koala,huh2025epic}. 
Multimodal QA datasets, including TVQA \cite{lei2018tvqa}, further highlight the importance of integrating dialogue, subtitles, and temporal reasoning for structured comprehension of complex video content. 

Multimodal corpora show that integrating visual, audio, and textual signals systematically shapes model design, promoting hierarchical, attention-driven, and memory-aware architectures. This unified analysis across datasets highlights how multimodal richness, temporal structure, and scale collectively drive architectural innovations, fulfilling our goal of a dataset-driven synthesis of video understanding challenges.

\section{Additional Benchmark Analysis}

Tables~\ref{tab:video_models} and~\ref{tab:video_models2} benchmark representative video models across recognition, detection, retrieval, localization, and question answering. These results show how dataset properties interact with architectural choices and pretraining strategies. We organize the discussion into two parts: (i) recognition and detection benchmarks focusing on spatiotemporal representation, and (ii) multimodal tasks such as retrieval and question answering that require video-language alignment. 

\subsection{Spatiotemporal Modeling for Recognition and Detection}

Short-clip, motion-centric datasets strongly favor architectures that explicitly model local spatiotemporal dynamics (see Table~\ref{tab:video_models}). On HMDB51 and UCF101, early Two-Stream variants and 3D CNNs consistently outperform others, highlighting the importance of capturing instantaneous motion cues. For instance, Two-Stream'16 achieves 93.5\% on UCF101 and 69.2\% on HMDB51, while RGB-I3D reaches 95.6\% and 74.8\%, respectively. These results suggest that for datasets with short, well-constrained clips, exploiting frame-level motion information through two-stream fusion or early 3D convolutions remains highly effective.
Long-range, compositional datasets benefit from architectures that model temporal dependencies across extended sequences. On Something-Something V1/V2 (SSv1/SSv2), sequential approaches such as TSM, TRN, and TSN show substantial gains over short-term models. For example, TSM attains 66.0\% on SSv2 and 52.6\% on SSv1, while TRN scores 55.5\% and 42.0\% on the same datasets, respectively. This emphasizes that capturing object interactions, temporal ordering, and higher-level sequence composition is critical, as simple motion cues are insufficient for datasets requiring understanding of temporal context.

Procedural and multi-agent datasets demand attention-based or transformer architectures capable of relational reasoning and large-scale context modeling. On AVA v2.2, transformer and self-supervised models such as MaskFeat and VideoMAE achieve 38.8-42.6 mAP, whereas traditional 3D CNNs underperform or are not reported. Epic-Kitchens-100 further highlights this trend: TSM achieves 38.3\% on Action, 67.9\% on Verb, and 49.0\% on Noun, while Motionformer reaches 44.5\%, 67.0\%, and 58.5\%, respectively. These results indicate that procedural or relational tasks benefit from architectures that integrate long-range temporal dependencies with feature attention and pretraining on large datasets.

Large-scale, generic action recognition datasets demonstrate that performance scales with model capacity and pretraining sophistication. On Kinetics-400/600/700, modern transformers (Swin, MViT, VideoMAE, InternVideo) consistently outperform early 3D CNNs, achieving top-1 accuracies above 80\% and up to 84\% (InternVideo and InternVideo2 on K700). This trend shows that dataset scale, model capacity, and self-supervised pretraining collectively influence performance, emphasizing the need for careful model selection when moving from small, constrained datasets to large, diverse video collections.
Cross-dataset patterns show a clear alignment between dataset characteristics and architectural design. Early 3D CNNs and two-stream networks excel at short, motion-sensitive clips, sequential models dominate compositional and interaction-heavy datasets, and transformer/self-supervised models achieve the best results on procedural, multi-agent, or densely annotated benchmarks. This alignment provides a practical roadmap for model selection: matching architectural inductive biases to the temporal, relational, and compositional properties of the target dataset maximizes performance while guiding design choices for scalability.

\subsection{Multimodal Alignment for Retrieval and Reasoning}

Early backbone architectures such as I3D, R(2+1)D, and SlowFast consistently excel in temporal action localization, achieving strong mAP on THUMOS'14 (up to 66.8\%) and moderate performance on ActivityNet and HACS (see Table~\ref{tab:video_models2}). These results highlight their ability to capture fine-grained spatiotemporal motion patterns. However, their lack of reported performance on retrieval or question answering benchmarks (\eg, MSR-VTT, MSVD-QA) highlights their limited capacity for multimodal reasoning or language grounding. This establishes a baseline: temporal modeling alone is insufficient for tasks that require semantic alignment across video and text.
The emergence of video-language pretrained models marks the next major shift. Architectures such as CLIP4Clip, Frozen, VIOLET, and ALPRO achieve substantial gains on retrieval benchmarks, with MSR-VTT R@1 reaching 32.5--42.1\%. These gains show that large-scale pretraining on paired video-text data enables robust cross-modal alignment, facilitating both retrieval and video question answering. Indeed, models like VideoCLIP and VIOLET also demonstrate strong QA performance (\eg, 92.1\% on MSVD-QA, 68.9\% on TGIF-QA), showing that semantic transfer extends beyond simple retrieval. Nevertheless, these models generally report limited or no performance on temporal localization benchmarks, suggesting that video-language alignment alone cannot fully replace specialized temporal reasoning.

The most pronounced performance leap occurs with large-scale models such as InternVideo and InternVideo2. These models achieve state-of-the-art retrieval results across multiple datasets (MSR-VTT: 62.8\%, ActivityNet: 74.1\%, VATEX: 75.5\%) while maintaining strong localization performance (THUMOS'14 mAP: 72.0\%). They also exhibit high accuracy on multiple-choice QA (MSR-VTT: 93.4\%, LSMDC: 77.3\%), highlighting their ability to generalize multimodal understanding to structured reasoning tasks. These results demonstrate that scaling both model size and pretraining diversity enhances not only cross-modal alignment but also downstream adaptability across datasets of varying complexity and granularity.
Instruction-tuned video-language models such as Video-ChatGPT, Valley, and Grounding-GPT introduce a complementary paradigm. By aligning video understanding with natural language instructions, these models excel in video QA, achieving top-1 accuracies exceeding 49\% on MSR-VTT-QA and TGIF-QA, despite minimal supervised fine-tuning. Their performance highlights the potential of instruction tuning to enable flexible, open-ended reasoning, although retrieval metrics remain largely unreported.

Across datasets, several trends emerge. Short-clip retrieval datasets like MSR-VTT and MSVD benefit most from large-scale video-language pretraining, whereas long-range, compositional datasets such as ActivityNet and DiDeMo show the importance of temporal reasoning. QA datasets, including MSVD-QA and TGIF-QA, require both semantic alignment and multimodal reasoning, favoring instruction-tuned architectures. Multiple-choice settings further emphasize the benefits of models that can integrate retrieval, localization, and reasoning capabilities.
Taken together, these patterns illustrate a clear trajectory: from modality-specific backbones optimized for temporal cues, through multimodal alignment via video-language pretraining, to instruction-tuned models capable of flexible zero-shot reasoning. Future video models will likely need to merge the temporal precision of early architectures with the broad multimodal and reasoning capabilities of large-scale, instruction-aligned systems, advancing toward truly general-purpose video understanding.

\section{Extended Perspectives on Dataset-Bias-Architecture Co-Evolution}

We now provide a prescriptive roadmap showing how dataset properties shape architectures, guiding model selection while balancing scalability and deployment.

\subsection{Dataset Limitations and Future Outlook}

\textbf{Dataset limitations.} Despite their pivotal role in shaping model architectures, existing datasets remain constrained by structural limitations that directly influence what models learn and how they generalize. A first challenge is dataset bias. Many benchmarks reflect narrow cultural or environmental contexts, \eg, sports, kitchens, or scripted movies, leading to strong priors that models can exploit without acquiring robust spatiotemporal reasoning skills. Architectures trained on such datasets may achieve high benchmark accuracy yet falter in real-world deployments, where actions, objects, and environments differ markedly from the training distribution. Bias and imbalance in class coverage further skew learning dynamics, amplifying context-specific shortcuts rather than transferable representations.
A second limitation lies in annotation cost and granularity. Fine-grained temporal labels, hierarchical task decompositions, and multimodal alignments are expensive and time-consuming to obtain. Consequently, many datasets provide only sparse or weak supervision, with limited temporal density or noisy boundaries. These constraints have architectural consequences: models trained under such supervision often overfit to annotated segments while ignoring unlabelled structure, motivating the rise of weakly supervised, self-supervised, and semi-automatic approaches that attempt to compensate for annotation sparsity.
Third, many datasets lack ecological validity. Curated short clips and trimmed action boundaries capture isolated moments rather than continuous, overlapping, and ambiguous workflows that typify everyday activity. As a result, architectures optimized for such curated data, such as clip-based 3D CNNs or trimmed-sequence transformers, struggle when confronted with egocentric videos, multi-agent dynamics, or long-horizon reasoning tasks. The gap between benchmark data and real-world complexity has fueled interest in architectures that incorporate memory, relational reasoning, causal inference, and multimodal grounding to cope with unconstrained environments.
Finally, evaluation remains fragmented across datasets, with heterogeneous metrics and inconsistent protocols that make it difficult to assess generalization (Tables \ref{tab:video_models} and \ref{tab:video_models2}). Models are often fine-tuned to maximize benchmark-specific accuracy or mean average precision, rather than being evaluated for broader capabilities such as compositional reasoning, causal inference, or robustness to distributional shift. This fragmentation not only obscures comparative progress but also shapes architectural incentives, leading to models tuned for leaderboard performance rather than general-purpose understanding.

\textbf{Future outlook.} These limitations point directly to the requirements of next-generation datasets and models. Future benchmarks should move beyond static, domain-specific corpora to embrace diversity, ecological validity, and scalability. Diversity involves curating datasets that capture a broad range of cultures, environments, and activity types, helping to reduce biases that limit generalization. Ecological validity requires capturing continuous, untrimmed, multimodal, and multi-agent activities, enabling models to reason over overlapping workflows and dynamic social contexts. Scalability demands annotation strategies that combine automated labeling, crowd-sourcing, and self-supervised alignment, ensuring that datasets can grow without prohibitive human cost. Crucially, future datasets should be explicitly designed to support not only recognition and localization but also higher-level reasoning tasks such as forecasting, causal analysis, and interactive decision-making.
Architectures should co-evolve to meet these demands. Long-horizon reasoning will require models with structured memory, hierarchical temporal abstractions, and recurrent attention mechanisms capable of spanning minutes or even hours of activity. Relational and causal modeling will benefit from graph-based and neuro-symbolic hybrids that can disentangle inter-agent dependencies and infer cause-effect relationships. Multimodal grounding will necessitate foundation models that seamlessly integrate visual, auditory, textual, and sensor streams, with mechanisms for balancing modality dominance and coping with misalignment. Moreover, continual and adaptive learning will become essential as datasets increasingly reflect dynamic, open-world conditions where models should adapt to new tasks, domains, and modalities without catastrophic forgetting.

Alongside dataset and architectural innovation, benchmarking practices should also evolve. Standardized cross-dataset protocols can provide a measure of true generalization, while new evaluation metrics should capture dimensions such as compositional generalization, reasoning faithfulness, robustness under noise, and computational efficiency. Open benchmarks that test causal inference, multi-step prediction, and cross-modal reasoning will better reflect the capabilities demanded by real-world video understanding systems.

Taken together, these findings highlight the mutual relationship between datasets and architectures: current benchmark limitations expose model blind spots, while future model requirements drive the need for more representative, scalable, and challenging datasets. Bridging this gap will require community-wide collaboration in dataset curation, annotation, benchmarking, and model design. If pursued systematically, this agenda has the potential to transform video understanding from narrow task performance toward general-purpose, robust, and socially responsible systems, closing the loop between data, inductive bias, and architectural evolution.

\subsection{Datasets as Engines of Architectural Innovation}

Datasets are not passive benchmarks; they are the principal structural force shaping model design. As summarized in Table II and reflected in performance trends in Tables \ref{tab:video_models} and \ref{tab:video_models2}, every major architectural transition in video understanding has been catalyzed by properties encoded in the data: short, trimmed motion corpora favored two-stream CNNs and early 3D ConvNets; long-range sequential datasets demanded temporal aggregation and memory; and multimodal, text-paired corpora precipitated cross-modal transformers and video-language pretraining. In this sense, datasets operate as inductive-bias generators that determine which invariances, \eg, temporal, relational, and semantic, models should internalize to succeed.

Motion complexity sets the limits of generalization. Coarse, high-amplitude datasets rewarded architectures that capture instantaneous motion (\eg, optical-flow streams and shallow 3D filters), but those same inductive biases often fail in cluttered or low-amplitude regimes. FineGym, Diving48, and AVA illustrate the opposite pressures: fine-grained micro-motions, multi-agent interactions, and sparse cues force models toward multi-scale temporal hierarchies, pose/part reasoning, and attention mechanisms that can isolate salient sub-trajectories. The practical implication is clear: training exclusively on coarse-action datasets leads to fragile transfer performance. Motion granularity should be present in the data if we expect robustness in nuanced real-world settings.
Compositional and hierarchical structure unlocks procedural reasoning. Instructional datasets such as Breakfast, YouCook2, COIN, and large weakly aligned corpora like HowTo100M show that activities are not atomic labels but sequences of sub-actions arranged taxonomically and contextually. Exposure to such structure allows models to learn reusable primitives and combine them zero-shot into unseen activities, an ability crucial for robotics, assistive AI, and instructional video analysis. Table II makes this visible in the \emph{Step/Hier} annotation columns: when the dataset records relations among actions rather than only their names, models learn mechanisms that transfer.

Temporal richness and multimodal alignment provide the foundation for deployable systems. Long-horizon corpora and egocentric datasets (\eg, Charades, EPIC-KITCHENS, Ego4D) compel models to track extended dependencies, handle overlapping actions, and disentangle ego-motion. When videos are paired with language, audio, and narration (\eg, ActivityNet Captions, HowTo100M), models are pressured to ground perception in text and sound, enabling cross-modal retrieval and zero-shot generalization. The performance patterns in Table~\ref{tab:video_models2} mirror this: video-language pretraining and instruction tuning lift retrieval and QA substantially, while pure visual pretraining alone is insufficient for semantic grounding.

These observations also expose concrete dataset gaps. Web-scale video-text corpora deliver breadth, yet their captions are noisy, weakly aligned in time, and culturally/language biased; fine-grained manipulation and low-amplitude motions are underrepresented; spatio-temporal boxes remain sparse outside a handful of benchmarks; and true long-horizon, multi-agent, multimodal datasets with dense alignment are rare. Cross-view and ego-exo bridging are inconsistently available, and compositional generalization is seldom stress-tested with principled splits. Addressing these gaps requires treating dataset design as a strategic lever rather than a scaling exercise. Simply enlarging class vocabularies or clip counts will not yield general video intelligence. The decisive ingredient is \emph{structure}: motion granularity, procedural and hierarchical annotations, temporal continuity across minutes or hours, precise audio/text alignment at sub-second resolution, and evaluation splits that diagnose composition and transfer. Datasets designed with these properties have historically driven the architectural innovations that underpin the field today; future corpora should be crafted to sustain this virtuous cycle.

\subsection{Open Challenges and Future Directions}

The patterns observed in Tables~\ref{tab:video_models} and \ref{tab:video_models2} illustrate a clear principle: datasets generate invariance pressures, which architectures evolve to accommodate. Trimmed, homogeneous clips permitted two-stream and 3D CNNs to dominate short-clip recognition; longer, untrimmed activities exposed their rigidity and motivated temporal convolutions and recurrent aggregation; long-horizon reasoning elevated transformers with scalable attention; rich human-object and multi-agent interactions stimulated relational encoders and graph-augmented models; and, finally, multimodal corpora forced fusion modules and large-scale video-language pretraining. The prescription that follows is not a linear ``replace the old with the new'', but a matching of inductive bias to dataset structure, together with a plan to close the remaining gaps.

For short and relatively homogeneous data, compact 3D CNNs and hybrid CNN-transformer encoders remain efficient and competitive, particularly when deployment constraints prioritize throughput and latency. As temporal span grows and events occur concurrently, attention mechanisms with memory compression, segment-level pooling, and hierarchical temporal pyramids become essential to preserve long-range context without sacrificing resolution at action boundaries. Where multi-agent interactions and human-object relations are central, relational modules and graph-enhanced transformers provide the inductive bias to track entities, roles, and contact dynamics over time. When audio and language are present, alignment losses and contrastive or generative video-language pretraining become the standard framework for retrieval, captioning, and question answering, as reflected in the substantial gains of CLIP-style and instruction-tuned models.

However, two system-level gaps remain. The first is \emph{temporal-semantic unification}: models that achieve precise temporal localization often lack open-vocabulary semantic understanding, whereas models with strong semantic capabilities (\eg, retrieval or QA) perform poorly at boundary-level localization.
A promising direction is to couple dense temporal detectors with token-level audio-text grounding, sharing representations across localization and language understanding heads. The second is \emph{long-horizon compositional reasoning}: current instruction-tuned systems excel at zero-shot QA but degrade on extended, multi-step procedures. Here, retrieval-augmented video understanding, indexing events and steps into a persistent memory and querying them with language, offers a practical path forward, especially when paired with datasets that provide chaptering, steps, and cross-modal timestamps.
Progress also depends on evaluation that measures what matters. Alongside top-1 and R@1, reporting should emphasize temporal mAP across IoU thresholds for localization, moment retrieval under compositional splits, QA accuracy under counterfactual and long-horizon subsets, calibration (\eg, ECE/Brier) for safety-critical use, and compute/latency metrics for deployment. Cross-dataset testing, ego to exo, lab to in-the-wild, language and culture shifts, should become a first-class protocol rather than an afterthought. These practices are directly tied to the contributions: we implement a dataset-centric perspective, relate structural properties to observed performance trends across recognition, localization, retrieval, and QA, and convert this synthesis into concrete guidance for selecting and designing models under real-world constraints.

The roadmap is therefore dual. On the modeling side, pursue architectures that integrate the temporal precision of CNNs, the hierarchical composition needed for procedures, the scalability of transformers for long horizons, and the grounding provided by multimodal pretraining and instruction tuning. On the data side, build the corpora that will pressure such models to emerge: long-form, multi-agent, multimodal datasets with sub-second alignment; annotations that expose steps, roles, and relations; splits that test compositional generalization and domain transfer; and curated hard negatives that probe fine-grained motion and language disambiguation. If dataset and model co-evolve along these lines, the field can move beyond recognition of isolated clips toward robust, general, and deployable video understanding, precisely the dataset-driven vision advanced by this survey.